\PassOptionsToPackage{table}{xcolor}
\documentclass[10pt]{article} %
\usepackage[accepted]{tmlr}
\usepackage[breaklinks,colorlinks,allcolors=mydarkblue]{hyperref}
\usepackage{url}

\newcommand{\rotarrow}{%
	\begin{tikzpicture}[scale=1,
		syncarrow/.style={
			line width=8pt,                    %
			line cap=round, line join=round,   %
			-{Stealth[round,length=28pt,width=26pt,bend]}  %
		}]
		\draw[syncarrow] (135:1) arc (135:-45:1);
		\draw[syncarrow] (-55:1) arc (-45:-225:1);
\end{tikzpicture}}

\title{Sequential Causal Discovery with Noisy \\ Language Model Priors}

\author{\name Prakhar Verma\thanks{Work done during an internship with Adobe Research} \email prakhar.verma@aalto.fi \\
      \addr ELLIS Institute Finland and Aalto University
      \AND
      \name David Arbour  \email arbour@adobe.com \\
      \addr Adobe Research
      \AND
      \name Sunav Choudhary  \email schoudha@adobe.com \\
      \addr Adobe Research
      \AND
      \name Harshita Chopra\thanks{Work done while the author was with Adobe Research} \email hchopra3@uw.edu \\
      \addr University of Washington, Seattle
      \AND
      \name Arno Solin  \email arno.solin@aalto.fi \\
      \addr ELLIS Institute Finland and Aalto University
      \AND
      \name Atanu R.\ Sinha  \email atr@adobe.com \\
      \addr Adobe Research
}

\def\month{04}  %
\def\year{2026} %
\def\openreview{\url{https://openreview.net/forum?id=wFs71JzEO7}} %

\usepackage{xcolor}
\definecolor{mydarkblue}{rgb}{0,0.08,0.45} 

\usepackage[utf8]{inputenc} %
\usepackage[T1]{fontenc}    %
\usepackage{url}            %
\usepackage{booktabs}       %
\usepackage{amsfonts}       %
\usepackage{nicefrac}       %
\usepackage{microtype}      %
\usepackage{amsmath}
\usepackage{amssymb}
\usepackage{mathtools}
\usepackage{amsthm}
\usepackage{bm}
\usepackage{multirow}


\usepackage{algorithm}
\usepackage{algorithmic}
\usepackage{arydshln}
\usepackage{tikz}
\usetikzlibrary{calc,arrows.meta,fit,backgrounds}
\usetikzlibrary{decorations.pathmorphing}
\usetikzlibrary{decorations.markings}
\usetikzlibrary{arrows.meta, shapes.misc}
\usetikzlibrary{bending}
\usepackage{pgfplots}
\usepackage{subcaption}
\usepackage{pifont}
\usepackage{fontawesome5}
\usepackage{enumitem}
\usepackage{wrapfig}
\algsetup{linenosize=\tiny}

\definecolor{PastelSky}{RGB}{204,229,255}
\colorlet{highlight}{PastelSky!50}

\usepackage[capitalize,nameinlink]{cleveref}
\crefname{section}{Sec.}{Secs.}
\crefname{proposition}{Prop.}{Props.}
\crefname{lemma}{Lem.}{Lems.}
\crefname{model}{Mod.}{Mods.}
\crefname{appendix}{App.}{Apps.}
\crefname{algorithm}{Alg.}{Algs.}
\usepackage{xspace}

\newcommand{\xmark}{\textcolor{black!30!red}{\ding{55}}}%

\renewcommand{\mid}{\,|\,}

\newlength{\figurewidth}
\newlength{\figureheight}

\usepackage{listings}
\usetikzlibrary{arrows.meta, positioning, shapes.geometric, fit}

\newcommand{\DD}{\mathbf{D}}
\newcommand{\HH}{\mathcal{H}}
\newcommand{\R}{\mathbb{R}}
\newcommand{\G}{\mathcal{G}}
\newcommand{\B}{\mathcal{B}}
\newcommand{\V}{\mathbf{V}}
\newcommand{\E}{\mathbf{E}}

\newcommand{\N}{\mathcal{N}}
\newcommand{\vtheta}{\bm{\theta}}
\newcommand{\vphi}{\bm{\phi}}
\newcommand{\vm}{\bm{m}}
\newcommand{\MS}{\bm{S}}

\newcommand{\GD}[1]{\G^{\text{D\textsubscript{#1}}}}
\newcommand{\VD}{\V^{\text{O}}}
\newcommand{\VL}{\V^{\text{L}}}
\newcommand{\ED}[1]{\E^{\text{D\textsubscript{#1}}}}

\newcommand{\GE}[1]{\G^{\text{X\textsubscript{#1}}}}
\newcommand{\VE}[1]{\V^{\text{X\textsubscript{#1}}}}
\newcommand{\EE}[1]{\E^{\text{X\textsubscript{#1}}}}
\newcommand{\IE}{\mathcal{I}^{\E}}
\newcommand{\IL}{\mathcal{I}^{\text{L}}}
\newcommand{\IP}{\mathcal{I}^{\text{P}}}
\newcommand{\IR}{\mathcal{I}^{\rho}}
\newcommand{\HE}[1]{\HH^{\E_{#1}}}
\newcommand{\HL}[1]{\HH^{\text{L}_{#1}}}
\newcommand{\me}{m^\E}
\newcommand{\ml}{m^\text{L}}
\newcommand{\vthetaO}{\vtheta^{{\text{O}}}}
\newcommand{\vthetaL}{\vtheta^{{\text{L}}}}

\newcommand{\eg}{\textit{e.g.,}\xspace}
\newcommand{\ie}{\textit{i.e.}\xspace}

\newcommand{\cf}{\textit{cf.}\xspace}

\newcommand{\iid}{\textit{i.i.d.}\xspace}

\newcommand{\arrowdot}{{\leftarrow \hspace{-.4em} \circ}}
\newcommand{\dotarrow}{{\circ \hspace{-.4em} \rightarrow}}
\newcommand{\dotarrowdot}{{\circ \hspace{-.4em} - \hspace{-.4em} \circ}}

\newcommand{\fllm}{f_\text{LM}}

\newcommand{\val}[2]{$#1$\textcolor{gray}{\tiny ${\pm}#2$}} 
\newcommand{\valb}[2]{$\mathbf{#1}$\textcolor{gray}{\tiny ${\pm}\mathbf{#2}$}} 

\renewcommand{\paragraph}[1]{\textbf{#1}~~}

\newcommand{\ours}{NLPSCM\xspace}
\newcommand{\dataset}[2]{\tikz\node[align=center]{\sc \strut#1 \\[2pt] \strut($d\,{=}\,#2$)};}

\begin{document}

\maketitle

\begin{abstract}
Causal discovery from observational data typically assumes access to complete data and availability of perfect domain experts. In practice, data often arrive in batches, are subject to sampling bias, and expert knowledge is scarce. Language Models (LMs) offer a surrogate for expert knowledge but suffer from hallucinations, inconsistencies, and bias. We present a hybrid framework that bridges these gaps by adaptively integrating sequential batch data with LM-derived noisy, expert knowledge while accounting for both \emph{data-induced} and \emph{LM-induced} biases. We propose a representation shift from Directed Acyclic Graph (DAG) to Partial Ancestral Graph (PAG), that accommodates ambiguities within a coherent framework, allowing grounding the \emph{global} LM knowledge in \emph{local} observational data. To guide LM interactions, we use a sequential optimization scheme that adaptively queries the most informative edges. Across varied datasets and LMs, we outperform prior work in structural accuracy and extend to parameter estimation, showing robustness to LM noise.
\end{abstract}

\section{Introduction}
Inference of causal relations from observational data remains a challenge in applications across healthcare, economics, business, and scientific discovery \citep{sanchez2022causal, tu2019neuropathic, sadeghi2023causal, ebert2012causal}. The challenge is addressed through a dual approach: applying causal learning algorithms to observational data while incorporating domain expertise to resolve structural uncertainties \citep{spirtes2000causation, neapolitan2004learning, spirtes2016causal, chickering2002optimal}. However, domain expertise can be a scarce resource \citep{he2008active, choo2023subset, mooij2016distinguishing, meek2013causal, constantinou2023impact}. Advanced Language Models (LMs) create opportunities to explore their potential as surrogate experts for causal discovery \citep{kiciman2023causal, willig2022can}. LMs generate informative priors or contraints \citep{takayama2025integratinglargelanguagemodels, long2022can, ban2023query}, improving accuracy when combined with data-driven algorithms. Yet, LMs pose their own challenges: hallucination, inconsistency, or failure to capture context-specific nuances \citep{ji2023survey, kiciman2023causal}. 

The challenges compound since in common applications, observational data arrive batch-wise at a cadence, instead of as a complete dataset. Examples include web and app metrics of all online firms, where, for example, data could arrive weekly. Privacy regulations and storage constraints may further restrict data access to a short look-back window. A given week's (batch) data may not be representative of the overall distribution, which we assume remains stationary. This kind of arrival introduces \emph{data-induced} bias, since the non-random draw of a batch suffers from sample selection bias (hereafter, selection bias~\cite{spirtes1995causal}) that distorts causal discovery. Separately, use of LMs, including \textit{large} ones, poses two problems: {\em (i)}~As surrogates for domain expertise, LMs introduce an \emph{LM-induced} bias---their responses in terms of informative causal priors are prone to hallucinations, contextual brittleness, and inconsistency \citep{ji2023survey, kiciman2023causal}. {\em (ii)}~The \textit{global} knowledge encoded in LMs may not align with domain-specific \textit{local} patterns emblematic of batch-wise data, leading to potentially biased learning. 

Inattention to the dual biases---data-induced and LM-induced---is a key gap in current approaches to causal discovery with LMs, which we address. First, 
we propose a change in representation shift from a Directed Acyclic Graph (DAG), which LM-augmented causal discovery methods currently use, to a Partial Ancestral Graph (PAG), to accommodate uncertainty in the causal structure arising from the dual biases. Second, we propose a novel Bayesian-inspired approach to causal structure discovery, where beliefs over causal structure are updated with new data-batch, while augmenting noisy LM-knowledge as priors
In support of PAG, we show that popular methods of pairwise and triplet prompting are \textit{overly-optimistic} (\cf \cref{tbl:llm_optimistic_nature}) and generate unreliable causal structure in the form of a DAG.

\begin{table*}[t!] 
	\scriptsize
	\caption{\textbf{LMs are overly optimistic:} LM based DAG-Pairwise and DAG-triplet prompting methods achieve high recall with low precision across temperatures on two common causal discovery datasets. This limitation calls for explicitly modeling data and LM biases. %
	SHD=Structural Hamming Distance.}
\label{tbl:llm_optimistic_nature}
\vspace*{-4pt}
\setlength{\tabcolsep}{4pt}
\begin{tikzpicture}
	\node (table) [inner sep=0pt] {%
\begin{tabular}{ccl ccc ccc}
	\toprule
	\textbf{Dataset} & \textbf{Temp.} & \textbf{Method} & \multicolumn{3}{c}{\textbf{GPT-3.5\textsubscript{turbo}}} &\multicolumn{3}{c}{\textbf{GPT-4o}}  \\
	\midrule
	&&&\textbf{SHD} $\downarrow$ & \textbf{Precision} $\uparrow$ & \textbf{Recall} $\uparrow$ & \textbf{SHD} $\downarrow$ & \textbf{Precision} $\uparrow$ & \textbf{Recall} $\uparrow$ \\
	\midrule
	\multirow{6}{*}{\rotatebox{90}{\sc Earthquake}} 
	& \multirow{2}{*}{0.0}   & Pairwise & \val{3.0}{0.0} & \val{0.57}{0.00} & \val{1.0}{0.0} & \val{2.0}{0.0} & \val{0.67}{0.00} & \val{1.0}{0.0} \\
	&     & Triplet  & \val{2.1}{0.3}  & \val{0.66}{0.03} & \val{1.0}{0.0}  & \val{4.8}{0.4} & \val{0.46}{0.02} & \val{1.0}{0.0}  \\
	\cdashline{2-9}[.4pt/2pt]
	& \multirow{2}{*}{0.5} & Pairwise & \val{2.8}{0.6} & \val{0.59}{0.05} & \val{1.0}{0.0}  & \val{2.2}{0.4} & \val{0.65}{0.04} & \val{1.0}{0.0} \\
	&     & Triplet  & \val{2.1}{0.3}  & \val{0.66}{0.03} & \val{1.0}{0.0}  & \val{4.4}{0.5} & \val{0.48}{0.03} & \val{1.0}{0.0} \\
		\cdashline{2-9}[.4pt/2pt]
	& \multirow{2}{*}{1.0}  & Pairwise & \val{3.2}{0.6} & \val{0.56}{0.05} & \val{1.0}{0.0}  & \val{1.8}{0.4} & \val{0.69}{0.05} & \val{1.0}{0.0} \\
	&     & Triplet  & \val{2.0}{0.6}   & \val{0.67}{0.07} & \val{1.0}{0.0}  & \val{6.2}{0.8} & \val{0.39}{0.03} & \val{1.0}{0.0} \\
	\midrule
	\multirow{6}{*}{\rotatebox{90}{\sc Asia}} 
	& \multirow{2}{*}{0.0}   & Pairwise & \val{25.2}{0.4}   & \val{0.24}{0.00} & \val{1.0}{0.0}  &\val{11.2}{0.4}& \val{0.42}{0.01} & \val{1.0}{0.0} \\
	&     & Triplet  & \val{24.5}{0.5}  & \val{0.25}{0.00} & \val{1.0}{0.0}  & \val{19.0}{0.6} & \val{0.30}{0.01} & \val{1.0}{0.0} \\
			\cdashline{2-9}[.4pt/2pt]
	& \multirow{2}{*}{0.5} & Pairwise & \val{23.4}{1.0}  & \val{0.26}{0.01} & \val{1.0}{0.0}  & \val{11.6}{0.5} & \val{0.48}{0.01} & \val{1.0}{0.0} \\
	&     & Triplet  & \val{24.0}{1.0}      & \val{0.25}{0.01} & \val{1.0}{0.0}  & \val{19.2}{0.8} &\val{0.29}{0.01}&\val{1.0}{0.0} \\
			\cdashline{2-9}[.4pt/2pt]
	& \multirow{2}{*}{1.0}   & Pairwise & \val{23.2}{1.5} & \val{0.25}{0.02} & \val{0.9}{0.1}  & \val{11.8}{0.4} & \val{0.40}{0.01} & \val{1.0}{0.0}  \\
	&     & Triplet  & \val{23.3}{1.0} & \val{0.26}{0.01} & \val{1.0}{0.0}  &\val{26.2}{1.7}&\val{0.23}{0.01}&\val{1.0}{0.0}  \\
	\bottomrule
\end{tabular}
	};
	
	\begin{scope}[on background layer]
		\draw[fill=highlight,draw=none,rounded corners,draw=none,dashed] 
		($(table.north east)+(0,-.5cm)$) rectangle ($(table.south east)+(-3.15cm,0)$);
		\draw[fill=highlight,draw=none,rounded corners,draw=none,dashed] 
		($(table.north east)+(-4.5cm,-.5cm)$) rectangle ($(table.south east)+(-7.65cm,0)$);
		
		\node[draw=none,fill=none,rounded corners=2pt,inner xsep=4pt,text width=8em] (box) at (8,1) {Overly optimistic behavior of the LM experts lead to high recall and low precision.};
		
		\node[draw=none,fill=none,rounded corners=2pt,inner xsep=4pt,text width=8em] (box2) at (8,-1.2) {\xmark~No \emph{global} \\ \phantom{\xmark}~causal structure\\ \xmark~No grounding \\ \phantom{\xmark}~to \emph{local} data\\\xmark~Need heuristics \\ \phantom{\xmark}~to postprocess};
		
		\draw[line width=1.5pt,highlight] (box.north west) -- (box.south west);
		
		\draw[highlight,line width=1.5pt] (box.west) edge[bend right=30] ($(table.north east)+(0,-1.5)$);
		
	\end{scope}
\end{tikzpicture}
\vspace*{-1em}
\end{table*}

\begin{figure*}[t!]
\centering
\vspace*{-1em}
\resizebox{\textwidth}{!}{%
	\begin{tikzpicture}[node distance=1.8cm and 2.0cm]
		\tikzset{
			process/.style={rectangle, rounded corners=5pt, draw=black, very thick, minimum height=1cm, minimum width=2.5cm, align=center},
			data/.style={minimum height=1cm, minimum width=2cm, align=center},
			cloud/.style={ellipse, draw=black, very thick, align=center, fill=gray!10},
			arrow/.style={-{Stealth[length=3mm]}, thick},
		}

		\node[minimum width=3cm] (data) {};

		\node[data] (datalabel) at ($(data)+(0,.6cm)$) {\sc Observational Data\\($\DD_{i}$)};

		\node[draw, rectangle, minimum width=2.6cm, minimum height=2.2cm, right=1.5cm of data,rounded corners=2pt] (pagbox) {};
		\node[circle, draw, minimum size=0.5cm] (A) at ([xshift=-0.8cm,yshift=0.6cm]pagbox.center) {A};
		\node[circle, draw, minimum size=0.5cm] (B) at ([xshift=-0.8cm,yshift=-0.6cm]pagbox.center) {B};
		\node[circle, draw, minimum size=0.5cm] (C) at ([xshift=0.8cm,yshift=-0.6cm]pagbox.center) {C};
		\node[circle, draw, minimum size=0.5cm] (D) at ([xshift=0.8cm,yshift=0.6cm]pagbox.center) {D};			
		\draw[-{Circle[open]}, thick] (B) -- (A);
		\draw[{Circle[open]}-{Circle[open]}, thick] (B) -- (C);
		\draw[->, thick] (D) -- (C);

		\node[data, below=.25cm of datalabel] (bprev) {\sc Background Knowledge\\$(\B_{i-1})$};
		\draw[arrow] (bprev.east) -- ($(pagbox.west)!(bprev.east)!(pagbox.west)$);
		\draw[arrow] (datalabel.east) -- ($(pagbox.west)!(datalabel.east)!(pagbox.west)$);

		\node[anchor=north,font=\scriptsize] at (pagbox.south) {As a PAG (see \cref{sec:method_dag_to_pag})};			

		\node[rectangle, minimum width=3.2cm, minimum height=3.2cm, right=1.6cm of pagbox,fill=black!05,rounded corners=2pt] (histogram) {};

		\node[anchor=north east] at (histogram.north east) {\scalebox{.15}{\rotarrow}};

		\foreach \x/\y/\label/\offset in {-0.8/0.6/{\{A,B\}}/0, 0/0/{\{B,C\}}/-0.25, 0.8/-0.6/{\{C,D\}}/0.1} {
			\begin{scope}[shift={(histogram.center)}]
				\draw[fill=black!30] (\x,-0.5+\y) rectangle ++(0.2,0.8);
				\draw[fill=black!50] (\x+0.3,-0.5+\y) rectangle ++(0.2,1.2);
				\draw[densely dashed, red,thick] (\x-0.2,\y+0.7+\offset) -- (\x+0.7,\y+0.7+\offset);
				\node[below] at (\x+0.2,\y-0.6) {\scriptsize \label};
			\end{scope}
		};

		\node[minimum width=3cm,right=2cm of histogram] (outputanchor) {};
		\node[data] (prevHistogram) at ($(outputanchor)+(0,.6cm)$) {\sc Cumulative Histograms\\$(\HE{(i-1)})$};
		\draw[arrow] (prevHistogram.west) -- ($(histogram.east)!(prevHistogram.west)!(histogram.east)$);

		\node[data, below=.25cm of prevHistogram] (output) {\sc \sc Updated Knowledge \\$\{\HE{i}, \B_i\}$};
		\draw[arrow] ($(histogram.east)!(output.west)!(histogram.east)$) -- (output.west);			

		\draw[arrow] (pagbox) -- (histogram);

		\node[above=of histogram, yshift=-1.75cm] (llm) {\sc Language Model};
		\node[anchor=west,font=\scriptsize] at (llm.east) {(with sequential optimization)};

		\draw[arrow,densely dashed,|-{Stealth[length=3mm]}] (output.north) -- ($(prevHistogram.south)+(0,.7em)$);
		\draw[arrow,densely dashed,|-{Stealth[length=3mm]}] (output.south) -- ++(0,-.6cm) -| (bprev.south);
		
	\end{tikzpicture}
}
\caption{\textbf{Overview of the \ours framework:} At each batch $\DD_i$, observational data is combined with accumulated background knowledge $\B_{(i-1)}$ as prior to estimate a PAG structure. A language model is then queried—under sequential optimization—to produce beliefs over possible causal relations and update $\B_{i}$. The updated $\{\HE{i},\B_i\}$ are fed back into the next iteration.}
\label{fig:lencd_overview}
\end{figure*}
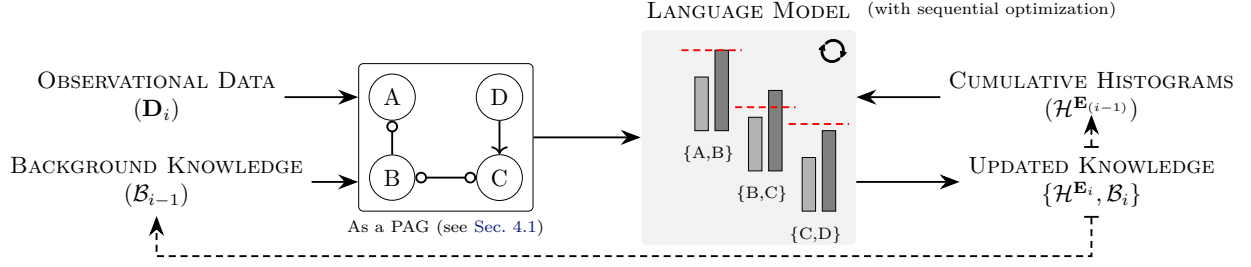

We introduce \ours (Noisy Languge Prior in Sequential Causal Modeling); see \cref{fig:lencd_overview} for an overview of the framework. \ours differs from existing methods that either rely solely on access to the complete observational data or treat LMs as primary discovery mechanism. In a novelty, the causal structure discovery itself is Bayesian-inspired in that the beliefs about causal structure from data are updated iteratively by information drawn from an LM, as data arrive in batches. 
That is, we adopt a \textit{data-first} approach, where for each batch, a traditional causal discovery algorithm, such as FCI from \citet{spirtes1995causal}, constructs an initial PAG conditioned on the background knowledge, which is then iteratively refined through optimized LM queries that leverage the global knowledge while remaining grounded in observed data. To maximize performance under limited budget, LM interactions are framed as a \emph{sequential optimization} problem, selecting the most important edges to query, while accumulating background knowledge over batches. 
Moreover, to complete the causal discovery process, the parameter (edge weights) estimation we propose is \textit{also} Bayesian, which incorporates potentially noisy LM priors on latent confounders and causal relationships. Our method uncovers cross-sectional causal relationships within each batch, rather than modeling temporal dependencies. \ours applies especially to situations where studying contemporaneous causal relations among metrics is separated from confounding temporal effects.\looseness-1

\paragraph{Contributions} We summarize the contributions as follows: \emph{(i)}~We propose a representation shift from DAGs to PAGs, in a hybrid setup of batch, observational data and LM as noisy expert, that inherently captures uncertainty in causal structure learning. \emph{(ii)}~A Bayesian-inspired algorithm for \textit{causal structure} discovery with sequential batch data treating LMs as noisy experts thus accounting for dual sources of bias. \emph{(iii)}~A \textit{sequential optimization} strategy for selecting maximally informative LM edge queries under fixed LM budget constraints. \emph{(iv)}~A Bayesian \textit{parameter estimation} algorithm that robustly integrates noisy LM priors with batched data. Taken together, these constitute ~\ours---an end-to-end causal discovery framework that jointly addresses both \emph{causal structure learning} and \emph{parameter estimation}.

\section{Literature Review} 
\label{sec:literature_review}
\paragraph{Traditional causal discovery}
Traditional causal discovery aims to recover the underlying causal structure from observational data by exploiting statistical dependencies, often formalized through graphical models such as DAGs and PAGs \citep{spirtes2016causal, pearl2009causality, neapolitan2004learning}. These approaches include constraint-based methods (\eg PC from \citet{spirtes2000causation}, FCI \citep{spirtes1995causal, zhang2008completeness}, RFCI from \citet{colombo2012learning}), score-based methods (\eg GES from \citet{chickering2002optimal}), and functional causal models (\eg LiNGAM \citep{shimizu2006linear, shimizu2011directlingam}), typically assuming causal sufficiency and faithfulness \citep{spirtes2000causation, zhang2012strong}. They rely on conditional independence tests or likelihood-based scoring to causal relationships \citep{shah2020hardness, zhang2011kernel, peters2017elements, glymour2019review}. However, these approaches often struggle under data scarcity, presence of latent confounders \citep{spirtes2000causation, monti2020causal}, or domain-specific constraints not captured by statistical patterns \citep{mooij2016distinguishing, peters2014causal}. These limitations are addressed by hybrid methods incorporating external domain knowledge \citep{meek1995strong, heckerman1995learning, pmlrv52ogarrio16} from human experts, refining causal graphs with new variables, modifying edge orientations, or resolving equivalence classes \citep{brouillard2020differentiable, wang2017permutation, constantinou2023impact, ban2023query}. This helps restrict the search space and improves identifiability, particularly when data is limited, as shown by \citet{wallace1996causal}. The evolution from purely statistical methods to knowledge-augmented approaches fuels advanced machine learning techniques to enhance causal discovery \citep{glymour2019review, scholkopf2021toward}. Also, as noted by \citet{baldi2020bayesian}, causal research distinguishes general \textit{vs.}  local settings and applies to diverse fields \citep{andrade2016global,bilal2024macroeconomic,geist2002proximate,kelly2011mitigating,mckinney2016global,mathers2009global}. \looseness-1

Causal discovery extends to streaming data of networks, stock markets, and sensor systems. Causal Bayesian learning and causal discovery with progressively streaming features are well studied \citep{darvish2018causalearn,yu2010causal,you2023local,li2021causality,you2021online,yu2010causal}. Instead, we focus on \textit{contemporaneous} relationships among a fixed set of variables, which yield data across batches. In batch-wise experimentation in medicine and A/B testing, \citet{bridgeford2021batch} examines associations between batches, with batch effects as causal effects, while \citet{zhang2024practical} examines adaptive batch-wise intervention. We confine to sequential, batch-wise observational data.\looseness-1

\paragraph{LM-augmented causal structure discovery} 
LM augmented causal discovery relies on LM's world knowledge and includes pairwise prompting \citep{willig2022can, long2022can, kiciman2023causal, jin2024can, long2023causal}, and triplet-based prompting incorporating voting proposed by \citet{vashishtha2025causal}. Hybrid frameworks integrate LM-generated insights in constraint-based methods or inform score-based approaches \citep{ban2023query, takayama2025integratinglargelanguagemodels}. Reliability of LM-derived constraints is sensitive to domain specificity and prompt framing \citep{kiciman2023causal, ji2023survey}. \citet{ban2023causal} proposes ILS-CSL involving iterative refinement of causal graphs by alternating between LM reasoning and statistical verification. Augmenting LM prompts with correlation matrices or statistical summaries is explored \citep{jiralerspong2024efficientcausalgraphdiscovery, susanti2025can}. Yet, LM-aided causal discovery shows inconsistent judgments across prompting strategies, difficulties with grounding, and biases. Recently, \citet{dasilva2025expertaidedcausaldiscoveryancestral} proposed the BFS method. It is designed for full dataset while our method is anchored in sequential batched data. Additionally, its update is around information gain and does not perform edge-weight estimation.\looseness-1

Deviating from the prior art, we present a Bayesian-inspired framework for causal structure discovery, where data arrive sequentially in batches, and we recognize dual uncertainties---arising from limited observational data and noisy LM responses. We depart from DAG-centric discovery to adopt the PAG, a more robust representation for evolving, uncertain causal structures. We iteratively refine the PAG across batches, while framing LM queries as a sequential optimization problem. 

\paragraph{Parameter estimation in SEMs} 
While existing hybrid methods that combine observational data with language models (LMs) primarily focus on causal structure discovery, they typically stop short of estimating causal effect parameters, which we refer to as \emph{parameter estimation}. In contrast, prior work on parameter estimation is predominantly framed within structural equation models (SEMs) and generally assumes either a known structure or purely data-driven priors.

Within the SEM framework, under assumptions of linearity and Gaussian noise \citep{bollen1989structural, shimizu2006linear}, classical approaches estimate parameters using maximum likelihood or two-stage least squares method, as described in \citet{pearl2009causality}. Learning of SEM parameters extends to nonlinear or nonparametric settings using neural networks or Gaussian processes, enabling flexible modeling of complex dependencies while retaining causal interpretability \citep{zheng2020learning, lachapelle2020gradient}, and helps in integrating expert knowledge with data-driven estimation\citep{peters2017elements, scholkopf2021toward}. However, the use of LM-derived priors and correlations for SEM parameter estimation remains largely unexplored.

To bridge this gap and provide an end-to-end causal discovery framework, we jointly perform causal structure discovery and SEM parameter estimation. The latter constitutes a key contribution of our work: we propose a principled parameter estimation procedure that integrates LM-derived noisy priors into SEMs, yielding consistent estimators of causal strengths despite misspecification of priors.

\section{Problem Setup: Sequential Causal Discovery with LMs}
\label{sec:problem_statement}
Traditional causal discovery methods uncover causal structure by exploiting statistical dependencies in observational data, typically assuming access to the complete dataset, and reliable domain knowledge. In contrast, we focus on the setting of sequential, batch-wise observational data. This setting introduces dual sources of bias: \emph{(i)}~potentially biased and limited batched observational data, and \emph{(ii)}~noisy LM responses. We assume the underlying population distribution $p(X)$ is stationary across batches. However, we allow for selection bias, where each batch may not be drawn randomly from the population distribution. That is, for a batch $i$, $p(X \mid batch = i) \neq p(X)$. Below we introduce the notation and the problem setup.

\paragraph{Problem statement} 
Given sequential batches of observational data subject to selection bias, we consider the problem of: \emph{(i)}~causal discovery in the presence of latent confounders, \emph{(ii)}~incorporating noisy priors provided by LMs, and \emph{(iii)}~parameter estimation of the inferred causal structure. We now define the problem formally.\looseness-1

\paragraph{Notation and setup} 
We define batches of observational data ${\mathcal{D} = \{\DD_i\}_{i=0}^N}$, where each $i$\textsuperscript{th} batch is a sample from the same underlying \emph{`true'} distribution, $\DD_i \sim \mathbb{D}$. Each batch contains same set of observed variables, $\VD$, with $\DD_i \in \R^{n_i \times d}$, where $n_i$ is the number of data points, varies by batch, and $d=|\VD|$ is the number of observed variables. $\R$ denotes real numbers. Any categorical value for an observed variable is encoded as a numerical value. All notations are succinctly shown in \cref{tbl:notations}.

For each batch $\DD_i$, a causal graph ${\GD{i}{=}(\VD, \ED{i})}$ is inferred using a standard causal discovery algorithm, where $\GD{i}$ is a PAG with $\VD$ nodes and $\ED{i}$ edges. 
We assume there exists a true causal graph ${\G{=}(\V, \E)}$, where $\VD {\subseteq} \V$ and ${\V{=}\{\VD, \VL\}}$ represents all the variables, both observed and unobserved latent, and $\E$ represents the true causal relationships. We assume that each confounder may affect two observed variables.

\paragraph{LM-augmented sequential causal discovery} 
Traditional hybrid approaches rely on domain experts to narrow the gap between the inferred causal graph $\GD{i}$ and the true causal graph $\G$ by either introducing knowledge about unobserved variables, effectively reducing the set $\V \setminus \VD$, and adding, removing, reorienting edges in the inferred graph, aiming to minimize the difference $\E \setminus \ED{i}$.

\ours's hybrid approach extends the prior art in two directions: \emph{(i)}~using an LM as a noisy expert to improve the causal structure $\GD{i}$, obtained via known causal discovery algorithm, where \emph{(ii)}~data arrive sequentially in batches. The LM (noisy e\textbf{X}pert), represented $X^i$, helps in reducing the Markov equivalence class to yield causal structure ($\GE{i}$).
Formally,
\begin{align}
f_{\text{CD}}: \DD_i \rightarrow \GD{i} \; ; \; \fllm: \GD{i} \rightarrow \GE{i} \, , 
\end{align}
where ${\GD{i} = (\VD, \ED{i})}$ and $\GE{i} = (\VE{i}, \EE{i})$. The aim is to reduce the discrepancy between the inferred graph and the true causal structure through LM expertise,
\begin{align}	
\G \setminus \GE{i} \leq \G \setminus \GD{i} \; ; \; 
\V \setminus \VE{i} \leq \V \setminus \VD \, \; ; \; 
\E \setminus \EE{i} \leq \E \setminus \ED{i} \, .
\end{align} 

The $\VE{i} \setminus \VD$ is the set of \textit{LM-suggested} variables while $\EE{i} \setminus \ED{i}$ is the set of \textit{LM-suggested} edges. To integrate the LM's noisy responses and address the inherent bias in batch data, our Bayesian-inspired causal discovery framework explicitly handles both \textit{data-induced} and \textit{LM-induced} biases.

\paragraph{Parameter estimation}
Once the LM-augmented causal structure $\GE{i}$ is obtained, we focus on estimating the parameters of the structure equation model (SEM) \ie the edge weights $\vtheta$ and the noise variance $\sigma^2$; $\vphi = \{\vtheta, \sigma^2\}$. A straightforward method to learn the parameters $\vphi$ is Maximum Likelihood Estimation (MLE), $\nabla_{\vphi} \log p(\DD_i \mid \GE{i}, \vphi)$.
However, a critical limitation appears in the presence of unobserved \textit{expert-suggested} variables ($\VL$) in the augmented causal structure, as the likelihood becomes intractable. We address this by proposing a Bayesian parameter estimation algorithm that incorporates the \textit{expert-suggested} information about the unobserved (latent) variable(s), $\VL$.\looseness-1

\begin{figure*}[t!]
\centering

\newcommand{\parallelarrow}[4]{%
	\draw[shorten >=8pt,shorten <=8pt,#4] 
	($(#1)!#3!-90:(#2)$) -- ($(#2)!-#3!-90:(#1)$);
}		

\begin{subfigure}[t]{.48\textwidth}
	\resizebox{\textwidth}{!}{%
		\begin{tikzpicture}[]

			\node (prompt-a) at ($(-3,0)+(-60:1)$) {\texttt{Prompt}};

			\node[circle] (A) at (0, 0) {A};
			\node[circle] (B) at (2, 0) {B};
			\node[circle] (C) at (-60:2) {C}; %

			\parallelarrow{A}{B}{2pt}{->} 
			\parallelarrow{B}{A}{2pt}{->} 

			\parallelarrow{B}{C}{2pt}{->} 
			\parallelarrow{C}{B}{2pt}{<-} 

			\parallelarrow{C}{A}{2pt}{<-} 
			\parallelarrow{A}{C}{2pt}{->} 

			\node[draw, rectangle, 
			fit=(A) (B) (C),   %
			inner sep=5pt,     %
			rounded corners   %
			] (llm-a) {};  
			\node[inner sep=2pt,font=\small\strut~,anchor=south west] at (llm-a.south west) {LM};

			\node[circle] (A) at (4, 0) {A};
			\node[circle] (B) at ($(A)+(2,0)$) {B};
			\node[circle] (C) at ($(A)+(-60:2)$) {C}; %

			\draw[shorten >=2pt,shorten <=2pt,->] (A) -- (C);
			\draw[shorten >=2pt,shorten <=2pt,->] (B) -- (C);

			\node[draw, rectangle, 
			fit=(A) (B) (C),   %
			inner sep=5pt,     %
			rounded corners   %
			] (dag) {};
			\node[inner sep=2pt,font=\small\strut~,anchor=south west] at (dag.south west) {DAG};

			\draw[line width=2pt,black,-latex,shorten >=3pt,shorten <=3pt] (prompt-a) -- node[above]{\resizebox{2em}{!}{\rotarrow}} (llm-a);
			\draw[line width=2pt,black,-latex,shorten >=3pt,shorten <=3pt] (llm-a) -- (dag);
			
	\end{tikzpicture}}
	\caption{Conventional DAG-LM}
	\label{fig:dag-llm}
\end{subfigure}
\hfill
\begin{subfigure}[t]{.48\textwidth}
	\resizebox{\textwidth}{!}{%
		\begin{tikzpicture}[]

			\node (prompt-a) at ($(-3,0)+(-60:1)$) {\texttt{Prompt}};

			\node[circle] (A) at (0, 0) {A};
			\node[circle] (B) at (2, 0) {B};
			\node[circle] (C) at (-60:2) {C}; %

			\parallelarrow{A}{B}{2pt}{{Circle[open,length=4pt]}->} 
			\parallelarrow{B}{A}{2pt}{{Circle[open,length=4pt]}-{Circle[open,length=4pt]}} 

			\parallelarrow{B}{C}{2pt}{->} 
			\parallelarrow{C}{B}{2pt}{<-} 

			\parallelarrow{C}{A}{2pt}{<-} 
			\parallelarrow{A}{C}{2pt}{->}

			\node[draw, rectangle, 
			fit=(A) (B) (C),   %
			inner sep=5pt,     %
			rounded corners   %
			] (llm-a) {};  
			\node[inner sep=2pt,font=\small\strut~,anchor=south west] at (llm-a.south west) {LM};

			\node[circle] (A) at (4, 0) {A};
			\node[circle] (B) at ($(A)+(2,0)$) {B};
			\node[circle] (C) at ($(A)+(-60:2)$) {C}; %

			\draw[shorten >=2pt,shorten <=2pt,->] (A) -- (C);
			\draw[shorten >=2pt,shorten <=2pt,->] (B) -- (C);  
			\draw[shorten >=2pt,shorten <=2pt,
			{Circle[open,length=4pt]}-{Circle[open,length=4pt]}] (A) -- (B);  

			\node[draw, rectangle, 
			fit=(A) (B) (C),   %
			inner sep=5pt,     %
			rounded corners   %
			] (pag) {};  
			\node[inner sep=2pt,font=\small\strut~,anchor=south west] at (pag.south west) {PAG};

			\draw[line width=2pt,black,-latex,shorten >=3pt,shorten <=3pt] (prompt-a) -- node[above]{\resizebox{2em}{!}{\rotarrow}} (llm-a);
			\draw[line width=2pt,black,-latex,shorten >=3pt,shorten <=3pt] (llm-a) -- (pag);
			
	\end{tikzpicture}}
	\caption{PAG-LM in \ours}
	\label{fig:pag-llm}
\end{subfigure}
\caption{\textbf{PAG-LM} streamlines how LMs compose the graph and allows for ambiguities to be indicated in the structure. \emph{(a)}~DAG is constructed by iterative prompting 
	(\raisebox{-2pt}{\protect\resizebox{1em}{!}{\protect\rotarrow}}) leading to ambiguities (\eg \hspace*{-.8em}\protect\tikz[baseline=-.8ex,inner sep=0]{\protect\node[] (a) {};\protect\node[right of=a] (b){}; \protect\parallelarrow{a}{b}{2pt}{->}\protect\parallelarrow{b}{a}{2pt}{->}}\hspace*{-.8em}) requiring heuristics 
	that cannot be represented. \emph{(b)}~\ours represents the causal structure as a PAG that implicitly allow ambiguities to be represented providing a richer representation ( \eg $\dotarrowdot$)\looseness-1}
\label{fig:pag_llm_illustration}
\vspace*{-.5em}
\end{figure*}

\section{Sequential Causal Structure Discovery with PAGs}

The promise of LMs as proxies for domain expert in causal structure discovery, as studied in prior art, typically queried via pairwise or triplet prompts, faces key limitations: \emph{(i)}~LMs may provide responses regardless of causal relations among other variables, resulting in inconsistent or cyclic causal graphs; \emph{(ii)}~LMs may hallucinate; \emph{(iii)}~LMs may be overly optimistic, predicting spurious causal relationships with high recall but low precision (\cf \cref{tbl:llm_optimistic_nature}). 
While prior art addresses \emph{(i)} and \emph{(ii)} through heuristics or auxiliary models to enforce acyclicity and consistency, they add complexity and potentially degrading performance. Crucially, \textit{(ii)} and \textit{(iii)} remain largely unexplored in the setting of sequential batch data where two distinct sources of bias emerge: \textit{LM-induced} bias and \textit{data-induced} bias.

\subsection{PAG to Incorporate Uncertainty}
\label{sec:method_dag_to_pag}
\begin{wraptable}{r}{0.45\textwidth}
	\raggedleft\scriptsize
	\vspace{-2em}
	\caption{\textbf{DAG to PAG:} Structural Intervention Distance (SID) between DAG-Pairwise (Voting) and PAG-Pairwise depicts benefit of PAG to represent inherent causal uncertainty.\looseness-1}
	\label{tbl:pag_llm}
	\setlength{\tabcolsep}{3pt}
\begin{tabular}{cclc}
	\toprule
	\textbf{Dataset} & \textbf{Temp.} & \textbf{Method} & \textbf{SID} $\downarrow$ \\
	\midrule
	\multirow{6}{*}{\rotatebox{90}{\sc Earthquake}} 
	& \multirow{2}{*}{0.0}   & Pairwise (Voting) & \val{(1.0, 1.0)}{(0.0, 0.0)}   \\
	 & & \cellcolor{highlight}{PAG-Pairwise}  &\cellcolor{highlight}{\val{(0.0, 0.0)}{(0.0, 0.0)}} \\
	\cdashline{2-4}[.4pt/2pt]
	& \multirow{2}{*}{0.5} & Pairwise (Voting)  & \val{(1.4, 1.4)}{(1.2, 1.2)}  \\
	 & & \cellcolor{highlight}{PAG-Pairwise}   & \cellcolor{highlight}{\val{(0.0, 0.0)}{(0.0, 0.0)}} \\
	\cdashline{2-4}[.4pt/2pt]
	& \multirow{2}{*}{1.0} & Pairwise (Voting)  & \val{(1.6, 1.6)}{(1.5, 1.5)} \\
	 &  & \cellcolor{highlight}{PAG-Pairwise}  & \cellcolor{highlight}{\val{(0.8, 0.8)}{(1.6, 1.6)}} \\
	\midrule
	\multirow{6}{*}{\rotatebox{90}{\sc Asia}} 
	& \multirow{2}{*}{0.0}   & Pairwise (Voting) &\val{(6.4, 6.4)}{(0.8, 0.8)} \\
	& & \cellcolor{highlight}{PAG-Pairwise}  & \cellcolor{highlight}{\val{(3.8, 3.8)}{(0.9, 0.9)}} \\
	\cdashline{2-4}[.4pt/2pt]
	& \multirow{2}{*}{0.5} & Pairwise (Voting) &\val{(5.2, 5.2)}{(1.7, 1.7)} \\
	 & & \cellcolor{highlight}{PAG-Pairwise} & \cellcolor{highlight}{\val{(3.5, 3.5)}{(0.8, 0.8)}} \\
	\cdashline{2-4}[.4pt/2pt]
	& \multirow{2}{*}{1.0}   & Pairwise (Voting) &\val{(5.4, 5.4)}{(1.8, 1.8)} \\
	&  & \cellcolor{highlight}{PAG-Pairwise} & \cellcolor{highlight}{\val{(4.0, 4.0)}{(1.9, 1.9)}}  \\
	\bottomrule
\end{tabular}
 	\vspace{-2em}
\end{wraptable}
Prompting methods (pairwise or triplet) constrain the response format to---\emph{`causal', `non-causal',} or \emph{`unknown'}---which prevent LMs from expressing uncertainty, thus exacerbating bias and inconsistency. To address this, we propose a representational shift from DAG to PAG when using LMs as proxy for experts. PAGs encode uncertainty and structural ambiguity in a principled manner, accommodating latent confounding and partial orientation (\cf \cref{fig:pag_llm_illustration}). Formally, we expand the limited edge set of DAG, ${\E^{\text{DAG}}{=} \{\rightarrow, \leftarrow, \cdot\}}$, to the richer set for PAG, ${\E^{\text{PAG}}{=} \{\rightarrow, \leftarrow, \leftrightarrow, \dotarrow, \arrowdot , \dotarrowdot, \cdot, -\}}$
\footnote{To capture selection bias, ambiguous edges $(\dotarrowdot, \dotarrow)$ and undirected edges `$-$' can be used. Not all causal discovery algorithms output undirected edges (\eg \url{https://causal-learn.readthedocs.io/en/latest/search_methods_index/Constraint-based causal discovery methods/FCI.html}), but they output ambiguous edges, which we use as proxy for selection bias.}, where $\cdot$ represents no causal relation. The expansion allows LM to select from more options and improve causal discovery (\cf \cref{tbl:pag_llm}). In PAG-pairwise, the LM is queried per variable pair to select from $\E^{\text{PAG}}$ (see \cref{app:prompts} for the full prompt).

The representational shift to PAG is a necessary pivot and starting point for addressing dual sources of bias---\textit{LM-induced} and \textit{data-induced}. The LM's error prone response constitutes a prior and motivates a novel Bayesian causal structure discovery framework in the sequential batch setting, whereby we integrate causal predictions from observational data and LMs in a sequential, iterative manner. 
Building on this key observation, we introduce the causal structure learning algorithm of \ours and then on the inferred structure introduce the parameter estimation algorithm.

\subsection{LM-augmented Causal Structure Discovery}
\label{sec:method_lm_causal_discovery}
As a principled way to incorporate the dual sources of bias, we adopt a Bayesian-inspired formulation. Formally, given data batch $\DD_i$, cumulative background knowledge $\B_{(i-1)}$ up to batch $(i-1)$, the posterior over $\G_i$ is obtained via Bayes' rules, 
\begin{equation}
\underbrace{p( \G_i \mid \DD_i , \B_{(i-1)})}_{\text{Posterior}} \propto \underbrace{p(\DD_i \mid \G_i)}_{\text{Likelihood}} \, \underbrace{p(\G_i \mid \B_{(i-1)})}_{\text{Prior}} \, .
\end{equation}
The prior $p(\G_{i} \mid \B_{(i-1)})$ is iteratively shaped by background knowledge, which accumulates across batches. Intuitively, the prior encodes the edge type between node pairs in the causal structure. Once the posterior is obtained, an LM is queried to update and obtain background knowledge $\B_{i}$ using \cref{eq:dynamic_bg} and \cref{eq:score}, which shapes the prior for the next batch. We note that this formulation serves as a conceptual motivation for the design of \ours; the implementation maintains uncertainty at the edge level (\cref{eq:hist_update,eq:dynamic_bg}) rather than computing a full posterior over graph space. Next we discuss the formulation.

Given the inherent stochasticity of LMs, its response can be viewed as a \textit{sample} from an implicit distribution over the edge types ($\E^{\text{PAG}}$). This allows explicit modeling of uncertainty in LM responses, ${\fllm^{(i)}(A, B, \text{Pr}) \sim p(E_{AB} \mid A, B)}$, where $A, B$ are the two nodes, Pr is the prompt, and $E_{AB} \in \E^{\text{PAG}}$.
Treating LM responses as noisy observations rather than ground truth, addresses the challenge of LM hallucination. As more batches are processed, the prompt $\text{Pr}$ becomes more informative, thereby decreasing uncertainty in LM responses.
Formally, we treat LM as a \textit{black-box causal edge sampler} and aggregate multiple LM samples into empirical histograms that are updated iteratively over batches,\looseness-1
\begin{equation}
	\label{eq:hist_update}
\HE{i}(A, B)[E_{AB}] = \HE{{i-1}}(A, B)[E_{AB}] + \mathbb{I}[f_{\text{LM}}^{(i)}(A, B) = E_{AB}] \, ,
\end{equation}
where $\HE{i}(A, B)$ represents the cumulative histogram up to batch $i$, and $E_{AB}$ represents a type of causal relation. These histograms define an approximate posterior distribution over edge types, capturing the LM's evolving beliefs about causal relationships. 
\begin{table*}[t!]
	\centering \scriptsize
	\setlength{\tabcolsep}{3.5pt}
	\caption{\textbf{Semantic Entropy in \ours.} We report the mean Shannon entropy of empirical histograms over PAG edge predictions (mean $\pm$ std over 5 runs) across sequential batches for {\sc User Level Data–I} and {\sc User Level Data–II}. The observed reduction in entropy indicates decreasing predictive uncertainty over PAG edge types as structural context is progressively incorporated.}
	\label{tbl:ambiguity_compression}
	\begin{tabular}{lccccccc}
		\toprule
		\textbf{Dataset} & \textbf{Batch-1} & \textbf{Batch-2} & \textbf{Batch-3} & \textbf{Batch-4}& \textbf{Batch-5}& \textbf{Batch-6}& \textbf{Batch-7}\\
		\midrule
		\sc{User Level Data - I} & $0.48 \pm 0.06$ & $0.50 \pm 0.05$ & $0.36 \pm 0.08$ & $0.17 \pm 0.13$ & $0.10 \pm 0.06$ & $0.12 \pm 0.06$ & $0.09 \pm 0.05$ \\
		\midrule
		\sc{User Level Data - II} & $0.31 \pm 0.25$ & $0.27 \pm 0.22$ & $0.18 \pm 0.18$ & $0.12 \pm 0.19$ & $0.07 \pm 0.15$ & $0.00 \pm 0.00$ & $0.00 \pm 0.00$ \\
		\bottomrule
	\end{tabular}
	\vspace*{-1.1em}
\end{table*}

To determine when accumulated LM evidence is sufficient to promote an edge to background knowledge, we define a dynamic threshold that balances distributional uncertainty and sampling uncertainty (motivated by the explore-exploit trade-off formalized in \cref{sec:method_lm_interaction}),
\begin{align}
\label{eq:dynamic_bg}
\tau_i^e = \alpha \times E_i^e \times T_i^e  + (1 - \alpha) \sqrt{T_i^e \left(1 - \frac{T_i^e}{T_i}\right)}, \quad \text{s.t.} \quad %
E_i^e = -\sum_j \frac{\HE{i}_{j,e}}{T_i^e} \times \log \left(\frac{\HE{i}_{j,e}}{T_i^e}\right) .
\end{align}
Here, for batch $i$ and edge $e$, $\tau_i^e$ denotes the threshold, $E_i^e$ the posterior entropy, $T_i^e$ the number of LM interactions, $T_i{=}\sum_e T_i^e$ the total interactions, and $\HE{i}_{j,e}$ the frequency of bin $j$ in $e$’s histogram.

Intuitively, the first term in $\tau_i^e$ of \cref{eq:dynamic_bg} accounts for uncertainty in the histogram edge distribution, while the second term handles the sampling uncertainty that decreases as more batches of data arrive. The hyperparameter $\alpha$ balances between these two terms. \cref{fig:structure_learning_ablation} showcases the efficiency of the proposed dynamic background threshold (\cf \cref{eq:dynamic_bg}), with the additional details discussed in \cref{sec:exp_structure_learning}.
The pseudo-code of the algorithm is outlined in \cref{alg:proposed-structure-learning}.

A key property of our sequential framework is that the prompt $\text{Pr}$ becomes progressively more informative across batches as $\GE{i}$ expands, providing increasingly rich structural context to the LM. This corresponds to conditioning on accumulated causal constraints, and therefore the model’s predictive uncertainty is expected to decrease over iterations. We empirically validate this behavior by reporting the mean histogram entropy across batches for the two real world datasets {\sc User Level Data–I} and {\sc User Level Data–II} in \cref{tbl:ambiguity_compression}, and observe a consistent reduction as the graph is incrementally refined.

From an uncertainty modeling perspective, our histogram-based formulation is closely related to recent work on semantic entropy in language models \citep{Farquhar2024LLMSemanticEntropy, nikitin2024kernel, kuhn2023semantic}. However, in our setting, \ours restricts the LM output space to the finite set of PAG edge types $\E^{\text{PAG}}$, yielding purely categorical predictions. Consequently, semantic entropy can be computed directly as the Shannon entropy of empirical histograms over sampled edge predictions, without requiring embedding-based clustering. While conceptually aligned with prior formulations of semantic entropy, this arises naturally in our setting from the discrete decision space induced by PAG edge selection. 

\begin{figure}[t!]
	\centering
	\vspace*{-6pt}
	\begin{minipage}[t]{.48\textwidth}
        {
        \let\AND\relax
		\begin{algorithm}[H]
			\caption{LM-augmented structure learning}
			\label{alg:proposed-structure-learning}
			\begin{algorithmic}[1]
				\REQUIRE $\DD_i, \HE{(i{-}1)}, \HL{(i{-}1)}, \B_{(i{-}1)}, \me, \ml$,
				\ENSURE  $\HE{i}, \HL{i}, \B_i$
				
				\STATE \textbf{Initial causal structure}
				\STATE $f_{\text{CD}}: \DD_i \times \B_{(i-1)} \rightarrow \GD{i}$
				
				\STATE \textbf{Expert-guided causal structure refinement}
				\STATE $\fllm: \GD{i} {\times} \HE{(i{-}1)} {\times} \B_{(i{-}1)} {\times} \me {\rightarrow} (\HE{i}, \B_i)$
				
				\STATE \textbf{Expert-suggested latent confounder}
				\FOR{$A \leftrightarrow B$ in $\B_i$}
				\STATE $\fllm: \HE{i} {\times} \HL{(i{-}1)} {\times} A {\times} B {\times} \ml {\rightarrow} \HL{i}$
				\ENDFOR
				
				\STATE \textbf{return} $\HE{i}, \HL{i},  \B_i$
				
			\end{algorithmic}
		\end{algorithm}
        }
		\vspace{-1em}
		\textbf{Notation:}\\
		$\me, \ml$: LM budget for edges and confounder\\
		$\HE{}, \HL{}$: histograms for edges and confounders\\
		$\IP, \IR$: Prompt for prior and corelation
		\end{minipage}
	\hfill
	\begin{minipage}[t]{.48\textwidth}
        {
        \let\AND\relax
		\begin{algorithm}[H]
			\caption{Bayesian Parameter Estimation}
			\label{alg:llm-causal-discovery-parameter-estimation}
			\begin{algorithmic}[1]
				\REQUIRE $\DD_i$, $\eta$, $\GE{i}, \IP, \IR$
				\ENSURE $\vphi$
				
				\STATE \textbf{Initialization}
				
				\STATE Warm-start  for $\vphi^{\text{O}}$\\
				$\quad \vphi^{\text{O}} \in \arg\max_{\vphi^{\text{O}}} p(\DD_i \mid \GE{i}_{-\VL}, \vphi^{\text{O}})$
				
				\STATE Get prior over $\VL$ \\
				$\quad \fllm: \GE{i} \times \VL \times \IP \rightarrow \N(\vm_p, \MS_p)$
				
				\STATE Get correlation $\rho(\VL, \VD)$ \\
				$\quad \fllm: \GE{i} \times \IR \rightarrow \rho(\VL, \VD)$
				
				\STATE Initialize $\vthetaL$ to $\rho(\VL, \VD)$ or randomly
				
				\STATE \textbf{Iterative Optimization}
				\WHILE{not converged}
				\STATE Compute posterior over $\VL$ using \cref{eq:e_step}
				
				\STATE Optimize $\vphi$ variables using \cref{eq:m_step}
				\ENDWHILE
				
				\STATE \textbf{return} $\vphi$
			\end{algorithmic}
		\end{algorithm}
        }
	\end{minipage}
\end{figure}
\subsection{LM Interaction: Sequential Optimization}  
\label{sec:method_lm_interaction}
Given the stochastic LM responses and cumulative histogram-based estimates of edge uncertainty, we next address how to allocate LM queries efficiently across candidate edges. We model LM interactions $\fllm$ as a \textit{sequential optimization} problem under a limited budget. At each batch $i$, up to $\ml$ calls to the LM are allowed. The objective is to strategically allocate these calls to refine the edge distribution $\HE{i}$ and expand the set of background knowledge $\B_i$.\looseness-1

In the edge refinement setting, each query corresponds to selecting a candidate edge $e$ and querying $\fllm$ to reduce uncertainty about its type. The LM's response is treated as a noisy sample from the underlying distribution over edge types. This induces a natural trade-off: we must \textit{explore} uncertain edges to improve estimates and \textit{exploit} promising edges that are likely to yield useful and increasing background knowledge.
Formally, we cast LM interactions as a sequential decision-making problem:
\begin{align}
\label{eq:seqential_decisions}
\textbf{Arms:} \quad & \mathcal{A} = \{ \text{All possible edges between variables, } \E^{\text{PAG}} \},  \\
\textbf{Reward:} \quad & r_k(e) = \text{Information gain from querying edge } e \text{ at step } k, \\
\textbf{Policy:} \quad & \pi: \HE{i} \times \GD{i} \rightarrow e \quad \text{(Edge selection rule)}.
\end{align}

The optimization objective is to maximize cumulative information gain over $\ml$ LM calls, balancing both the expansion of background knowledge, and the uncertainty reduction in $\HE{i}$.
\begin{table*}[t!]
\centering \scriptsize
\caption{\textbf{\ours improves causal discovery:} We experiment with six datasets--number of observed variables ranges 5 (small) to 37 (large)-- using two paradigms: \textit{Only-Data} and \textit{Data-LM}. We evaluate with 5 metrics: \textit{Modified SHD, SID, Precision, Recall, F1}. All methods use GPT-3.5\textsubscript{turbo} as an LM with temperature $1$. We report mean and standard deviation over $5$ runs and perform significance test with $\alpha=0.05$.} 
\label{tbl:causal_discovery_metrics}
\setlength{\tabcolsep}{3.5pt}
\resizebox{\textwidth}{!}{
\begin{tabular}{cclccccc}
	\toprule
	\textbf{Dataset} & \textbf{Approach} & \textbf{Method} & \textbf{Mod. SHD $\downarrow$} & \textbf{SID $\downarrow$} & \textbf{Precision $\uparrow$} & \textbf{Recall $\uparrow$} & \textbf{F1 Score $\uparrow$} \\
	\midrule
\multirow{7}{*}{\rotatebox{90}{\dataset{Earthquake}{5}}} & \multirow{4}{*}{Only-Data} & FCI-Cumulative & \val{2.00}{0.00} & \val{(0.00, 5.00)}{(0.00, 0.00)} & \valb{1.00}{0.00} & \val{0.50}{0.00} & \val{0.67}{0.00} \\
&& FCI-Vanilla & \val{3.60}{0.80} & \val{(8.20, 9.20)}{(3.60, 1.60)} & \val{0.20}{0.40} & \val{0.05}{0.10} & \val{0.08}{0.16} \\
&& FCI-Iterative & \val{5.00}{1.67} & \val{(12.20, 12.20)}{(4.66, 4.66)} & \val{0.30}{0.27} & \val{0.20}{0.19} & \val{0.24}{0.22} \\
&& FCI-Heuristics & \val{3.60}{0.80} & \val{(8.20, 9.20)}{(3.60, 1.60)} & \val{0.20}{0.40} & \val{0.05}{0.10} & \val{0.08}{0.16} \\
\cdashline{2-8}[.4pt/1pt]
& \multirow{3}{*}{Data-LM} & LLM-first & \val{6.00}{0.82} & \val{(15.00, 15.00)}{(0.82, 0.82)} & \val{0.11}{0.16} & \val{0.08}{0.12} & \val{0.09}{0.13} \\
&& ILS-CSL & \val{2.38}{0.96} & \val{(5.25, 6.50)}{(0.83, 2.69)} & \valb{0.88}{0.22} & \val{0.44}{0.21} & \val{0.56}{0.21} \\
\rowcolor{highlight}\cellcolor{white} &\cellcolor{white}& \ours & \valb{1.00}{0.63} & \valb{(2.20, 2.20)}{(1.60, 1.60)} & \valb{1.00}{0.00} & \valb{0.75}{0.16} & \valb{0.85}{0.11} \\
	\midrule
\multirow{7}{*}{\rotatebox{90}{\dataset{Asia}{8}}} & \multirow{4}{*}{Only-Data} & FCI-Cumulative & \val{7.00}{0.00} & \val{(23.00, 49.00)}{(0.00, 0.00)} & \val{0.00}{0.00} & \val{0.00}{0.00} & \val{0.00}{0.00} \\
&& FCI-Vanilla & \val{7.80}{0.75} & \val{(30.00, 35.00)}{(5.90, 2.45)} & \val{0.00}{0.00} & \val{0.00}{0.00} & \val{0.00}{0.00} \\
&& FCI-Iterative & \val{8.00}{1.26} & \val{(33.00, 33.00)}{(7.46, 7.46)} & \val{0.45}{0.24} & \val{0.23}{0.15} & \val{0.29}{0.18} \\
&& FCI-Heuristics & \val{7.80}{0.75} & \val{(30.00, 35.00)}{(5.90, 2.45)} & \val{0.00}{0.00} & \val{0.00}{0.00} & \val{0.00}{0.00} \\
\cdashline{2-8}[.4pt/1pt]
& \multirow{3}{*}{Data-LM} & LLM-first & \val{7.33}{0.94} & \val{(27.67, 27.67)}{(2.49, 2.49)} & \val{0.58}{0.12} & \val{0.29}{0.06} & \val{0.39}{0.08} \\
&& ILS-CSL & \val{6.50}{0.50} & \val{(28.50, 28.50)}{(3.20, 3.20)} & \valb{0.79}{0.12} & \val{0.28}{0.11} & \val{0.40}{0.12} \\
\rowcolor{highlight}\cellcolor{white} &\cellcolor{white}& \ours & \valb{4.60}{1.02} & \valb{(13.60, 13.60)}{(3.83, 3.83)} & \valb{0.80}{0.12} & \valb{0.60}{0.12} & \valb{0.67}{0.08} \\
	\midrule
	\multirow{7}{*}{\rotatebox{90}{\tiny\dataset{User Level Data-I}{9}}} & \multirow{4}{*}{Only-Data} & FCI-Cumulative & \val{15.00}{2.77} & \val{(47.80, 47.80)}{(4.62, 4.62)} & \val{0.72}{0.18} & \val{0.37}{0.09} & \val{0.47}{0.09} \\
	&& FCI-Vanilla & \val{21.30}{3.50} & \val{(61.60, 61.60)}{(5.54, 5.54)} & \val{0.41}{0.18} & \val{0.22}{0.10} & \val{0.29}{0.13} \\
	&& FCI-Iterative & \valb{\phantom{0}5.60}{1.20} & \val{(23.40, 23.40)}{(7.94, 7.94)} & \valb{0.93}{0.04} & \val{0.77}{0.02} & \valb{0.84}{0.03} \\
	&& FCI-Heuristics & \val{21.30}{3.50} & \val{(61.60, 61.60)}{(5.54, 5.54)} & \val{0.41}{0.18} & \val{0.22}{0.10} & \val{0.29}{0.13} \\
	\cdashline{2-8}[.4pt/1pt]
	& \multirow{3}{*}{Data-LM} & LLM-first & \val{\phantom{0}9.68}{1.67} & \val{(38.12, 38.12)}{(4.06, 4.06)} & \val{0.80}{0.06} & \val{0.66}{0.04} & \val{0.72}{0.05} \\
	&& ILS-CSL & \val{\phantom{0}9.20}{1.72} & \val{(34.20, 34.20)}{(3.49, 3.49)} & \val{0.82}{0.05} & \val{0.66}{0.05} & \val{0.73}{0.05} \\
	 &  & \cellcolor{highlight} \ours & \cellcolor{highlight} \valb{\phantom{0}5.00}{0.71} & \cellcolor{highlight} \valb{(13.75, 13.75)}{(2.49, 2.49)} & \cellcolor{highlight} \valb{0.90}{0.04} & \cellcolor{highlight} \valb{0.83}{0.02} & \cellcolor{highlight} \valb{0.86}{0.02} \\
	\midrule
\multirow{7}{*}{\rotatebox{90}{\tiny\dataset{User Level Data-II}{8}}} & \multirow{4}{*}{Only-Data} & FCI-Cumulative & \val{19.80}{2.04} & \valb{(40.00, 40.00)}{(3.03, 3.03)} & \val{0.15}{0.07} & \val{0.10}{0.04} & \val{0.12}{0.05} \\
&& FCI-Vanilla & \val{17.40}{1.83} & \valb{(36.80, 39.40)}{(2.04, 3.38)} & \val{0.07}{0.13} & \val{0.02}{0.03} & \val{0.03}{0.05} \\
&& FCI-Iterative & \val{20.60}{1.77} & \val{(42.80, 43.40)}{(4.07, 4.22)} & \val{0.06}{0.08} & \val{0.05}{0.07} & \val{0.05}{0.07} \\
&& FCI-Heuristics & \val{17.40}{1.83} & \valb{(36.80, 39.40)}{(2.04, 3.38)} & \val{0.07}{0.13} & \val{0.02}{0.03} & \val{0.03}{0.05} \\
\cdashline{2-8}[.4pt/1pt]
& \multirow{3}{*}{Data-LM} & LLM-first & \val{18.75}{0.43} & \val{(44.00, 44.00)}{(1.00, 1.00)} & \val{0.18}{0.01} & \valb{0.17}{0.00} & \valb{0.18}{0.00} \\
&& ILS-CSL & \valb{16.90}{1.50} & \val{(44.40, 47.80)}{(6.05, 3.49)} & \val{0.16}{0.03} & \val{0.10}{0.04} & \val{0.12}{0.03} \\
 & & \cellcolor{highlight} \ours & \cellcolor{highlight} \valb{16.33}{1.80} & \cellcolor{highlight} \valb{(40.17, 40.17)}{(3.14, 3.14)} & \cellcolor{highlight} \valb{0.26}{0.04} & \cellcolor{highlight} \valb{0.18}{0.06} & \cellcolor{highlight} \valb{0.20}{0.04} \\
 \midrule
\multirow{7}{*}{\rotatebox{90}{\dataset{Child}{19}}} & \multirow{4}{*}{Only-Data} & FCI-Cumulative & \val{27.50}{0.00} & \valb{(111.00, 131.00)}{(0.00, 0.00)} & \val{0.38}{0.00} & \val{0.36}{0.00} & \val{0.37}{0.00} \\
&& FCI-Vanilla & \val{28.00}{1.48} & \val{(129.20, 133.20)}{(10.46, 10.76)} & \val{0.38}{0.04} & \val{0.26}{0.05} & \val{0.31}{0.04} \\
&& FCI-Iterative & \val{32.10}{1.16} & \val{(149.00, 164.40)}{(7.16, 10.33)} & \val{0.27}{0.03} & \val{0.26}{0.04} & \val{0.26}{0.03} \\
&& FCI-Heuristics & \val{28.00}{1.48} & \val{(129.20, 133.20)}{(10.46, 10.76)} & \val{0.38}{0.04} & \val{0.26}{0.05} & \val{0.31}{0.04} \\
\cdashline{2-8}[.4pt/1pt]
& \multirow{3}{*}{Data - LLM} & LLM-first & \val{31.67}{2.05} & \val{(172.00, 172.00)}{(13.42, 13.42)} & \val{0.29}{0.05} & \val{0.30}{0.05} & \val{0.30}{0.05} \\
&& ILS-CSL  & \val{32.00}{2.10} & \val{(154.00, 154.00)}{(26.53, 26.53)} & \val{0.28}{0.05} & \val{0.28}{0.05} & \val{0.28}{0.05} \\
\rowcolor{highlight}\cellcolor{white} &\cellcolor{white}& \ours & \valb{25.50}{0.89} & \valb{(103.40, 112.20)}{(8.09, 7.05)} & \valb{0.43}{0.01} & \valb{0.44}{0.02} & \valb{0.43}{0.01} \\
\midrule
\multirow{7}{*}{\rotatebox{90}{\dataset{Alarm}{37}}} & \multirow{4}{*}{Only-Data} & FCI-Cumulative & \valb{45.00}{0.00} & \valb{(626.00, 626.00)}{(\phantom{0}0.00, \phantom{0}0.00)} & \val{0.25}{0.00} & \val{0.02}{0.00} & \val{0.04}{0.00} \\
&& FCI-Vanilla & \val{49.50}{1.61} & \val{(617.80, 699.20)}{(29.23, 68.43)} & \val{0.00}{0.00} & \val{0.00}{0.00} & \val{0.00}{0.00} \\
&& FCI-Iterative & \val{52.40}{6.21} & \valb{(612.20, 636.80)}{(49.87, 43.83)} & \valb{0.33}{0.14} & \val{0.12}{0.05} & \val{0.17}{0.07} \\
&& FCI-Heuristics & \val{49.50}{1.61} & \val{(617.80, 699.20)}{(29.23, 68.43)} & \val{0.00}{0.00} & \val{0.00}{0.00} & \val{0.00}{0.00} \\
\cdashline{2-8}[.4pt/1pt]
& \multirow{3}{*}{Data - LLM} & LLM-first & \val{52.33}{3.09} & \val{(673.33, 673.33)}{(25.63, 25.63)} & \valb{0.33}{0.08} & \val{0.13}{0.02} & \val{0.19}{0.03} \\
&& ILS-CSL  & \val{51.20}{1.60} & \val{(657.20, 657.20)}{(24.07, 24.07)} & \val{0.34}{0.05} & \val{0.10}{0.01} & \val{0.15}{0.02} \\
\rowcolor{highlight}\cellcolor{white} &\cellcolor{white}& \ours & \val{50.90}{1.32} & \valb{(589.80, 591.00)}{(26.48, 25.59)} & \valb{0.42}{0.03} & \valb{0.22}{0.02} & \valb{0.29}{0.02} \\
\bottomrule
\end{tabular} }
\end{table*}
To implement $\pi$, since the true information gain $r_k(e)$ is not known before querying, we propose a scoring function that serves as a proxy for the expected reward, jointly accounting for epistemic uncertainty, proximity to background knowledge thresholds, and exploration,
\begin{align}
S_i^e = w_1 E_i^e + w_2 \left(\frac{1}{TD_i^e}\right) + w_3 \sqrt{\frac{\log T_i}{T_i^e}}, \quad \text{s.t.} \quad TD_i^e = \tau_i - \max(\HE{i}(e)) \, ,
\label{eq:score}
\end{align}
where $TD_i^e$ is the threshold distance from being included in background knowledge, $E_i^e, T_i, T_i^e$ are as defined in \cref{eq:dynamic_bg}, and $w_1, w_2, w_3$ are hyper-parameters controlling the trade-off. After querying, the LM's response updates the histogram $\HE{i}(e)$, which constitutes the realized information gain. At each step, the edge ${e^* = \arg\max_e S_i^e}$ is selected for LM interaction. \cref{fig:structure_learning_ablation} showcases the effectiveness of the proposed selection score (\cf \cref{eq:score}), with details discussed in \cref{sec:exp_structure_learning}. This formulation generalizes to other expert-guided tasks (\eg confounder detection) by redefining the arms and reward function. 

While the connection to multi-armed bandits and the derivation of regret bounds are appealing in this setup, LM's stochastic nature, inter-dependent arms, and implicit priors induced by causal structure constraints warrant caution. We discuss this in detail in \cref{app:sequential_optimization}.

\section{Bayesian Parameter Estimation}
\label{sec:method_parameter_estimation}
Once we obtain a causal structure $\GE{i}$, we address the critical task of \textit{parameter estimation} within the Structural Equation Model (SEM). The parameters ${\vphi = \{\vtheta, \sigma^2\}}$ include edge weights (coefficients) and noise parameters. With $\GE{i}$ potentially containing both observed and latent variables $\VE{i} {=} \{\VD, \VL\}$, we represent observed variable edges as $\vthetaO$ and latent confounder edges as $\vthetaL$, giving $\vtheta = \{\vthetaO, \vthetaL\}$.

When no latent confounders exist ($\VL {=} \emptyset$), standard Maximum Likelihood Estimation (MLE) optimizes the parameters $\vphi{=}\{\vthetaO, \sigma^2\}$ as ${\vphi = \arg\max_{\vphi} \log p(\DD_i \mid \GE{i}, \vphi)}$
using a conventional gradient-based methods. We focus on the more important and challenging scenario where latent confounders exist ($\VL {\neq} \emptyset$).

With latent confounders, MLE is ill-posed and intractable. We instead employ an iterative Expectation--Maximization (EM) algorithm that incorporates LM-provided probability $p(\VL)$ and correlation $\rho(\VD, \VL)$ about latent confounders. Specifically, we propose the following EM steps:
\begin{itemize}[noitemsep,topsep=0pt,leftmargin=*]
\item \textbf{E-step:} Compute conditional posterior of latent confounder(s) given $\DD_i$ and SEM parameters $\vphi$,
\begin{align}
	\label{eq:e_step}
	p(\VL \mid \GE{i}, \DD_i, \vphi) \propto p(\DD_i \mid \GE{i}, \VL, \vphi) p(\VL).
\end{align}
\item \textbf{M-step:} Update parameters by maximizing the expected log-likelihood, incorporating LM-provided regularization for latent confounder edges,
\begin{align}
	\label{eq:m_step}
	\vphi \in \arg \max_{\vphi} &\E_{p(\VL \mid \GE{i}, \DD_i, \vphi)}[
	\log p(\DD_i \mid \GE{i}, \VL , \vphi)] - \lambda \|{\vthetaL - (\rho(\VD, \VL) \sigma_{\VD}\sigma_{\VL}^{-1})}\|_2 \, .
\end{align}
\end{itemize}

\cref{alg:llm-causal-discovery-parameter-estimation} details the EM parameter estimation algorithm. In \cref{sec:exp_parameter_estimation}, we demonstrate the robustness and recovery capability of the proposed parameter estimation algorithm.
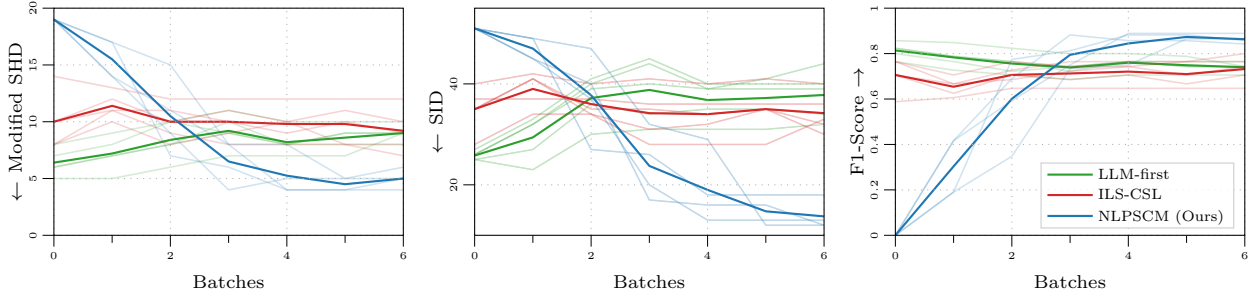
\begin{figure*}[t!]
	\centering\footnotesize
	\setlength{\figurewidth}{.28\textwidth}
	\setlength{\figureheight}{.65\figurewidth}
	\scriptsize
	\pgfplotsset{axis on top,scale only axis,width=\figurewidth,height=\figureheight, ylabel near ticks,ylabel style={yshift=-2pt},xlabel style={yshift=3pt},y tick label style={rotate=90}, legend style={nodes={scale=0.8, transform shape}},tick label style={font=\tiny,scale=.8}}
	\pgfplotsset{legend cell align={left},every axis/.append style={legend style={draw=none,inner xsep=2pt, inner ysep=0.5pt, nodes={inner sep=2pt, text depth=0.1em},fill=white,fill opacity=0.8}}}
	\pgfplotsset{grid style={dotted, gray},xlabel={Batches},minor x tick num=1}
	\begin{subfigure}{0.32\textwidth}
\begin{tikzpicture}

\definecolor{crimson2143940}{RGB}{214,39,40}
\definecolor{darkgray176}{RGB}{176,176,176}
\definecolor{forestgreen4416044}{RGB}{44,160,44}
\definecolor{steelblue31119180}{RGB}{31,119,180}

\begin{axis}[
tick align=outside,
tick pos=left,
x grid style={darkgray176},
xmajorgrids,
xmin=0, xmax=6,
xtick style={color=black},
y grid style={darkgray176},
ylabel={$\leftarrow$ Modified SHD},
ymajorgrids,
ymin=0, ymax=20,
ytick style={color=black}
]
\addplot [line width=1.5pt, thick, forestgreen4416044]
table {%
0 6.4
1 7.2
2 8.4
3 9.2
4 8.2
5 8.6
6 9
};
\addplot [semithick, forestgreen4416044, opacity=0.2]
table {%
0 8
1 9
2 10
3 11
4 10
5 10
6 10
};
\addplot [semithick, forestgreen4416044, opacity=0.2]
table {%
0 7
1 8
2 10
3 9
4 8
5 9
6 9
};
\addplot [semithick, forestgreen4416044, opacity=0.2]
table {%
0 5
1 5
2 6
3 7
4 7
5 7
6 9
};
\addplot [semithick, forestgreen4416044, opacity=0.2]
table {%
0 6
1 7
2 8
3 9
4 8
5 8
6 8
};
\addplot [semithick, forestgreen4416044, opacity=0.2]
table {%
0 6
1 7
2 8
3 10
4 8
5 9
6 9
};
\addplot [line width=1.5pt, thick, crimson2143940]
table {%
0 10
1 11.3999996185303
2 10
3 10
4 9.80000019073486
5 9.80000019073486
6 9.19999980926514
};
\addplot [semithick, crimson2143940, opacity=0.2]
table {%
0 10
1 11
2 11
3 10
4 9
5 10
6 9
};
\addplot [semithick, crimson2143940, opacity=0.2]
table {%
0 10
1 12
2 10
3 11
4 10
5 11
6 10
};
\addplot [semithick, crimson2143940, opacity=0.2]
table {%
0 8
1 10
2 8
3 9
4 10
5 8
6 7
};
\addplot [semithick, crimson2143940, opacity=0.2]
table {%
0 14
1 13
2 12
3 12
4 12
5 12
6 12
};
\addplot [semithick, crimson2143940, opacity=0.2]
table {%
0 8
1 11
2 9
3 8
4 8
5 8
6 8
};
\addplot [line width=1.5pt, thick, steelblue31119180]
table {%
0 19
1 15.5
2 10.5
3 6.5
4 5.25
5 4.5
6 5
};
\addplot [semithick, steelblue31119180, opacity=0.2]
table {%
0 19
1 14
2 11
3 8
4 4
5 4
6 4
};
\addplot [semithick, steelblue31119180, opacity=0.2]
table {%
0 19
1 14
2 9
3 4
4 5
5 5
6 6
};
\addplot [semithick, steelblue31119180, opacity=0.2]
table {%
0 19
1 17
2 7
3 6
4 4
5 4
6 5
};
\addplot [semithick, steelblue31119180, opacity=0.2]
table {%
0 19
1 17
2 15
3 8
4 8
5 5
6 5
};
\end{axis}

\end{tikzpicture}
 	\end{subfigure}
	\hfill
	\begin{subfigure}{0.32\textwidth}
\begin{tikzpicture}

\definecolor{crimson2143940}{RGB}{214,39,40}
\definecolor{darkgray176}{RGB}{176,176,176}
\definecolor{forestgreen4416044}{RGB}{44,160,44}
\definecolor{steelblue31119180}{RGB}{31,119,180}

\begin{axis}[
tick align=outside,
tick pos=left,
x grid style={darkgray176},
xmajorgrids,
xmin=0, xmax=6,
xtick style={color=black},
y grid style={darkgray176},
ylabel={$\leftarrow$ SID },
ymajorgrids,
ymin=10, ymax=55,
ytick style={color=black}
]
\path [draw=forestgreen4416044, fill=forestgreen4416044, opacity=0.1]
(axis cs:0,25.8)
--(axis cs:0,25.8)
--(axis cs:1,29.4)
--(axis cs:2,37.2)
--(axis cs:3,38.8)
--(axis cs:4,36.8)
--(axis cs:5,37.2)
--(axis cs:6,37.8)
--(axis cs:6,37.8)
--(axis cs:6,37.8)
--(axis cs:5,37.2)
--(axis cs:4,36.8)
--(axis cs:3,38.8)
--(axis cs:2,37.2)
--(axis cs:1,29.4)
--(axis cs:0,25.8)
--cycle;

\path [draw=forestgreen4416044, fill=forestgreen4416044, opacity=0.1]
(axis cs:0,27)
--(axis cs:0,27)
--(axis cs:1,33)
--(axis cs:2,41)
--(axis cs:3,45)
--(axis cs:4,40)
--(axis cs:5,40)
--(axis cs:6,40)
--(axis cs:6,40)
--(axis cs:6,40)
--(axis cs:5,40)
--(axis cs:4,40)
--(axis cs:3,45)
--(axis cs:2,41)
--(axis cs:1,33)
--(axis cs:0,27)
--cycle;

\path [draw=forestgreen4416044, fill=forestgreen4416044, opacity=0.1]
(axis cs:0,25)
--(axis cs:0,25)
--(axis cs:1,27)
--(axis cs:2,36)
--(axis cs:3,34)
--(axis cs:4,35)
--(axis cs:5,35)
--(axis cs:6,34)
--(axis cs:6,34)
--(axis cs:6,34)
--(axis cs:5,35)
--(axis cs:4,35)
--(axis cs:3,34)
--(axis cs:2,36)
--(axis cs:1,27)
--(axis cs:0,25)
--cycle;

\path [draw=forestgreen4416044, fill=forestgreen4416044, opacity=0.1]
(axis cs:0,25)
--(axis cs:0,25)
--(axis cs:1,23)
--(axis cs:2,30)
--(axis cs:3,31)
--(axis cs:4,31)
--(axis cs:5,31)
--(axis cs:6,32)
--(axis cs:6,32)
--(axis cs:6,32)
--(axis cs:5,31)
--(axis cs:4,31)
--(axis cs:3,31)
--(axis cs:2,30)
--(axis cs:1,23)
--(axis cs:0,25)
--cycle;

\path [draw=forestgreen4416044, fill=forestgreen4416044, opacity=0.1]
(axis cs:0,26)
--(axis cs:0,26)
--(axis cs:1,32)
--(axis cs:2,40)
--(axis cs:3,44)
--(axis cs:4,39)
--(axis cs:5,39)
--(axis cs:6,39)
--(axis cs:6,39)
--(axis cs:6,39)
--(axis cs:5,39)
--(axis cs:4,39)
--(axis cs:3,44)
--(axis cs:2,40)
--(axis cs:1,32)
--(axis cs:0,26)
--cycle;

\path [draw=forestgreen4416044, fill=forestgreen4416044, opacity=0.1]
(axis cs:0,26)
--(axis cs:0,26)
--(axis cs:1,32)
--(axis cs:2,39)
--(axis cs:3,40)
--(axis cs:4,39)
--(axis cs:5,41)
--(axis cs:6,44)
--(axis cs:6,44)
--(axis cs:6,44)
--(axis cs:5,41)
--(axis cs:4,39)
--(axis cs:3,40)
--(axis cs:2,39)
--(axis cs:1,32)
--(axis cs:0,26)
--cycle;

\path [draw=crimson2143940, fill=crimson2143940, opacity=0.1]
(axis cs:0,35)
--(axis cs:0,35)
--(axis cs:1,39)
--(axis cs:2,36)
--(axis cs:3,34.2)
--(axis cs:4,34)
--(axis cs:5,35)
--(axis cs:6,34.2)
--(axis cs:6,34.2)
--(axis cs:6,34.2)
--(axis cs:5,35)
--(axis cs:4,34)
--(axis cs:3,34.2)
--(axis cs:2,36)
--(axis cs:1,39)
--(axis cs:0,35)
--cycle;

\path [draw=crimson2143940, fill=crimson2143940, opacity=0.1]
(axis cs:0,35)
--(axis cs:0,35)
--(axis cs:1,41)
--(axis cs:2,34)
--(axis cs:3,31)
--(axis cs:4,32)
--(axis cs:5,35)
--(axis cs:6,32)
--(axis cs:6,32)
--(axis cs:6,32)
--(axis cs:5,35)
--(axis cs:4,32)
--(axis cs:3,31)
--(axis cs:2,34)
--(axis cs:1,41)
--(axis cs:0,35)
--cycle;

\path [draw=crimson2143940, fill=crimson2143940, opacity=0.1]
(axis cs:0,40)
--(axis cs:0,40)
--(axis cs:1,42)
--(axis cs:2,40)
--(axis cs:3,41)
--(axis cs:4,40)
--(axis cs:5,41)
--(axis cs:6,40)
--(axis cs:6,40)
--(axis cs:6,40)
--(axis cs:5,41)
--(axis cs:4,40)
--(axis cs:3,41)
--(axis cs:2,40)
--(axis cs:1,42)
--(axis cs:0,40)
--cycle;

\path [draw=crimson2143940, fill=crimson2143940, opacity=0.1]
(axis cs:0,35)
--(axis cs:0,35)
--(axis cs:1,41)
--(axis cs:2,35)
--(axis cs:3,35)
--(axis cs:4,34)
--(axis cs:5,35)
--(axis cs:6,30)
--(axis cs:6,30)
--(axis cs:6,30)
--(axis cs:5,35)
--(axis cs:4,34)
--(axis cs:3,35)
--(axis cs:2,35)
--(axis cs:1,41)
--(axis cs:0,35)
--cycle;

\path [draw=crimson2143940, fill=crimson2143940, opacity=0.1]
(axis cs:0,37)
--(axis cs:0,37)
--(axis cs:1,37)
--(axis cs:2,37)
--(axis cs:3,36)
--(axis cs:4,36)
--(axis cs:5,36)
--(axis cs:6,36)
--(axis cs:6,36)
--(axis cs:6,36)
--(axis cs:5,36)
--(axis cs:4,36)
--(axis cs:3,36)
--(axis cs:2,37)
--(axis cs:1,37)
--(axis cs:0,37)
--cycle;

\path [draw=crimson2143940, fill=crimson2143940, opacity=0.1]
(axis cs:0,28)
--(axis cs:0,28)
--(axis cs:1,34)
--(axis cs:2,34)
--(axis cs:3,28)
--(axis cs:4,28)
--(axis cs:5,28)
--(axis cs:6,33)
--(axis cs:6,33)
--(axis cs:6,33)
--(axis cs:5,28)
--(axis cs:4,28)
--(axis cs:3,28)
--(axis cs:2,34)
--(axis cs:1,34)
--(axis cs:0,28)
--cycle;

\path [draw=steelblue31119180, fill=steelblue31119180, opacity=0.1]
(axis cs:0,51)
--(axis cs:0,51)
--(axis cs:1,47)
--(axis cs:2,37.75)
--(axis cs:3,23.75)
--(axis cs:4,19)
--(axis cs:5,14.75)
--(axis cs:6,13.75)
--(axis cs:6,13.75)
--(axis cs:6,13.75)
--(axis cs:5,14.75)
--(axis cs:4,19)
--(axis cs:3,23.75)
--(axis cs:2,37.75)
--(axis cs:1,47)
--(axis cs:0,51)
--cycle;

\path [draw=steelblue31119180, fill=steelblue31119180, opacity=0.1]
(axis cs:0,51)
--(axis cs:0,51)
--(axis cs:1,45)
--(axis cs:2,37)
--(axis cs:3,20)
--(axis cs:4,13)
--(axis cs:5,13)
--(axis cs:6,13)
--(axis cs:6,13)
--(axis cs:6,13)
--(axis cs:5,13)
--(axis cs:4,13)
--(axis cs:3,20)
--(axis cs:2,37)
--(axis cs:1,45)
--(axis cs:0,51)
--cycle;

\path [draw=steelblue31119180, fill=steelblue31119180, opacity=0.1]
(axis cs:0,51)
--(axis cs:0,51)
--(axis cs:1,45)
--(axis cs:2,40)
--(axis cs:3,17)
--(axis cs:4,16)
--(axis cs:5,16)
--(axis cs:6,12)
--(axis cs:6,12)
--(axis cs:6,12)
--(axis cs:5,16)
--(axis cs:4,16)
--(axis cs:3,17)
--(axis cs:2,40)
--(axis cs:1,45)
--(axis cs:0,51)
--cycle;

\path [draw=steelblue31119180, fill=steelblue31119180, opacity=0.1]
(axis cs:0,51)
--(axis cs:0,51)
--(axis cs:1,49)
--(axis cs:2,27)
--(axis cs:3,26)
--(axis cs:4,18)
--(axis cs:5,18)
--(axis cs:6,18)
--(axis cs:6,18)
--(axis cs:6,18)
--(axis cs:5,18)
--(axis cs:4,18)
--(axis cs:3,26)
--(axis cs:2,27)
--(axis cs:1,49)
--(axis cs:0,51)
--cycle;

\path [draw=steelblue31119180, fill=steelblue31119180, opacity=0.1]
(axis cs:0,51)
--(axis cs:0,51)
--(axis cs:1,49)
--(axis cs:2,47)
--(axis cs:3,32)
--(axis cs:4,29)
--(axis cs:5,12)
--(axis cs:6,12)
--(axis cs:6,12)
--(axis cs:6,12)
--(axis cs:5,12)
--(axis cs:4,29)
--(axis cs:3,32)
--(axis cs:2,47)
--(axis cs:1,49)
--(axis cs:0,51)
--cycle;

\addplot [line width=1.5pt, thick, forestgreen4416044, opacity=1]
table {%
0 25.8
1 29.4
2 37.2
3 38.8
4 36.8
5 37.2
6 37.8
};
\addplot [semithick, forestgreen4416044, opacity=0.2]
table {%
0 27
1 33
2 41
3 45
4 40
5 40
6 40
};
\addplot [semithick, forestgreen4416044, opacity=0.2]
table {%
0 25
1 27
2 36
3 34
4 35
5 35
6 34
};
\addplot [semithick, forestgreen4416044, opacity=0.2]
table {%
0 25
1 23
2 30
3 31
4 31
5 31
6 32
};
\addplot [semithick, forestgreen4416044, opacity=0.2]
table {%
0 26
1 32
2 40
3 44
4 39
5 39
6 39
};
\addplot [semithick, forestgreen4416044, opacity=0.2]
table {%
0 26
1 32
2 39
3 40
4 39
5 41
6 44
};
\addplot [line width=1.5pt, thick, crimson2143940, opacity=1]
table {%
0 35
1 39
2 36
3 34.2
4 34
5 35
6 34.2
};
\addplot [semithick, crimson2143940, opacity=0.2]
table {%
0 35
1 41
2 34
3 31
4 32
5 35
6 32
};
\addplot [semithick, crimson2143940, opacity=0.2]
table {%
0 40
1 42
2 40
3 41
4 40
5 41
6 40
};
\addplot [semithick, crimson2143940, opacity=0.2]
table {%
0 35
1 41
2 35
3 35
4 34
5 35
6 30
};
\addplot [semithick, crimson2143940, opacity=0.2]
table {%
0 37
1 37
2 37
3 36
4 36
5 36
6 36
};
\addplot [semithick, crimson2143940, opacity=0.2]
table {%
0 28
1 34
2 34
3 28
4 28
5 28
6 33
};
\addplot [line width=1.5pt, thick, steelblue31119180, opacity=1]
table {%
0 51
1 47
2 37.75
3 23.75
4 19
5 14.75
6 13.75
};
\addplot [semithick, steelblue31119180, opacity=0.2]
table {%
0 51
1 45
2 37
3 20
4 13
5 13
6 13
};
\addplot [semithick, steelblue31119180, opacity=0.2]
table {%
0 51
1 45
2 40
3 17
4 16
5 16
6 12
};
\addplot [semithick, steelblue31119180, opacity=0.2]
table {%
0 51
1 49
2 27
3 26
4 18
5 18
6 18
};
\addplot [semithick, steelblue31119180, opacity=0.2]
table {%
0 51
1 49
2 47
3 32
4 29
5 12
6 12
};
\end{axis}

\end{tikzpicture}
 	\end{subfigure}
	\hfill
	\begin{subfigure}{0.32\textwidth}
\begin{tikzpicture}

\definecolor{crimson2143940}{RGB}{214,39,40}
\definecolor{darkgray176}{RGB}{176,176,176}
\definecolor{forestgreen4416044}{RGB}{44,160,44}
\definecolor{lightgray204}{RGB}{204,204,204}
\definecolor{steelblue31119180}{RGB}{31,119,180}

\begin{axis}[
legend cell align={left},
legend style={
  fill opacity=0.8,
  draw opacity=1,
  text opacity=1,
  at={(0.97,0.03)},
  anchor=south east,
  draw=lightgray204
},
tick align=outside,
tick pos=left,
x grid style={darkgray176},
xmajorgrids,
xmin=0, xmax=6,
xtick style={color=black},
y grid style={darkgray176},
ylabel={F1-Score $\rightarrow$},
ymajorgrids,
ymin=0, ymax=1,
ytick style={color=black}
]
\addplot [line width=1.5pt, thick, forestgreen4416044]
table {%
0 0.813781512555604
1 0.783244206724675
2 0.756209150277382
3 0.738730158680484
4 0.759999999950627
5 0.748836261728092
6 0.739831932723606
};
\addlegendentry{LLM-first}
\addplot [semithick, forestgreen4416044, opacity=0.2, forget plot]
table {%
0 0.764705882303633
1 0.727272727223875
2 0.705882352891869
3 0.685714285664653
4 0.705882352891869
5 0.705882352891869
6 0.705882352891869
};
\addplot [semithick, forestgreen4416044, opacity=0.2, forget plot]
table {%
0 0.799999999950367
1 0.764705882303633
2 0.722222222172377
3 0.74285714280751
4 0.764705882303633
5 0.74285714280751
6 0.74285714280751
};
\addplot [semithick, forestgreen4416044, opacity=0.2, forget plot]
table {%
0 0.857142857093225
1 0.848484848435996
2 0.823529411715398
3 0.799999999950367
4 0.799999999950367
5 0.787878787829936
6 0.74285714280751
};
\addplot [semithick, forestgreen4416044, opacity=0.2, forget plot]
table {%
0 0.823529411715398
1 0.787878787829936
2 0.764705882303633
3 0.74285714280751
4 0.764705882303633
5 0.764705882303633
6 0.764705882303633
};
\addplot [semithick, forestgreen4416044, opacity=0.2, forget plot]
table {%
0 0.823529411715398
1 0.787878787829936
2 0.764705882303633
3 0.722222222172377
4 0.764705882303633
5 0.74285714280751
6 0.74285714280751
};
\addplot [line width=1.5pt, thick, crimson2143940]
table {%
0 0.705882352891869
1 0.654055258418202
2 0.706126814312827
3 0.712511671285655
4 0.720672268858125
5 0.709803921519411
6 0.732100840286697
};
\addlegendentry{ILS-CSL}
\addplot [semithick, crimson2143940, opacity=0.2, forget plot]
table {%
0 0.705882352891869
1 0.666666666617815
2 0.685714285664653
3 0.722222222172377
4 0.74285714280751
5 0.705882352891869
6 0.74285714280751
};
\addplot [semithick, crimson2143940, opacity=0.2, forget plot]
table {%
0 0.705882352891869
1 0.624999999951758
2 0.705882352891869
3 0.685714285664653
4 0.705882352891869
5 0.666666666617815
6 0.705882352891869
};
\addplot [semithick, crimson2143940, opacity=0.2, forget plot]
table {%
0 0.764705882303633
1 0.705882352891869
2 0.764705882303633
3 0.74285714280751
4 0.74285714280751
5 0.764705882303633
6 0.799999999950367
};
\addplot [semithick, crimson2143940, opacity=0.2, forget plot]
table {%
0 0.588235294068339
1 0.606060606011754
2 0.647058823480104
3 0.647058823480104
4 0.647058823480104
5 0.647058823480104
6 0.647058823480104
};
\addplot [semithick, crimson2143940, opacity=0.2, forget plot]
table {%
0 0.764705882303633
1 0.666666666617815
2 0.727272727223875
3 0.764705882303633
4 0.764705882303633
5 0.764705882303633
6 0.764705882303633
};
\addplot [line width=1.5pt, thick, steelblue31119180]
table {%
0 0
1 0.303571428546319
2 0.601066850046734
3 0.794546568579392
4 0.844596171752797
5 0.873312387968583
6 0.863250468463766
};
\addlegendentry{\ours (Ours)}
\addplot [semithick, steelblue31119180, opacity=0.2, forget plot]
table {%
0 0
1 0.416666666633681
2 0.592592592550892
3 0.749999999951758
4 0.888888888839043
5 0.888888888839043
6 0.888888888839043
};
\addplot [semithick, steelblue31119180, opacity=0.2, forget plot]
table {%
0 0
1 0.416666666633681
2 0.689655172368609
3 0.882352941127163
4 0.857142857093225
5 0.857142857093225
6 0.842105263107895
};
\addplot [semithick, steelblue31119180, opacity=0.2, forget plot]
table {%
0 0
1 0.190476190458957
2 0.774193548339646
3 0.812499999951758
4 0.882352941127163
5 0.882352941127163
6 0.857142857093225
};
\addplot [semithick, steelblue31119180, opacity=0.2, forget plot]
table {%
0 0
1 0.190476190458957
2 0.347826086927788
3 0.733333333286889
4 0.749999999951758
5 0.864864864814901
6 0.864864864814901
};
\end{axis}

\end{tikzpicture}
 	\end{subfigure}
	\vspace*{-1em}
	\caption{\textbf{\sc User Level Data - I:} Performance evolution across batches for \textit{Data-LM} methods. Left: Modified Structural Hamming Distance ($\downarrow$), Middle: Structural Intervention Distance ($\downarrow$), and Right: F1-Score ($\uparrow$). \ours consistently outperforms other approaches as data accumulation progresses.\looseness-1}
	\label{fig:user_1_metrics}
    \vspace*{-1em}
\end{figure*}	

\section{Experiments}
To provide a robust empirical examination, we conduct experiments on four language models: GPT-3.5\textsubscript{turbo}, GPT4.1\textsubscript{nano}, Llama3.1\textsubscript{8B-instruct}, Qwen3\textsubscript{4B-instruct} and six datasets: {\sc Earthquake} from \citet{korb2010bayesian}, {\sc Asia} from \citet{lauritzen1988local}, {\sc User Level Data - I}, {\sc User Level Data - II} from \citet{googleanalytics}, {\sc Child} from \citet{spiegelhalter1993bayesian}, and {\sc Alarm} from \citet{beinlich1989alarm}. They range in number of observed variables (nodes) from small (5) to medium (19) to large (37). {\sc Earthquake}, {\sc Asia}, {\sc Child}, and {\sc Alarm} datasets are standard benchmarks, providing observational data and ground-truth causal structures. To simulate a streaming batch setting, each dataset is split into batches. For the two {\sc User Level Data}, which contain only observational data, the underlying DAG is inferred using DirectLiNGAM, an algorithm proposed by \citet{shimizu2011directlingam}.
We treat the DAG inferred from DirectLiNGAM as the ground truth for evaluation. Further details on the data sets and simulation process are provided in \cref{app:dataset_details}.

Our evaluation metrics include a \textit{modified} Structural Hamming Distance (Mod.\ SHD), which extends SHD to account for uncertain edges in PAG; Structural Intervention Distance (SID); and precision, recall, and F1-score for causal relations that are certain. Together, these metrics assess structural accuracy, interventional soundness, and edge-wise discovery performance (details in \cref{app:metrics}).

\subsection{Structure Learning}
\label{sec:exp_structure_learning}

We evaluate \ours against \textit{Only-Data} and \textit{Data-LM} baselines in \cref{tbl:causal_discovery_metrics}. As shown in \cref{tbl:llm_optimistic_nature}, \textit{Only-LM} methods exhibit \textit{overly optimistic} behavior, producing globally plausible but locally unreliable causal structures. This highlights the need for data-grounded post-processing. \cref{tbl:causal_discovery_metrics} compares \ours with several baselines, including multiple FCI variants (cumulative, vanilla, iterative, heuristics), as well as \textit{Data-LM} approaches (LM-first, and ILS-CSL proposed by \citet{ban2023causal}). Across all evaluation metrics, \ours consistently outperforms baselines, which also holds for different LM temperatures (\cref{tbl:causal_discovery_metrics_all_t}). While \cref{tbl:causal_discovery_metrics} reports metrics for the final batch, we also show performance evolution across batches, a crucial step in sequential settings (\cf \cref{fig:user_1_metrics}). We justify the use of FCI over other causal discovery algorithms in \cref{app:other_causal_discovery_algorithms}. We provide more experiment details in \cref{app:experiment_details}.

Finally, going beyond GPT-3.5\textsubscript{turbo} (\cref{tbl:causal_discovery_metrics}), %
results with recent LMs, GPT-4o and GPT-5 (\cref{tbl:gpt4o_gpt5_results}) show %
good performance gains for \ours across LMs. The much higher inference cost of GPT-4o and GPT-5, over GPT-3.5\textsubscript{turbo} constrain their use for large set of experiments.
\begin{figure*}[t!]
	\centering\scriptsize
	\setlength{\figurewidth}{.28\textwidth}
	\setlength{\figureheight}{.65\figurewidth}
	\pgfplotsset{axis on top,scale only axis,width=\figurewidth,height=\figureheight, ylabel near ticks,ylabel style={yshift=-2pt},xlabel style={yshift=3pt},y tick label style={rotate=90}, legend style={nodes={scale=0.8, transform shape}},tick label style={font=\tiny,scale=.8}}
	\pgfplotsset{legend cell align={left},every axis/.append style={legend style={draw=none,inner xsep=2pt, inner ysep=0.5pt, nodes={inner sep=2pt, text depth=0.1em},fill=white,fill opacity=0.8}}}
	\pgfplotsset{grid style={dotted, gray},xlabel={Batch sequence},minor x tick num=1}	
	\begin{subfigure}{0.32\textwidth}
\begin{tikzpicture}

\definecolor{crimson2143940}{RGB}{214,39,40}
\definecolor{gray}{RGB}{128,128,128}
\definecolor{lightgray204}{RGB}{204,204,204}
\definecolor{steelblue31119180}{RGB}{31,119,180}

\begin{axis}[
legend cell align={left},
legend style={fill opacity=0.8, draw opacity=1, text opacity=1, draw=lightgray204},
tick align=outside,
tick pos=left,
x grid style={gray},
xlabel={Batches},
xmajorgrids,
xmin=0, xmax=6,
xtick style={color=black},
y grid style={gray},
ylabel={$\leftarrow$ Modified SHD},
ymajorgrids,
ymin=3, ymax=20,
ytick style={color=black}
]
\addplot [very thin, crimson2143940, opacity=0.5, forget plot]
table {%
0 17
1 16
2 11
3 9
4 6
5 6
6 6
};
\addplot [very thin, crimson2143940, opacity=0.5, forget plot]
table {%
0 18
1 15
2 11
3 8
4 7
5 6
6 6
};
\addplot [very thin, crimson2143940, opacity=0.5, forget plot]
table {%
0 17
1 14
2 11
3 7
4 7
5 6
6 6
};
\addplot [very thin, crimson2143940, opacity=0.5, forget plot]
table {%
0 17
1 14
2 8
3 6
4 6
5 6
6 6
};
\addplot [very thin, steelblue31119180, opacity=0.5, forget plot]
table {%
0 13
1 9
2 6
3 5
4 5
5 5
6 4
};
\addplot [very thin, steelblue31119180, opacity=0.5, forget plot]
table {%
0 17
1 12
2 8
3 6
4 6
5 6
6 5
};
\addplot [very thin, steelblue31119180, opacity=0.5, forget plot]
table {%
0 17
1 11
2 7
3 6
4 6
5 6
6 5
};
\addplot [very thin, steelblue31119180, opacity=0.5, forget plot]
table {%
0 18
1 14
2 9
3 7
4 6
5 6
6 5
};
\addplot [very thin, steelblue31119180, opacity=0.5, forget plot]
table {%
0 16
1 12
2 5
3 5
4 5
5 5
6 4
};
\addplot [very thin, steelblue31119180, opacity=0.5, forget plot]
table {%
0 19
1 12
2 9
3 6
4 6
5 6
6 5
};
\addplot [line width=1.5pt, thick, steelblue31119180]
table {%
0 16.6666660308838
1 11.6666669845581
2 7.33333349227905
3 5.83333349227905
4 5.66666650772095
5 5.66666650772095
6 4.66666650772095
};
\addlegendentry{\ours-Score}
\addplot [line width=1.5pt, thick, crimson2143940]
table {%
0 17.25
1 14.75
2 10.25
3 7.5
4 6.5
5 6
6 6
};
\addlegendentry{Random}
\end{axis}

\end{tikzpicture}
 	\end{subfigure}
	\hfill
	\begin{subfigure}{0.32\textwidth}
\begin{tikzpicture}

\definecolor{crimson2143940}{RGB}{214,39,40}
\definecolor{gray}{RGB}{128,128,128}
\definecolor{lightgray204}{RGB}{204,204,204}
\definecolor{steelblue31119180}{RGB}{31,119,180}

\begin{axis}[
legend cell align={left},
legend style={
  fill opacity=0.8,
  draw opacity=1,
  text opacity=1,
  at={(0.97,0.03)},
  anchor=south east,
  draw=lightgray204
},
tick align=outside,
tick pos=left,
x grid style={gray},
xlabel={Batches},
xmajorgrids,
xmin=0, xmax=6,
xtick style={color=black},
y grid style={gray},
ylabel={F1-Score $\rightarrow$ },
ymajorgrids,
ymin=0.0552630019426694, ymax=0.93471126306568,
ytick style={color=black}
]
\addplot [very thin, crimson2143940, opacity=0.5, forget plot]
table {%
0 0.190476018140746
1 0.384614991124663
2 0.645160815817206
3 0.742856646530944
4 0.842104763158192
5 0.842104763158192
6 0.842104763158192
};
\addplot [very thin, crimson2143940, opacity=0.5, forget plot]
table {%
0 0.0999999050000903
1 0.399999635200333
2 0.645160815817206
3 0.777777279321307
4 0.810810311176349
5 0.842104763158192
6 0.842104763158192
};
\addplot [very thin, crimson2143940, opacity=0.5, forget plot]
table {%
0 0.190476018140746
1 0.461538068047673
2 0.645160815817206
3 0.810810311176349
4 0.810810311176349
5 0.842104763158192
6 0.842104763158192
};
\addplot [very thin, crimson2143940, opacity=0.5, forget plot]
table {%
0 0.190476018140746
1 0.461538068047673
2 0.749999517578435
3 0.833332834876841
4 0.842104763158192
5 0.842104763158192
6 0.842104763158192
};
\addplot [very thin, steelblue31119180, opacity=0.5, forget plot]
table {%
0 0.479999635200277
1 0.727272238751476
2 0.833332834876841
3 0.864864365230384
4 0.864864365230384
5 0.864864365230384
6 0.894736342105543
};
\addplot [very thin, steelblue31119180, opacity=0.5, forget plot]
table {%
0 0.260869277883114
1 0.599999535555915
2 0.777777279321307
3 0.842104763158192
4 0.842104763158192
5 0.842104763158192
6 0.871794372123889
};
\addplot [very thin, steelblue31119180, opacity=0.5, forget plot]
table {%
0 0.190476018140746
1 0.645160815817206
2 0.810810311176349
3 0.842104763158192
4 0.842104763158192
5 0.842104763158192
6 0.871794372123889
};
\addplot [very thin, steelblue31119180, opacity=0.5, forget plot]
table {%
0 0.181817946281297
1 0.499999563775891
2 0.742856646530944
3 0.810810311176349
4 0.842104763158192
5 0.842104763158192
6 0.871794372123889
};
\addplot [very thin, steelblue31119180, opacity=0.5, forget plot]
table {%
0 0.272727037190286
1 0.571428135204415
2 0.864864365230384
3 0.864864365230384
4 0.864864365230384
5 0.864864365230384
6 0.894736342105543
};
\addplot [very thin, steelblue31119180, opacity=0.5, forget plot]
table {%
0 0.0952379229028062
1 0.599999535555915
2 0.742856646530944
3 0.842104763158192
4 0.842104763158192
5 0.842104763158192
6 0.871794372123889
};
\addplot [line width=1.5pt, thick, steelblue31119180]
table {%
0 0.246854639599754
1 0.607309970776803
2 0.795416347277795
3 0.844475555185282
4 0.849691297182256
5 0.849691297182256
6 0.879441695451107
};
\addlegendentry{\ours-Score}
\addplot [line width=1.5pt, thick, crimson2143940]
table {%
0 0.167856989855582
1 0.426922690605085
2 0.671370491257514
3 0.79119426797636
4 0.82645753716727
5 0.842104763158192
6 0.842104763158192
};
\addlegendentry{Random}
\end{axis}

\end{tikzpicture}
 	\end{subfigure}
	\hfill
	\begin{subfigure}{0.32\textwidth}
\begin{tikzpicture}

\definecolor{crimson2143940}{RGB}{214,39,40}
\definecolor{gray}{RGB}{128,128,128}
\definecolor{lightgray204}{RGB}{204,204,204}
\definecolor{steelblue31119180}{RGB}{31,119,180}

\begin{axis}[
legend cell align={left},
legend style={fill opacity=0.8, draw opacity=1, text opacity=1, draw=lightgray204},
tick align=outside,
tick pos=left,
width=\figurewidth,
height=\figureheight,
x grid style={gray},
xlabel={Batches},
xmajorgrids,
xmin=0, xmax=6,
xtick style={color=black},
y grid style={gray},
ylabel={$\leftarrow$ Modified SHD},
ymajorgrids,
ymin=3, ymax=20,
ytick style={color=black}
]
\addplot [very thin, crimson2143940, opacity=0.5, forget plot]
table {%
0 19
1 19
2 17
3 15
4 14
5 15
6 14
};
\addplot [very thin, crimson2143940, opacity=0.5, forget plot]
table {%
0 19
1 19
2 20
3 18
4 17
5 16
6 16
};
\addplot [very thin, crimson2143940, opacity=0.5, forget plot]
table {%
0 19
1 17
2 14
3 12
4 11
5 7
6 5
};
\addplot [very thin, crimson2143940, opacity=0.5, forget plot]
table {%
0 19
1 17
2 16
3 15
4 11
5 11
6 6
};
\addplot [very thin, crimson2143940, opacity=0.5, forget plot]
table {%
0 18
1 18
2 16
3 14
4 11
5 9
6 9
};
\addplot [very thin, steelblue31119180, opacity=0.5, forget plot]
table {%
0 16
1 13
2 5
3 5
4 5
5 5
6 5
};
\addplot [very thin, steelblue31119180, opacity=0.5, forget plot]
table {%
0 16
1 10
2 5
3 4
4 4
5 5
6 7
};
\addplot [very thin, steelblue31119180, opacity=0.5, forget plot]
table {%
0 18
1 13
2 12
3 5
4 5
5 4
6 4
};
\addplot [very thin, steelblue31119180, opacity=0.5, forget plot]
table {%
0 16
1 13
2 7
3 7
4 5
5 6
6 6
};
\addplot [very thin, steelblue31119180, opacity=0.5, forget plot]
table {%
0 12
1 9
2 4
3 5
4 6
5 7
6 7
};
\addplot [line width=1.5pt, thick, steelblue31119180]
table {%
0 15.6000003814697
1 11.6000003814697
2 6.59999990463257
3 5.19999980926514
4 5
5 5.40000009536743
6 5.80000019073486
};
\addlegendentry{\ours-threshold}
\addplot [line width=1.5pt, thick, crimson2143940]
table {%
0 18.7999992370605
1 18
2 16.6000003814697
3 14.8000001907349
4 12.8000001907349
5 11.6000003814697
6 10
};
\addlegendentry{Fixed threshold}
\end{axis}

\end{tikzpicture}
 	\end{subfigure}
	\vspace*{-1em}
	\caption{\textbf{Structure learning ablation:} The impact of two key components: \textit{selection score} and \textit{dynamic background threshold}. (Left, Middle)~Modified SHD~($\downarrow$) and F1-score~($\uparrow$) on the {\sc User Level Data - I} dataset, comparing \ours---\textit{selection score} against random selection. (Right)~Modified SHD($\downarrow$) comparing \ours--- \textit{dynamic threshold} with a conventional fixed threshold.\looseness-1}
	\label{fig:structure_learning_ablation}
    \vspace*{-1em}
\end{figure*}
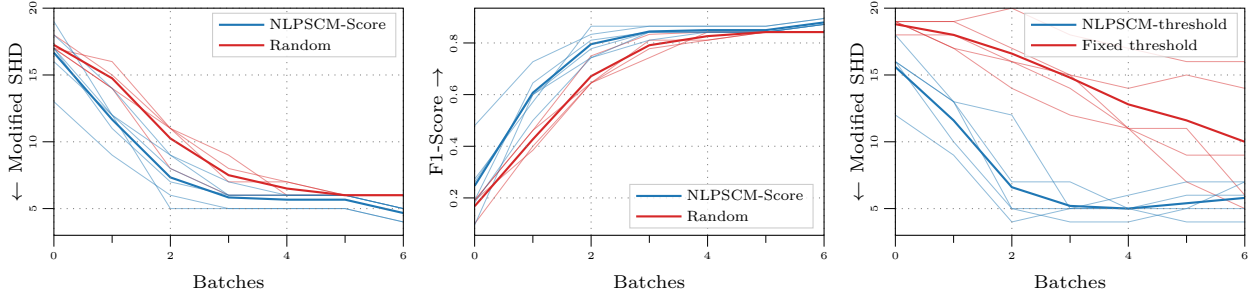	

\paragraph{Structure learning ablations}
We ablate two key components of \ours: \textit{(i)}~selection score for sequential optimization, and \textit{(ii)}~dynamic background threshold. Results of ablations performed on {\sc User Level Data - I} are shown in \cref{fig:structure_learning_ablation}. 

Effectiveness of proposed selection score (\cref{eq:score}) in guiding edge selection under a fixed budget of LM queries, is compared against a random-selection baseline. \cref{fig:structure_learning_ablation} (Left, Middle) shows that \ours achieves significantly better Mod.\ SHD and F1-score across batches, demonstrating the benefit of a principled edge selection policy in sequential structure learning. 

We also assess the impact of dynamic background threshold (\cref{eq:dynamic_bg}) used to promote edges from the histogram $\HE{}$ into the background knowledge $\B$. \cref{fig:structure_learning_ablation} (Right) reports Mod.\ SHD over batches, highlighting the advantages of a dynamic threshold over a conventional fixed one.  The adaptive mechanism yields more stable and accurate graph recovery throughout the learning process.

\subsection{Other LM Families and Memorization in LMs}
We evaluate \ours using three \textit{recent} LMs across different families: Llama3.1\textsubscript{8B-instruct}, Qwen3\textsubscript{4B-instruct}, and GPT4.1\textsubscript{nano}. The first two are open models. We evaluate on two datasets: {\sc Child}  and the real-world {\sc User Level Data - I} from \citet{googleanalytics}. Measuring performance on the 5 metrics over 5 runs, the results in \cref{tbl:llama_qwen_results} are consistent with those of the previously presented GPT3.5\textsubscript{Turbo} model, demonstrating robustness across LMs with different training data, and architectures.\looseness-1

We introduce an additional baseline which does causal discovery using LM and Pearson correlation in the prompt, namely BFS, as proposed by \citet{jiralerspong2024efficientcausalgraphdiscovery}. A comparison of \ours with BFS also serves an important purpose by throwing light on the issue of potential memorization of datasets seen in training by LMs. As noted by \citet{jiralerspong2024efficientcausalgraphdiscovery}, their method relies on the LM's training knowledge (Sec.~5 of their paper), thereby recognizing the dependence on memorization. To make a fair comparison, we use a recent model GPT4.1\textsubscript{nano}. Results reported in the bottom panel of \cref{tbl:llama_qwen_results} show that for both {\sc Child} and {\sc User Level Data - I},  
\ours beats BFS handily. To elucidate about memorization, we compare the difference-of-differences in metric-values between BFS and \ours. The difference is \textit{smaller} for {\sc Child} dataset, a standard causal discovery benchmark that is likely to be a part of the LM training data. In contrast, the difference is \emph{much larger} for the {\sc User Level Data-I} dataset. The latter dataset or its causal graph is unlikely to have been seen by any LM due to the processing and construction of attributes we performed in this dataset, from the publicly available large data at \citet{googleanalytics}. That BFS performs worse highlights its reliance on memorized knowledge, whereas \ours demonstrates greater robustness across datasets.\looseness-1

\begin{table*}[t!]
	\centering \scriptsize
	\caption{\textbf{\ours improves causal discovery across language model families:} We showcase results on the {\sc Child} and {\sc User level data-I} datasets using three language models: \textit{Llama3.1\textsubscript{8B-instruct}}, \textit{Qwen3\textsubscript{4B-instruct}} and a GPT-4.1\textsubscript{nano}. We report 5 metrics: \textit{Modified SHD, SID, Precision, Recall, F1}, with mean and standard deviation over $5$ runs and perform significance test with $\alpha=0.05$.}
	\label{tbl:llama_qwen_results}
	\setlength{\tabcolsep}{3.5pt}
	\resizebox{\textwidth}{!}{
\begin{tabular}{cllccccc}
	\toprule
	\textbf{Dataset} & \textbf{Model} & \textbf{Method} & \textbf{Mod. SHD $\downarrow$} & \textbf{SID $\downarrow$} & \textbf{Precision $\uparrow$} & \textbf{Recall $\uparrow$} & \textbf{F1 Score $\uparrow$}\\
	\midrule
	\multirow{11}{*}{\rotatebox{90}{\tiny{ \sc Child}}} & \multirow{3}{*}{Llama3.1\textsubscript{8B-instruct}} & LLM-first 
	    & \val{45.80}{1.94} 
	    & \val{(211.20,\,211.20)}{(15.25,\,15.25)} 
	    & \val{0.13}{0.01} 
	    & \val{0.19}{0.02}
        & \val{0.16}{0.01}\\
    && ILS-CSL 
        & \val{27.60}{2.24} 
        & \val{(146.40,\,146.40)}{(15.93,\,15.93)} 
        & \val{0.37}{0.05} 
        & \val{0.38}{0.07}
        & \val{0.38}{0.06}\\
    & &\cellcolor{highlight} \ours 
	    &\cellcolor{highlight} \valb{24.60}{0.80} 
	    &\cellcolor{highlight} \valb{(101.80,\,101.80)}{(8.11,\,8.11)} 
	    &\cellcolor{highlight} \valb{0.44}{0.01} 
	    &\cellcolor{highlight} \valb{0.45}{0.02}
        &\cellcolor{highlight} \valb{0.44}{0.01}\\
    \cdashline{2-8}[.4pt/1pt]
    &\multirow{3}{*}{Qwen3\textsubscript{4B-instruct}} & LLM-first 
	    & \val{35.80}{2.04} 
	    & \val{(173.80,\,173.80)}{(14.13,\,14.13)} 
	    & \val{0.27}{0.02} 
	    & \valb{0.42}{0.02}
        & \val{0.34}{0.01} \\
    && ILS-CSL 
        & \val{30.60}{2.42} 
        & \val{(196.80,\,196.80)}{(14.13,\,14.13)} 
        & \val{0.30}{0.06} 
        & \val{0.28}{0.05}
        & \val{0.29}{0.05}\\
    && \cellcolor{highlight} \ours 
	    & \cellcolor{highlight} \valb{21.20}{0.75} 
	    & \cellcolor{highlight} \valb{(107.80,\,107.80)}{(13.42,\,13.42)} 
	    & \cellcolor{highlight} \valb{0.51}{0.01} 
	    & \cellcolor{highlight} \val{0.32}{0.03}
        & \cellcolor{highlight} \valb{0.39}{0.02}\\
    \cdashline{2-8}[.4pt/1pt]
    &\multirow{5}{*}{GPT4.1\textsubscript{nano}} & LLM-first 
	    & \val{31.83}{2.61}
        & \val{(203.50, 203.50)}{(23.06, 23.06)}
        & \val{0.16}{0.01}
        & \val{0.14}{0.03}
        & \val{0.15}{0.02} \\
    && ILS-CSL 
        & \val{31.60}{2.06}
        & \valb{(147.40, 147.40)}{(11.25, 11.25)}
        & \val{0.29}{0.05}
        & \val{0.19}{0.04}
        & \val{0.23}{0.03} \\
    && BFS 
        & \val{34.75}{7.80}
        & \val{(252.50, 252.50)}{(23.36, 23.36)}
        & \val{0.18}{0.08}
        & \val{0.12}{0.04}
        & \val{0.14}{0.03}  \\
    && BFS\textsubscript{corr} 
        & \val{32.30}{4.16}
        & \val{(301.10, 301.10}{(14.42, 14.42)}
        & \val{0.21}{0.03}
        & \val{0.15}{0.01}
        & \val{0.18}{0.01} \\
    && \cellcolor{highlight} \ours 
	    & \valb{27.90}{1.61} \cellcolor{highlight}
	    & \valb{(143.50, 165.50)}{(12.23, 35.97)} \cellcolor{highlight}
	    & \valb{0.42}{0.06} \cellcolor{highlight}
	    & \valb{0.23}{0.04} \cellcolor{highlight}
        & \valb{0.30}{0.03} \cellcolor{highlight}\\
	\midrule
	\multirow{11}{*}{\rotatebox{90}{\tiny{ \sc User Level Data-I}}} & \multirow{3}{*}{Llama3.1\textsubscript{8B-instruct}} & LLM-first 
	    & \val{11.50}{0.50} 
	    & \val{(42.50,\,42.50)}{(1.50,\,1.50)} 
	    & \val{0.74}{0.01} 
	    & \val{0.61}{0.03}
        & \val{0.67}{0.02} \\
    && ILS-CSL 
        & \val{9.75}{1.64} 
        & \val{(32.25,\,32.25)}{(3.56,\,3.56)} 
        & \val{0.78}{0.05} 
        & \val{0.66}{0.03}
        & \val{0.72}{0.04}\\
    && \cellcolor{highlight}\ours 
	    & \cellcolor{highlight}\valb{7.50}{0.45} 
	    & \cellcolor{highlight}\valb{(25.80,\,25.40)}{(2.32,\,4.22)} 
	    & \cellcolor{highlight}\valb{0.92}{0.07} 
	    & \cellcolor{highlight}\valb{0.69}{0.01}
        & \cellcolor{highlight}\valb{0.79}{0.01} \\
    \cdashline{2-8}[.4pt/1pt]
    &\multirow{3}{*}{Qwen3\textsubscript{4B-instruct}} & LLM-first 
	    & \val{23.67}{1.11} 
	    & \val{(64.33,\,64.33)}{(0.94,\,0.94)} 
	    & \val{0.35}{0.04} 
	    & \val{0.28}{0.04}
        & \val{0.31}{0.04}\\
    && ILS-CSL 
        & \val{20.60}{4.18} 
        & \val{(56.80,\,56.80)}{(7.36,\,7.36)} 
        & \val{0.45}{0.13} 
        & \val{0.36}{0.10}
        & \val{0.40}{0.12}\\
    & & \cellcolor{highlight} \ours 
	    & \cellcolor{highlight} \valb{9.80}{2.04} 
	    & \cellcolor{highlight} \valb{(26.60,\,26.60)}{(6.83,\,6.83)} 
	    & \cellcolor{highlight} \valb{0.91}{0.08} 
	    & \cellcolor{highlight} \valb{0.54}{0.08}
        & \cellcolor{highlight} \valb{0.67}{0.07}\\
    \cdashline{2-8}[.4pt/1pt]
    &\multirow{5}{*}{GPT4.1\textsubscript{nano}} & LLM-first 
	    & \val{19.42}{1.14}
	    & \val{(72.10, 72.10)}{(5.36, 5.36)}
	    & \val{0.76}{0.07}
	    & \val{0.32}{0.03}
	    & \val{0.45}{0.03} \\
    && ILS-CSL 
        & \val{18.00}{1.41}
        & \val{(54.00, 54.00)}{(2.53, 2.53)}
        & \val{0.53}{0.05}
        & \valb{0.42}{0.03}
        & \val{0.47}{0.04} \\
    && BFS 
        & \val{36.60}{8.96}
        & \val{(65.80, 65.80)}{(2.17, 2.17)}
        & \val{0.13}{0.06}
        & \val{0.29}{0.14}
        & \val{0.18}{0.06}\\
    && BFS\textsubscript{corr} 
        & \val{24.00}{6.24}
        & \val{(42.20, 42.20)}{(6.53, 6.53)}
        & \val{0.17}{0.12}
        & \val{0.28}{0.11}
        & \val{0.21}{0.11}\\
    && \cellcolor{highlight} \ours 
	    & \valb{10.60}{2.13} \cellcolor{highlight}
	    & \valb{(34.80, 35.20)}{(5.56, 5.84)} \cellcolor{highlight}
	    & \valb{0.97}{0.04} \cellcolor{highlight}
	    & \valb{0.43}{0.06} \cellcolor{highlight}
        & \valb{0.60}{0.06} \cellcolor{highlight}\\
	\bottomrule
\end{tabular} 	}
\end{table*}
\subsection{Parameter Estimation}
\label{sec:exp_parameter_estimation}
The datasets used in \cref{sec:exp_structure_learning} do not contain latent confounders in their causal structures. Consequently, parameter estimation reduces to standard maximum likelihood estimation, which \ours (\cref{sec:method_parameter_estimation}) replicates by design. To meaningfully evaluate \ours in the presence of latent confounders, we consider a real-world dataset: the {\sc red-wine quality} from \citet{cortez2009modeling} (details in \cref{app:dataset_details}).

For the {\sc red-wine quality}, \ours's structure learning algorithm predicts a latent confounder between variables \textit{quality} and \textit{density}. Applying DirectLiNGAM on the full dataset indicates that the true confounder is \textit{alcohol content}, aligning with domain knowledge. Following \cref{sec:method_lm_causal_discovery}, we query an LM to identify potential latent confounders using world knowledge. \cref{fig:confounder_prediction} shows a histogram of LM predictions, with \textit{alcohol\_content} emerging as the top candidate. We incorporate this LM-provided confounder into \ours's parameter estimation pipeline and query the LM for its marginal distribution. \cref{tbl:confounder_variable_prediction} presents the LM-provided Gaussian distributions. Notably, when queried with obscure variables (\eg (cat, mouse)), LM often defaults to unit Gaussian $\N(0, 1)$.

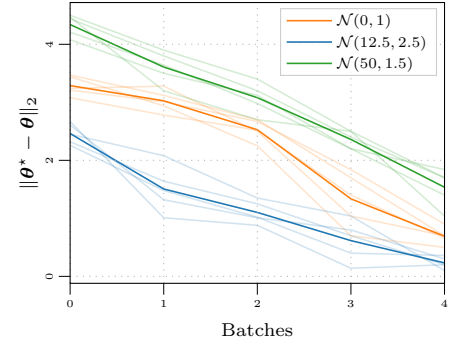
\begin{wrapfigure}{r}{0.35\textwidth}
	\raggedleft\scriptsize
	\setlength{\figurewidth}{.3\textwidth}
	\setlength{\figureheight}{.75\figurewidth}
	\pgfplotsset{axis on top,scale only axis,width=\figurewidth,height=\figureheight, ylabel near ticks,ylabel style={yshift=-2pt},xlabel style={yshift=3pt},y tick label style={rotate=90}, legend style={nodes={scale=0.8, transform shape}},tick label style={font=\tiny,scale=.8}}
	\pgfplotsset{legend cell align={left},every axis/.append style={legend style={draw=none,inner xsep=2pt, inner ysep=0.5pt, nodes={inner sep=2pt, text depth=0.1em},fill=white,fill opacity=0.8}}}
	\pgfplotsset{grid style={dotted, gray}}	
\begin{tikzpicture}

\definecolor{darkgray176}{RGB}{176,176,176}
\definecolor{darkorange25512714}{RGB}{255,127,14}
\definecolor{forestgreen4416044}{RGB}{44,160,44}
\definecolor{lightgray204}{RGB}{204,204,204}
\definecolor{steelblue31119180}{RGB}{31,119,180}

\begin{axis}[
legend cell align={left},
legend style={fill opacity=0.8, draw opacity=1, text opacity=1, draw=lightgray204},
tick align=outside,
tick pos=left,
width=\figurewidth,
height=\figureheight,
x grid style={darkgray176},
xmajorgrids,
xmin=0, xmax=4,
xtick style={color=black},
y grid style={darkgray176},
ymajorgrids,
xlabel=Batches,
ylabel=$\|\vtheta^\star - \vtheta\|_2$,
ymin=-0.12, ymax=4.72,
ytick style={color=black}
]
\addplot [semithick, darkorange25512714, opacity=0.2, forget plot]
table {%
0 3.24
1 3.28
2 2.5
3 1.04
4 0.7
};
\addplot [semithick, darkorange25512714, opacity=0.2, forget plot]
table {%
0 3.44
1 2.94
2 2.25
3 0.7
4 0.5
};
\addplot [semithick, darkorange25512714, opacity=0.2, forget plot]
table {%
0 3.08
1 2.78
2 2.52
3 1.4
4 0.66
};
\addplot [semithick, darkorange25512714, opacity=0.2, forget plot]
table {%
0 3.47
1 3.12
2 2.7
3 1.7
4 0.7
};
\addplot [semithick, darkorange25512714, opacity=0.2, forget plot]
table {%
0 3.21
1 3.01
2 2.67
3 1.84
4 0.9
};
\addplot [semithick, steelblue31119180, opacity=0.2, forget plot]
table {%
0 2.45
1 2.08
2 1.35
3 1.04
4 0.1
};
\addplot [semithick, steelblue31119180, opacity=0.2, forget plot]
table {%
0 2.32
1 1.64
2 1.25
3 0.7
4 0.2
};
\addplot [semithick, steelblue31119180, opacity=0.2, forget plot]
table {%
0 2.26
1 1.48
2 1.02
3 0.4
4 0.36
};
\addplot [semithick, steelblue31119180, opacity=0.2, forget plot]
table {%
0 2.61
1 1.32
2 1.01
3 0.8
4 0.3
};
\addplot [semithick, steelblue31119180, opacity=0.2, forget plot]
table {%
0 2.67
1 1.01
2 0.88
3 0.14
4 0.2
};
\addplot [semithick, forestgreen4416044, opacity=0.2, forget plot]
table {%
0 4.5
1 3.9
2 3.4
3 2.5
4 1.04
};
\addplot [semithick, forestgreen4416044, opacity=0.2, forget plot]
table {%
0 4.44
1 3.8
2 3.2
3 2.4
4 1.7
};
\addplot [semithick, forestgreen4416044, opacity=0.2, forget plot]
table {%
0 4.08
1 3.5
2 3.12
3 2.2
4 1.4
};
\addplot [semithick, forestgreen4416044, opacity=0.2, forget plot]
table {%
0 4.47
1 3.2
2 2.7
3 2.5
4 1.7
};
\addplot [semithick, forestgreen4416044, opacity=0.2, forget plot]
table {%
0 4.21
1 3.64
2 2.98
3 2.2
4 1.84
};
\addplot [semithick, darkorange25512714]
table {%
0 3.288
1 3.026
2 2.528
3 1.336
4 0.692
};
\addlegendentry{$\N(0, 1)$}
\addplot [semithick, steelblue31119180]
table {%
0 2.462
1 1.506
2 1.102
3 0.616
4 0.232
};
\addlegendentry{$\N(12.5, 2.5)$}
\addplot [semithick, forestgreen4416044]
table {%
0 4.34
1 3.608
2 3.08
3 2.36
4 1.536
};
\addlegendentry{$\N(50, 1.5)$}
\end{axis}

\end{tikzpicture}
\\[-.8em]
	\caption{\textbf{Parameter Estimation:} Convergence of parameters and robustness to prior misspecification as more batches are processed.}
	\label{fig:parameter_estimation}
	\vspace*{-1em}
\end{wrapfigure}

To assess parameter recovery, we track evolution of $\vtheta$ over sequential batches. As a performance metric, we compute the $\ell_2$-norm error $\|\vtheta^\star - \vtheta\|_2$, where $\vtheta^\star$ denotes the parameters obtained via MLE assuming the confounder (\textit{alcohol\_content}) is observed. \cref{fig:parameter_estimation} visualizes this convergence behavior.\looseness-1

\paragraph{Parameter estimation: robustness and recovery}
We demonstrate robustness of the proposed parameter estimation algorithm under misspecified or ill-informed priors. Based on domain knowledge and observational data, we estimate latent confounder \textit{alcohol\_content} to follow distribution $\N(11, 1.0)$. To stress-test \ours, we experiment with three alternative priors: $\N(12.5, 2.5)$ (suggested by GPT-3.5\textsubscript{turbo}), $\N(0, 1)$, and a severely misspecified prior $\N(50, 1.5)$.
\cref{fig:parameter_estimation} illustrates the evolution of the learned parameters $\vtheta$ across training batches for each prior. As expected and aligned with the Bayesian principle, convergence is slower when initialized with an inaccurate prior. Nevertheless, the model progressively refines its estimates as more data is processed, ultimately converging towards $\vtheta^\star$. This demonstrates both the robustness and recovery capabilities of \ours's Bayesian parameter estimation algorithm---even under poor initialization.
Additionally, we incorporate LM-suggested Pearson correlation coefficients in the \textit{M-step} objective to further guide estimation, \cref{eq:m_step} (see discussion and results in \cref{app:parameter_estimation}.).

\begin{figure*}[t!]
	\centering
	\begin{minipage}[t]{0.6\textwidth}
		\centering
		\footnotesize
		\captionof{table}{\textbf{LM-predicted priors for confounding variable:} LM suggest relevant Gaussian priors when the confounder is meaningful and default to $\N(0, 1)$ when uncertain.}
		\label{tbl:confounder_variable_prediction}
		\setlength{\tabcolsep}{2.5pt}
		\resizebox{\textwidth}{!}{%
			\begin{tabular}{lcc}
				\toprule
				\textbf{Variables} & \textbf{GPT-3.5\textsubscript{turbo}} & \textbf{GPT-4o}\\
				\midrule
				density$\leftarrow$ alcohol $\rightarrow$ quality & $\N(12.5, 2.5)$ & $\N(10.5, 1.2)$  \\
				density $\leftarrow$ alcohol $\rightarrow$ volatile-acidity & $\N(12.5, 2.5)$ & $\N(10.5, 1.2)$  \\
				cat $\leftarrow$ alcohol $\rightarrow$ mouse & $\N(0, 1)$ & $\N(0, 1)$ \\
				bed $\leftarrow$ alcohol $\rightarrow$ shopping & $\N(0, 1)$ & $\N(0, 1)$ \\
				\bottomrule
				\vspace*{1em}
		\end{tabular}}
	\end{minipage}
	\hfill		
	\begin{minipage}[t]{.38\textwidth}
		\vspace*{0pt}
		\raggedleft
		\tiny
		\begin{tikzpicture}
			\begin{axis}[
				ybar,
				bar width=0.3cm,
				width=\textwidth,
				height=.65\textwidth,
				enlarge x limits=0.1,
				ylabel={},
				ylabel={Count},
				symbolic x coords={
					alcohol\_content,
					grape\_quality,
					grape\_maturity,
					grape\_ripeness,
					grape\_type,
					wine\_age,
					sugar\_content,
					acidity\_level,
					vineyard,
					Other
				},
				xtick=data,
				xtick pos=bottom,
				x tick label style={rotate=45, anchor=east},
				ylabel style={yshift=-2em},
				nodes near coords,
				ymin=0,
				ymax=35
				]
				\addplot coordinates {
					(alcohol\_content, 27)
					(grape\_quality, 22)
					(grape\_maturity, 8)
					(grape\_ripeness, 6)
					(grape\_type, 4)
					(wine\_age, 5)
					(sugar\_content, 4)
					(acidity\_level, 4)
					(vineyard, 4)
					(Other, 9)
				};
			\end{axis}
		\end{tikzpicture}
		\captionof{figure}{LM predicted confounders.}
		\label{fig:confounder_prediction}
	\end{minipage}
	\vspace*{-1em}
\end{figure*}	

\section{Discussion and Conclusion}
We present \ours (Noisy Language Prior in Sequential Causal Modeling)---a Bayesian-inspired framework for causal structure discovery and parameter (edge weights) estimation in sequential, batch-wise data settings. By treating language models (LMs) as noisy surrogate experts, \ours addresses the dual \emph{LM-induced} and \emph{data-induced} biases. A key contribution is the representation shift from DAGs to PAGs, allowing uncertainty and confounders to be modeled explicitly. Through LM interactions modeled in sequential optimization framework and EM style iterative parameter estimation algorithm, \ours improves both structural accuracy and parameter recovery in hybrid \textit{data-LM} pipelines. \ours leverages global LM knowledge while staying grounded in local data, and offers a robust foundation for hybrid causal discovery in the presence of sequential, batched data. 

\paragraph{Limitations and Future Work} 
In the Bayesian formulation in \ours, incorporating Dirichlet or hierarchical Bayesian priors over LM judgments could provide another way to capture LM uncertainty.
Extending to fully probabilistic inference over graph structures, rather than edge-level belief tracking, is another promising direction.
Future work may also explore adaptive calibration of LM responses or memory-based accumulation of observational data across batches, following \citet{pmlr-v202-chang23a}. 
We employ FCI for causal discovery due to its compatibility with our setup (Section~\ref{sec:problem_statement}). The current framework assumes that each latent confounder may affect two observed variables; extending to multi-variable confounding is a direction for future work. Systematically evaluating the robustness of \ours for other causal discovery algorithms remains important future work. Additionally, extending our approach to incorporate interventional data and active learning strategies could improve sample efficiency and the quality of discovered causal structures. Systematic evaluation of LM accuracy in confounder identification across diverse domains also remains a useful future effort. Finally, a theoretical analysis of sequential optimization and a bandit-style framework remains an open problem, discussed preliminarily in \cref{app:sequential_optimization}.

\paragraph{Broader Impact} 
\ours combines observational data with LM-derived knowledge to discover causal structures, which may inform downstream decisions in domains such as healthcare, policy, and business. As with any causal discovery method, the learned structures reflect the assumptions and limitations of the underlying algorithms, data quality, and in our case, the reliability of LM priors. We recommend that practitioners treat discovered structures as hypotheses to be validated through domain expertise or interventional studies before acting on them.
\section*{Acknowledgments}
This work was primarily conducted while PV was an intern at Adobe Research, Bangalore. PV and AS acknowledge funding from the Research Council of Finland (grants 362408, 339730). We acknowledge CSC-IT Center for Science, Finland, and the Aalto Science-IT project for the computational resources.

\bibliographystyle{tmlr}

\clearpage

\setcounter{figure}{0}
\renewcommand\thefigure{A\arabic{figure}}
\setcounter{table}{0}
\renewcommand{\thetable}{A\arabic{table}}
\setcounter{equation}{0}
\renewcommand{\theequation}{A\arabic{equation}}
\setcounter{algorithm}{0}
\renewcommand{\thealgorithm}{A\arabic{algorithm}}

\appendix
\section*{Appendix}
We organize the appendix as follows: \cref{app:dataset_details} provides detailed descriptions of all datasets used and simulation details; \cref{app:other_causal_discovery_algorithms} discusses the choice of FCI over other causal discovery algorithms; \cref{app:metrics} outlines the evaluation metrics employed to assess method performance; \cref{app:sequential_optimization} discusses the connection between the proposed sequential optimization approach and the multi-armed bandit framework; \cref{app:parameter_estimation} describes the parameter estimation algorithm and its robustness to LM-derived information; and \cref{app:experiment_details} presents the experimental setup along with additional implementation and hyperparameter details.\looseness-1

\section{Dataset and Simulation Details}
\label{app:dataset_details}

\paragraph{\sc Earthquake}
The {\sc Earthquake} dataset is a widely used synthetic benchmark in causal discovery. It comprises a small, acyclic, and semantically meaningful graph over five binary variables: \textit{Burglary}, \textit{Earthquake}, \textit{Alarm}, \textit{JohnCalls}, and \textit{MaryCalls}. In this structure, an earthquake or burglary can trigger the alarm, which in turn influences whether John or Mary calls. The ground-truth causal graph is depicted in \cref{fig:earthquake_true_graph}. For our experiments, we aggregate data from six CSV files (\texttt{earthquake\_250\_{1--6}.csv}) available at \url{https://github.com/andrewli77/MINOBS-anc/}, resulting in a combined dataset of 1500 samples. As the dataset contains Boolean values (Yes/No) we convert them to a numerical format (1/0) producing the full dataset $\R^{(1500, 5)}$.

\paragraph{\sc Asia}  
The {\sc Asia} dataset is another canonical benchmark in causal discovery and probabilistic inference. It consists of eight binary variables (\eg \textit{VisitToAsia}, \textit{Tuberculosis}, \textit{Smoking}, \textit{LungCancer}) structured in a directed acyclic graph (DAG) with known semantics, as shown in \cref{fig:asia_true_graph}. We construct the dataset by merging all samples from the six CSV files (\texttt{asia\_250\_{1--6}.csv}) provided at \url{https://github.com/andrewli77/MINOBS-anc/}, yielding a total of 1500 samples. As with the {\sc Earthquake} dataset, all boolean values are converted to numerical format producing the full dataset $\R^{(1500, 8)}$.

\paragraph{\sc user level data - I} 
The {\sc User Level Data - I} dataset is derived from the publicly available Google Analytics Sample Dataset, accessible via Google BigQuery (detailed in \citep{googleanalytics}). It captures real-world e-commerce interactions from the Google Merchandise Store. We focus on a curated subset of user-level features that reflect engagement, browsing behavior, and purchase intent. The selected variables include \textit{Proximity to Transaction}, \textit{Number of Add To Cart}, \textit{Number of Product Clicks}, \textit{Number of Sessions on iOS}, \textit{Number of Promo Hits (Android)}, \textit{Number of Cheap Products Viewed}, \textit{Number of Page Hits}, \textit{Time Spent per Session}, and \textit{Number of Promo Hits (Others)}.
Since the true causal structure is not available, we estimate it using the DirectLiNGAM algorithm applied to the observational data, which also provides edge weights representing causal strengths. We then simulate data from this estimated DAG using a structural equation model (SEM) with linear relationships and additive Gaussian noise. \cref{fig:user_data_1_true_graph} shows the causal structure with strengths estimated by DirectLiNGAM which acts as a ground truth for us.

\paragraph{\sc user level data - II} 
The {\sc User-Level Data II} dataset is constructed from the same Google Analytics Sample corpus as {\sc User-Level Data I}, but emphasizes a distinct set of behavioral features associated with user engagement. The selected variables include: \textit{Number of Hits}, \textit{Number of Unique URLs}, \textit{Time Spent}, \textit{Total Active Days Last Month}, \textit{Number of Sessions Last Month}, \textit{Number of Hits Last Month}, \textit{Number of Page Hits Last Month}, and \textit{Number of Hits on Social Network}. These features capture both cumulative and recent patterns of user interaction, providing a rich view of user behavior. Since the true causal structure is not available, as with {\sc User-Level Data I}, we estimate the underlying causal structure using DirectLiNGAM and simulate data from the resulting DAG using a linear SEM with additive Gaussian noise. This serves as the true causal structure for subsequent evaluation and experimentation. \cref{fig:user_data_2_true_graph} shows causal structure with strengths estimated by DirectLiNGAM which acts as a ground truth for us.

\paragraph{\sc Child}
The {\sc Child} dataset is a widely used synthetic benchmark in causal discovery. It comprises a acyclic, and semantically meaningful graph over 20 variables: \textit{Age, Birth Asphyxia, Disease Type, Sickness, Hypoxia in O\textsubscript{2}, Grunting, Lung Parenchyma, Lower Body O\textsubscript{2}, CO\textsubscript{2}, Chest X-Ray, LVH, Cardiac Mixing, Birth Defect, Pulmonary Stenosis, Duct flow, Cyanosis, Heart Disorder, Temperature, Heart-Rate}, and \textit{PV SAT}. The ground-truth causal graph is shown in \citet[Figure 2]{spiegelhalter1993bayesian}. For our experiments, we aggregate data from six CSV files (\texttt{child\_2000\_{1--6}.csv}) available at \url{https://github.com/andrewli77/MINOBS-anc/}, resulting in a combined dataset of 12000 samples. We convert all the variables to a categorical format producing the full dataset $\R^{(12000, 20)}$.

\paragraph{\sc Alarm}
The {\sc Alarm} dataset is a widely used synthetic benchmark in causal discovery. It comprises a acyclic, and semantically meaningful graph over 37 variables. For our experiments, we aggregate data from the CSV files (\texttt{alarm\_1000\_{1--6}.csv}) available at \url{https://github.com/andrewli77/MINOBS-anc/}. All the variables are converted to categorical format and the details about it can be found in \citet{beinlich1989alarm}.

\paragraph{\sc Red Wine Quality}  
The {\sc Red Wine Quality} dataset is a real-world dataset from the UCI Machine Learning Repository, comprising 11 physicochemical attributes of red wine samples—such as \textit{density}, \textit{alcohol}, and \textit{residual sugar}—along with a quality rating. We utilize this dataset for \textit{parameter estimation} experiments with linear SEMs \ie of the form $A = \theta_1 B + \epsilon_i$ where $\epsilon_i\sim \N(0, \sigma^2)$. As shown in \cref{fig:redwine_true_graph}, the variable \textit{alcohol} acts as a confounder between \textit{density} and \textit{quality}, making it suitable for parameter estimation evaluation. The dataset used in our experiments is sourced from the CMU Example Causal Datasets repository: \url{https://github.com/cmu-phil/example-causal-datasets/blob/main/real/wine-quality/data/winequality-red.continuous.txt}. It contains a total of 1599 samples.

\begin{table*}[t!]
	\centering
	\footnotesize
	\caption{Glossary of notations used throughout the paper, categorized by their role in datasets, causal structures, parameters, expert interactions, and other components.}
	\label{tbl:notations}
	\resizebox{\textwidth}{!}{
\begin{tabular}{@{}ll ll ll@{}}
	\toprule
	\textbf{Symbol} & \textbf{Description} &
	\textbf{Symbol} & \textbf{Description} &
	\textbf{Symbol} & \textbf{Description} \\
	\midrule
	\multicolumn{6}{@{}l}{\textbf{Dataset and Observations}} \\
	$ \mathbb{D}$ & True data distribution & 
	$\mathcal{D}$ & Full observed dataset & 
	$\DD_i$ & Observed data at batch $i$ \\
	$n_i$ & Number of data points in batch $i$ & 
	$\VD$ & Set of observed variables & 
	$d$ & Cardinality of $\VD$ \\
	$\VL$ & Unobserved (latent) variables & 
	$k$ & Look-back batch window size & 
	$\B_i$ & Background knowledge till batch $i$ \\
	\midrule
	\multicolumn{6}{@{}l}{\textbf{Causal Graphs and Structures}} \\
	$\G$ & True underlying causal graph & 
	$\V, \E$ & Nodes and edges in $\G$ & 
	$\GD{i}$ & Causal graph after batch $i$ \\
	$\VD, \ED{i}$ & Nodes/edges in $\GD{i}$ & 
	$\GE{i}$ & Expert-provided graph till batch {i} & 
	$\VE{i}, \EE{i}$ & Nodes and edges in $\GE{i}$ \\
	$A, B$ & Arbitrary variables in graph & 
	$L$ & Latent confounder in causal structure& 
	&  \\
	\midrule
	\multicolumn{6}{@{}l}{\textbf{Parameters and Models}} \\
	$\vtheta$ & Edge weights of the causal DAG & 
	$\vthetaO, \vthetaL$ & Edge weights for observed, latent variable & 
	$\vphi$ & SEM parameters $\{\vtheta, \sigma^2\}$ \\
	$\theta_{A}$ & Edge weights associated with $A$ & 
	$\vphi_{A}$ & Parameters for $A$ variable ($\theta_A, \sigma^2$) & 
	$\epsilon$ & SEM noise \\
	$\sigma^2$ & SEM noise variance & 
	$\rho(A, B)$ & Correlation value between $A, B$ & 
	& \\
	\midrule
	\multicolumn{6}{@{}l}{\textbf{Expert Interaction and Histograms}} \\
	$\alpha$ & Expert noise level & 
	$\HE{i}$ & Edge histogram till batch $i$ & 
	$\HL{i}$ & Latent variable histogram till batch $i$ \\
	$\IE, \IL$ & Instructions for edge/latent expert & 
	$\IP, \IR$ & Instructions for prior/correlation expert. & 
	$\me, \ml$ & Maximum expert calls for edge/latent \\
	\midrule
	\multicolumn{6}{@{}l}{\textbf{Others}} \\
	$\R$ & Set of real numbers & 
	$\N$ & Gaussian distribution & 
	$\eta$ & Learning rate \\
	$\lambda$ & Regularizer parameter & 
	$\vm_p$ & Prior mean & 
	$\MS_p$ & Prior covariance \\
	\bottomrule
\end{tabular}
 	}
	\vspace*{-.5em}
\end{table*}

\section{Comparisons with Other Causal Discovery Algorithms}
\label{app:other_causal_discovery_algorithms}
In this paper, we focus on a setting where both \emph{selection bias} and \emph{latent confounding} are present. To accommodate both aspects in this general setting, it is necessary to shift from DAGs to Partial Ancestral Graphs (PAGs), since this explicitly captures uncertainty due to latent confounders and selection bias. The Fast Causal Inference (FCI) algorithm naturally aligns with these assumptions, making it a principled choice for the proposed framework, \ours.

Moreover, the primary aim of this paper is not to benchmark causal structure discovery algorithms, but rather to demonstrate how \ours can refine causal structure over time by integrating noisy expert knowledge (from LMs) with observational data. Notably, \ours is designed to be algorithm-agnostic, and can be used in a plug-and-play manner with any PAG-generating causal discovery method. Below, we briefly discuss why several modern algorithms are not directly applicable:

\begin{enumerate}
	\item \textbf{Greedy Equivalence Search (GES):} GES proposed by \citet{chickering2002optimal} assumes both causal sufficiency and the absence of selection bias. These assumptions do not hold in our sequential setting, where both latent confounders and selection bias may be present.
	\item \textbf{DAG-NoCURL and DAG-NoTEARS:} DAG-NOCURL by  \citet{pmlr-v139-yu21a} and DAG-NoTEARS by \citet{NEURIPS2018_e347c514} are continuous optimization-based methods that output DAGs under the assumption of causal sufficiency, without accounting for latent confounders. Thus, they do not apply to our setting, which requires a PAG-based representation to model uncertainty.
	\item \textbf{DAGMA:} \citet{NEURIPS2022_36e2967f} proposed DAGMA: a score-based causal discovery method that also formulates the problem as a continuous optimization task. Like DAG-NoCURL and DAG-NoTEARS, it outputs deterministic DAGs and assumes no latent confounding, making it incompatible with the sequential setting.
	\item \textbf{LiNGAM:} LiNGAM \citep[Linear Non-Gaussian Acyclic Model,][]{shimizu2014lingam} assumes that all relevant variables are observed, \ie there are no hidden confounders. This causal sufficiency assumption renders it unsuitable for the sequential setting.
	\item \textbf{Recursive Causal Discovery (RCD):} Unlike FCI, RCD proposed by \citet{pmlr-v108-maeda20a} produces a DAG rather than a PAG, focusing on efficient DAG learning. Hence, when dealing with latent confounders and selection bias requiring PAG representation, RCD can not be employed.
\end{enumerate}

We note that enhanced variants of FCI such as RFCI, GFCI, and FCI+ are fully compatible with \ours and can be seamlessly integrated bringing their advantage to \ours as well. We keep the research to study convergence impact of different algorithms for future work.

\section{Evaluation Metrics}
\label{app:metrics}
We describe the evaluation metrics used to assess the performance of the causal discovery method.

\subsection{Modified Structural Hamming Distance (Mod.\ SHD)}
To evaluate the structural accuracy of methods, we use a modified version of the Structural Hamming Distance (SHD) tailored for Partial Ancestral Graphs (PAGs). Unlike standard causal graphs, PAGs include multiple edge types that reflect varying degrees of causal certainty. This necessitates a more nuanced treatment of structural differences.

For example, consider the true edge $A \rightarrow B$. If one predicted PAG contains $A \leftrightarrow B$ (a definite bidirected edge) and another contains $A \dotarrow B$ (an edge with uncertainty at one endpoint), the former should incur a higher penalty since it reflects stronger, incorrect causal commitment. That is,
\begin{equation}
	\text{Mod.\ SHD} (A \rightarrow B, A \leftrightarrow B) > \text{Mod.\ SHD}(A \rightarrow B, A \dotarrow B) \, .
\end{equation}

Standard SHD counts the total number of missing edges, extra edges, and orientation mismatches. Our modified SHD refines the orientation mismatch term by assessing the endpoints of each edge separately. We assign a penalty of $1.0$ for definite orientation errors (\eg $\rightarrow$ vs.\ $\leftarrow$) and $0.5$ for uncertain mismatches (\eg $\rightarrow$ vs.\ $\dotarrow$). This weighting scheme better reflects the confidence associated with different edge types and penalizes definitive errors more heavily than uncertainty.

\subsection{Structural Intervention Distance (SID)}
The Structural Intervention Distance (SID) measures the robustness of a learned causal graph in supporting correct interventional reasoning. Unlike purely structural metrics such as SHD, SID evaluates whether the predicted graph implies the correct set of causal effects under interventions.

Formally, SID counts the number of intervention targets for which the predicted graph induces an incorrect adjustment set relative to the true graph. An SID of zero indicates that the learned structure supports identically correct interventional distributions as the ground truth, even if some edge directions are incorrect but do not affect adjustment validity.

SID is particularly useful in settings where downstream causal queries, rather than exact graph recovery, are the primary concern. It captures whether the learned structure preserves the correct set of (in)dependencies required for estimating interventional effects, making it a practical and task-aligned evaluation metric for causal discovery.

Since SID is not directly defined for PAGs, we compute it by mapping the learned PAG into a CPDAG-compatible adjacency representation. Directed edges are preserved, while partially oriented or bidirected edges are encoded as undirected or partially specified edges. We then use the \textit{SID\_CPDAG} function from the R SID package, which computes (lower, upper) bounds on SID between a true DAG and an equivalence class. When the bounds coincide, the learned structure fully determines interventional distances. We note that this mapping is an approximation, as PAGs represent equivalence classes under latent confounding that are more general than CPDAGs, and the reported SID values should be interpreted as approximate bounds.
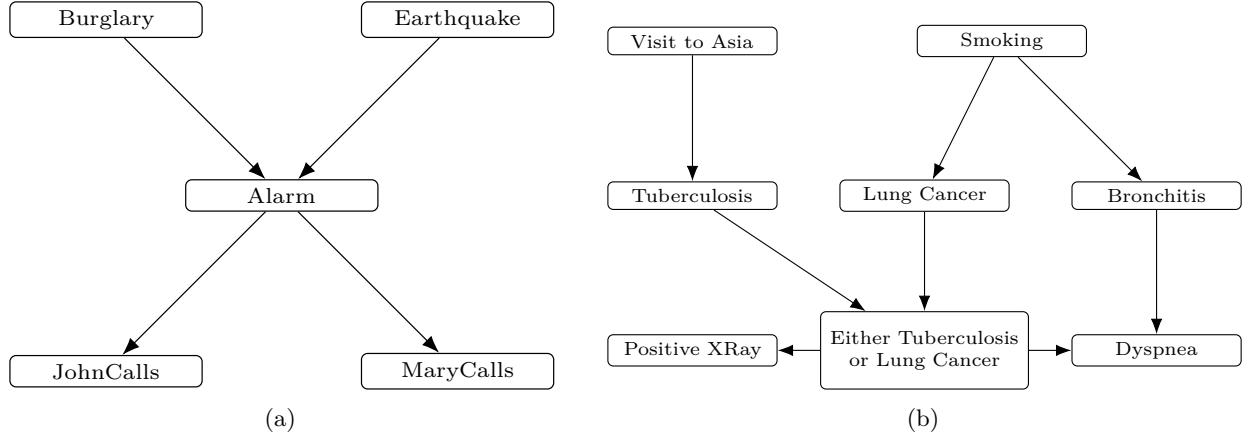
\begin{figure*}[t!]
	\centering
	\scriptsize
	\begin{subfigure}{.45\textwidth}
		\resizebox{\textwidth}{!}{		
\begin{tikzpicture}[
	node/.style={draw, rounded corners=2pt, align=center, font=\scriptsize, text width=2cm},
arrow/.style={-{Latex[length=2mm]}, thin}
	]

	\node[node] (Burglary) at (0,2) {Burglary};
	\node[node] (Earthquake) at (4,2) {Earthquake};
	\node[node] (Alarm) at (2,0) {Alarm};
	\node[node] (John) at (0,-2) {JohnCalls};
	\node[node] (Mary) at (4,-2) {MaryCalls};

	\draw[arrow] (Burglary) -- (Alarm);
	\draw[arrow] (Earthquake) -- (Alarm);
	\draw[arrow] (Alarm) -- (John);
	\draw[arrow] (Alarm) -- (Mary);
	
\end{tikzpicture} 		}
		\caption{}
		\label{fig:earthquake_true_graph}		
	\end{subfigure}
	\hfill
	\begin{subfigure}{.52\textwidth}
		\resizebox{\textwidth}{!}{
\begin{tikzpicture}[
	node/.style={draw, rounded corners=2pt, align=center, font=\scriptsize, text width=2cm},
arrow/.style={-{Latex[length=2mm]}, thin}
	]

	\node[node] (Asia) at (0,4) {Visit to  Asia};
	\node[node] (Smoking) at (4,4) {Smoking};
	\node[node] (Tuberculosis) at (0,2) {Tuberculosis};
	\node[node] (LungCancer) at (3,2) {Lung Cancer};
	\node[node] (Bronchitis) at (6,2) {Bronchitis};
	\node[node] (Either) at (3,0) [draw, rectangle, minimum size=1cm, align=center, text width=2.5cm] 
{Either Tuberculosis\\or Lung Cancer};
	\node[node] (XRay) at (0, 0) {Positive XRay};
	\node[node] (Dyspnea) at (6,0) {Dyspnea};

	\draw[arrow] (Asia) -- (Tuberculosis);
	\draw[arrow] (Smoking) -- (LungCancer);
	\draw[arrow] (Smoking) -- (Bronchitis);
	\draw[arrow] (Tuberculosis) -- (Either);
	\draw[arrow] (LungCancer) -- (Either);
	\draw[arrow] (Either) -- (XRay);
	\draw[arrow] (Either) -- (Dyspnea);
	\draw[arrow] (Bronchitis) -- (Dyspnea);
	
\end{tikzpicture} 		}
		\caption{}
		\label{fig:asia_true_graph}
	\end{subfigure}
	\caption{\textbf{Causal Structure Learning Datasets} \textit{(a)}~Ground-truth causal graph for the \textsc{Earthquake} dataset, where an earthquake or burglary can trigger an alarm, which in turn causes calls from John and Mary; 
		\textit{(b)}~Ground-truth structure for the \textsc{Asia} dataset, a classic dataset illustrating causal relations between variables such as smoking, lung disease, and visits to Asia.}
	\label{app:true_asia_earthquake_graph}
	\vspace*{-1em}
\end{figure*}
\subsection{Precision, Recall, F1-Score}
In addition to structural metrics, we report standard classification metrics---Precision, Recall, and F1-Score---computed over predicted directed causal relationships of the form \( A \rightarrow B \).

A predicted edge \( A \rightarrow B \) is considered a true positive if it matches a directed edge in the ground-truth causal graph. Precision measures the fraction of predicted directed edges that are correct, while Recall quantifies the fraction of ground-truth directed edges that are successfully recovered. F1-Score is the harmonic mean of Precision and Recall, providing a balanced summary of accuracy and completeness:\looseness-1
\begin{align}
	\text{Precision} &= \frac{|\text{Correctly predicted directed edges } A \rightarrow B|}{|\text{Predicted directed edges } A \rightarrow B|} \,, \\
	\text{Recall} &= \frac{|\text{Correctly predicted directed edges } A \rightarrow B|}{|\text{Ground-truth directed edges } A \rightarrow B|} \,, \\
	\text{F1 Score} &= 2 \cdot \frac{\text{Precision} \cdot \text{Recall}}{\text{Precision} + \text{Recall}} \,.
\end{align}
These metrics focus exclusively on directed edges and ignore uncertain or undirected edge types. As such, they offer a more task-specific evaluation of causal directionality recovery, which is critical in downstream applications that rely on explicit causal mechanisms.

\section{Sequential Optimization as Multi-Armed Bandit}
\label{app:sequential_optimization}
The LM interaction in \ours proposed as sequential optimization (\cf \cref{sec:method_lm_interaction}), can presumably be interpreted as a \textit{stochastic multi-armed bandit (MAB)} problem, providing a useful lens for understanding the design of the proposed score selection strategy (\cref{eq:score}).

\paragraph{From Edge Querying to Bandits}
In MAB problems, a learner chooses among $K$-\emph{arms}, each associated with an unknown reward distribution, and aims to maximize cumulative reward over $T$ pulls by balancing \textit{exploration} (learning about uncertain arms) and \textit{exploitation} (favoring known high-reward arms).

In our setting (as discussed in \cref{eq:seqential_decisions}):
\begin{itemize}
	\item Each candidate edge $e$ is treated as an \textbf{arm}.
	\item Pulling arm $e$ means \textbf{querying the LM} about edge $e$'s causal type.
	\item \textbf{Reward} is the \textbf{information gain} from that query---specifically, the reduction in uncertainty over edge $e$’s type.
	\item Total LM query budget $m_L$ acts as the $T$ \textbf{horizon}.
\end{itemize}

This framing is motivated by the need to \textit{adaptively allocate} a limited number of LM queries to edges that are either currently uncertain (exploration) or close to being included in background knowledge (exploitation). As in bandits, where arms may yield noisy feedback, LM responses are noisy, that is, stochastic and depend on context and prompt. Hence, each edge must be queried multiple times to obtain a reliable posterior.
\begin{figure*}[t!]
	\centering
	\scriptsize
\begin{tikzpicture}[
	node/.style={draw, rounded corners=2pt, align=center, font=\scriptsize, text width=2.2cm},
	arrow/.style={-{Latex[length=2mm]}, thin}
	]

	\node[node] (P2T) at (-0.3, 2) {Proximity to\\Transaction};
	\node[node] (AddCart) at (3, 2) {Number of\\Add to Cart};
	\node[node] (ProdClick) at (0, -1) {Number of\\Product Clicks};
	\node[node] (SessionsiOS) at (6.5, 1.5) {Number of\\Sessions on iOS};
	\node[node] (PromoAndroid) at (6, 0.2) {Number of\\Promo Hits (Android)};
	\node[node] (CheapViewed) at (1.8, -3.5) {Number of \\Cheap Products\\Viewed};
	\node[node] (PageHits) at (6, -2.6) {Number of\\Page Hits};
	\node[node] (TimeSpent) at (9, -1) {Time Spent\\per Session};
	\node[node] (PromoOthers) at (9.5, -3) {Number of\\Promo Hits\\(Others)};

	\draw[arrow] (P2T) -- node[midway, fill=white, font=\tiny]{0.12} (AddCart);
	\draw[arrow] (AddCart) -- node[midway, fill=white, font=\tiny]{0.24} (ProdClick);
	\draw[arrow] (AddCart) -- node[midway, fill=white, font=\tiny]{0.18} (SessionsiOS);
	\draw[arrow] (AddCart) -- node[midway, fill=white, font=\tiny]{0.38} (PromoAndroid);
	\draw[arrow] (AddCart) -- node[midway, fill=white, font=\tiny]{0.52} (CheapViewed);
	\draw[arrow] (AddCart) -- node[midway, fill=white, font=\tiny]{0.40} (PageHits);
	
	\draw[arrow] (ProdClick) -- node[midway, fill=white, font=\tiny]{0.23} (SessionsiOS);
	\draw[arrow] (ProdClick) -- node[midway, fill=white, font=\tiny]{0.15} (PromoAndroid);
	\draw[arrow] (ProdClick) -- node[midway, fill=white, font=\tiny]{0.32} (CheapViewed);
	\draw[arrow] (ProdClick) -- node[midway, fill=white, font=\tiny]{0.15} (PageHits);
	
	\draw[arrow] (PromoAndroid) -- node[midway, fill=white, font=\tiny]{0.13} (CheapViewed);
	\draw[arrow] (PromoAndroid) -- node[midway, fill=white, font=\tiny]{0.32} (PageHits);
	\draw[arrow] (PromoAndroid) -- node[midway, fill=white, font=\tiny]{0.17} (TimeSpent);
	\draw[arrow] (PromoAndroid) -- node[midway, fill=white, font=\tiny]{0.63} (PromoOthers);
	
	\draw[arrow] (CheapViewed) -- node[midway, fill=white, font=\tiny]{0.25} (PageHits);
	\draw[arrow] (CheapViewed) -- node[midway, fill=white, font=\tiny]{0.29} (PromoOthers);
	
	\draw[arrow] (PageHits) -- node[midway, fill=white, font=\tiny]{0.65} (TimeSpent);
	\draw[arrow] (PageHits) -- node[midway, fill=white, font=\tiny]{0.11} (PromoOthers);
	
	\draw[arrow] (TimeSpent) -- node[midway, fill=white, font=\tiny]{-0.09} (PromoOthers);	
\end{tikzpicture} 	\caption{Ground truth causal structure for the {\sc user level data - I} illustrating the causal relationships between user behavioral metrics. The edge weights represent estimated causal strengths derived from observational data.}
	\label{fig:user_data_1_true_graph}
	\vspace*{-1em}
\end{figure*}
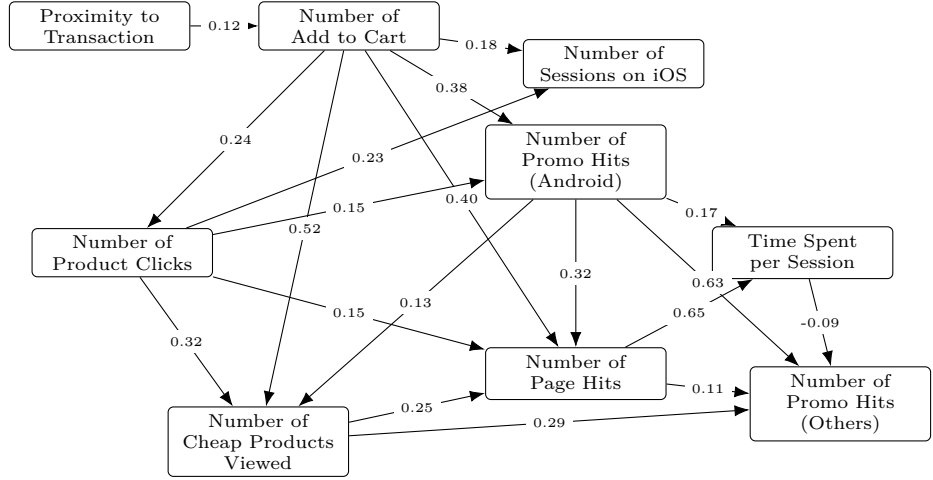
\subsection{UCB-Inspired Selection Score}
The selection score proposed in \cref{eq:score} resembles a classic \textit{Upper Confidence Bound (UCB)} strategy,
\begin{equation}
	S_i^e = w_1 E_i^e + w_2 \left(\frac{1}{TD_i^e}\right) + w_3 \sqrt{\frac{\log T_i}{T_i^e}}, \quad \text{s.t.} \quad TD_i^e = \tau_i - \max(\HE{i}(e)) \, .
\end{equation}
This selection score balances three factors:
\begin{itemize}
	\item \textbf{Uncertainty (Exploration):} $E^e_i$ is the entropy of the edge's histogram---capturing how uncertain the current belief over $e$ is. High entropy indicates high uncertainty, encouraging exploration.
	\item \textbf{Threshold Proximity (Exploitation):} $1/TDe_i$ quantifies how close edge $e$ is to being included in background knowledge. A small $TDe_i$ yields a large score, encouraging exploitation of promising edges.
	\item \textbf{Exploration Bonus:} The term $\sqrt{\log T_i / T^e_i}$ mirrors the optimism term in UCB1. It grows slowly with total queries and shrinks with more queries to edge $e$, thus favoring edges that have not been sampled much.
\end{itemize}

Together, this scoring function implements a principled query policy that prioritizes under-explored, informative, and nearly-threshold edges. Each LM query corresponds to a pull of the edge with the highest $S^e_i$, analogous to greedy selection in UCB-based bandits.

\subsection{Regret Interpretation}
Under idealized assumptions of \iid edge-level information gains, bounded support, and stationary LM behavior, the expected cumulative \textit{regret} after $T$ queries is:
\begin{equation}
	\mathcal{R}(T) = \sum_{i=1}^T (\mu^* - \mu_{a_i}) \, ,
\end{equation}
where $\mu^*$ is the maximum expected gain over all edges, and $a_i$ is the edge queried at step $i$. Classical UCB algorithms ensures that
\begin{equation}
	\mathcal{R}(T) = O\left(\sum_{e:\Delta_e > 0} \frac{\log T}{\Delta_e} \right), \quad \text{where } \Delta_e = \mu^* - \mu_e.
\end{equation}
This sublinear growth implies that the average regret $\mathcal{R}(T)/T \to 0$ as $T \to \infty$, \ie the learner asymptotically focuses on optimal arms.
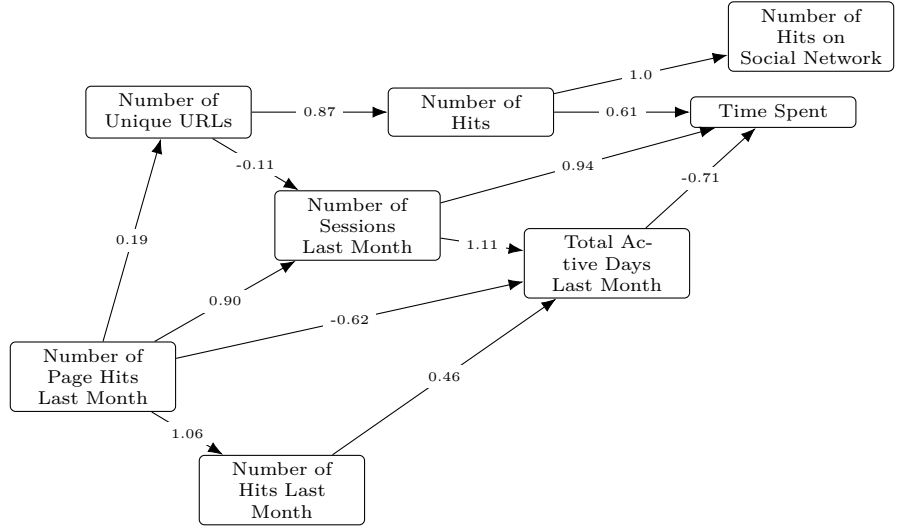
\begin{figure*}[t!]
	\centering
	\scriptsize
\begin{tikzpicture}[
	node/.style={draw, rounded corners=2pt, align=center, font=\scriptsize, text width=2cm},
	arrow/.style={-{Latex[length=2mm]}, thin}
	]

	\node[node] (UniqueURLs) at (-1.5, 2) {Number of\\Unique URLs};
	\node[node] (Hits) at (2.5, 2) {Number of\\Hits};
	\node[node] (TimeSpent) at (6.5, 2) {Time Spent};
	\node[node] (ActiveDays) at (4.3, 0) {Total Active Days\\Last Month};
	\node[node] (SessionsLastMonth) at (1, 0.5) {Number of\\Sessions\\Last Month};
	\node[node] (HitsLastMonth) at (0, -3) {Number of\\Hits Last Month};
	\node[node] (PageHitsLastMonth) at (-2.5, -1.5) {Number of\\Page Hits\\Last Month};
	\node[node] (SocialHits) at (7, 3) {Number of\\Hits on\\Social Network};

	\draw[arrow] (UniqueURLs) -- node[midway, fill=white, font=\tiny]{0.87} (Hits);
	\draw[arrow] (Hits) -- node[midway, fill=white, font=\tiny]{0.61} (TimeSpent);
	\draw[arrow] (ActiveDays) -- node[midway, fill=white, font=\tiny]{-0.71} (TimeSpent);
	\draw[arrow] (SessionsLastMonth) -- node[midway, fill=white, font=\tiny]{0.94} (TimeSpent);
	\draw[arrow] (PageHitsLastMonth) -- node[midway, fill=white, font=\tiny]{0.19} (UniqueURLs);
	\draw[arrow] (SessionsLastMonth) -- node[midway, fill=white, font=\tiny]{1.11} (ActiveDays);
	\draw[arrow] (HitsLastMonth) -- node[midway, fill=white, font=\tiny]{0.46} (ActiveDays);
	\draw[arrow] (PageHitsLastMonth) -- node[midway, fill=white, font=\tiny]{-0.62} (ActiveDays);
	\draw[arrow] (UniqueURLs) -- node[midway, fill=white, font=\tiny]{-0.11} (SessionsLastMonth);
	\draw[arrow] (PageHitsLastMonth) -- node[midway, fill=white, font=\tiny]{0.90} (SessionsLastMonth);
	\draw[arrow] (PageHitsLastMonth) -- node[midway, fill=white, font=\tiny]{1.06} (HitsLastMonth);
	\draw[arrow] (Hits) -- node[midway, fill=white, font=\tiny]{1.0} (SocialHits);
	
\end{tikzpicture}

 	\caption{Ground truth causal structure for the {\sc user level data - II} illustrating the causal relationships between user activity metrics such as hits, time spent, session history, and social network interactions. The edge weights represent estimated causal strengths derived from observational data.}
	\label{fig:user_data_2_true_graph}
	\vspace*{-1em}
\end{figure*}

Applied to \ours, this suggests that if LM responses were non-stochastic, the selection score strategy concentrates LM calls on most informative edges, making increasingly efficient use of limited expert budget. 
However, we \textbf{caution} against this linkage to regret bounds. The above analysis rests on assumptions that do not strictly hold in the proposed \ours setting due to:
\begin{itemize}
	\item \textbf{Stochasticity:} The LM’s behavior is batch-context dependent and evolves as context accumulates. Reward distributions (information gains) are not fixed.
	\item \textbf{Dependent arms:} Updates to the posterior of one edge can influence others due to graph constraints, violating arm independence.
	\item \textbf{Implicit priors:} LM outputs are influenced by prompts, prior batches, and temperature, introducing structured, non-\iid noise.
\end{itemize}

These violations of MAB assumptions mean that classical regret bounds cannot be directly applied. Still, the MAB abstraction provides valuable intuition for designing and analyzing selection policies under uncertainty.

\subsection{Future Work}
A full theoretical analysis of regret in this setting would require modeling \textit{non-stationary, dependent} reward structures. Promising directions include \textit{contextual and Bayesian bandits} that incorporate evolving priors, \textit{combinatorial bandits} for structured edge dependencies, and \textit{information-theoretic regret bounds} based on entropy reduction. We leave these extensions for future work.

\section{Parameter Estimation}
\label{app:parameter_estimation}
\begin{wrapfigure}{r}{0.35\textwidth}
	\vspace*{-2em}
	\raggedleft\scriptsize
	\setlength{\figurewidth}{.3\textwidth}
	\setlength{\figureheight}{.75\figurewidth}
	\pgfplotsset{axis on top,scale only axis,width=\figurewidth,height=\figureheight, ylabel near ticks,ylabel style={yshift=-2pt},xlabel style={yshift=3pt},y tick label style={rotate=90}, legend style={nodes={scale=0.8, transform shape}},tick label style={font=\tiny,scale=.8}}
	\pgfplotsset{legend cell align={left},every axis/.append style={legend style={draw=none,inner xsep=2pt, inner ysep=0.5pt, nodes={inner sep=2pt, text depth=0.1em},fill=white,fill opacity=0.8}}}
	\pgfplotsset{grid style={dotted, gray}}	
\begin{tikzpicture}

\definecolor{darkgray176}{RGB}{176,176,176}
\definecolor{darkorange25512714}{RGB}{255,127,14}
\definecolor{forestgreen4416044}{RGB}{44,160,44}
\definecolor{lightgray204}{RGB}{204,204,204}
\definecolor{steelblue31119180}{RGB}{31,119,180}

\begin{axis}[
legend cell align={left},
legend style={fill opacity=0.8, draw opacity=1, text opacity=1, draw=lightgray204},
tick align=outside,
tick pos=left,
x grid style={darkgray176},
xmajorgrids,
xmin=0, xmax=4,
xtick style={color=black},
y grid style={darkgray176},
ymajorgrids,
ymin=-0.0615, ymax=3.2715,
xlabel=Batches,
ylabel=$\|\vtheta^\star - \vtheta\|_2$,
ytick style={color=black}
]
\addplot [semithick, darkorange25512714, opacity=0.1, forget plot]
table {%
0 2.11
1 1.38
2 1.05
3 0.64
4 0.12
};
\addplot [semithick, darkorange25512714, opacity=0.1, forget plot]
table {%
0 2.12
1 1.24
2 0.95
3 0.4
4 0.15
};
\addplot [semithick, darkorange25512714, opacity=0.1, forget plot]
table {%
0 2.06
1 1.18
2 0.93
3 0.35
4 0.26
};
\addplot [semithick, darkorange25512714, opacity=0.1, forget plot]
table {%
0 2.11
1 1.12
2 0.81
3 0.4
4 0.14
};
\addplot [semithick, darkorange25512714, opacity=0.1, forget plot]
table {%
0 2.07
1 1.21
2 0.48
3 0.09
4 0.1
};
\addplot [semithick, steelblue31119180, opacity=0.1, forget plot]
table {%
0 2.45
1 2.08
2 1.35
3 1.04
4 0.1
};
\addplot [semithick, steelblue31119180, opacity=0.1, forget plot]
table {%
0 2.32
1 1.64
2 1.25
3 0.7
4 0.2
};
\addplot [semithick, steelblue31119180, opacity=0.1, forget plot]
table {%
0 2.26
1 1.48
2 1.02
3 0.4
4 0.36
};
\addplot [semithick, steelblue31119180, opacity=0.1, forget plot]
table {%
0 2.61
1 1.32
2 1.01
3 0.8
4 0.3
};
\addplot [semithick, steelblue31119180, opacity=0.1, forget plot]
table {%
0 2.67
1 1.01
2 0.88
3 0.14
4 0.2
};
\addplot [semithick, forestgreen4416044, opacity=0.1, forget plot]
table {%
0 3.11
1 2.38
2 1.75
3 1.44
4 1.12
};
\addplot [semithick, forestgreen4416044, opacity=0.1, forget plot]
table {%
0 3.12
1 2.24
2 1.75
3 1.4
4 1.15
};
\addplot [semithick, forestgreen4416044, opacity=0.1, forget plot]
table {%
0 3.06
1 2.18
2 1.73
3 1.35
4 1.26
};
\addplot [semithick, forestgreen4416044, opacity=0.1, forget plot]
table {%
0 3.11
1 2.12
2 1.61
3 1.4
4 1.14
};
\addplot [semithick, forestgreen4416044, opacity=0.1, forget plot]
table {%
0 3.07
1 2.21
2 1.58
3 1.09
4 1.1
};
\addplot [semithick, darkorange25512714]
table {%
0 2.094
1 1.226
2 0.844
3 0.376
4 0.154
};
\addlegendentry{Informative correlation}
\addplot [semithick, steelblue31119180]
table {%
0 2.462
1 1.506
2 1.102
3 0.616
4 0.232
};
\addlegendentry{No correlation}
\addplot [semithick, forestgreen4416044]
table {%
0 3.094
1 2.226
2 1.684
3 1.336
4 1.154
};
\addlegendentry{Misspecified correlation}
\end{axis}

\end{tikzpicture}
\\[-.8em]
	\caption{\textbf{Parameter Estimation:} Convergence of parameters with LM-suggested correlation as more batches are processed.}
	\label{fig:parameter_estimation_correlation}
	\vspace*{-3em}
\end{wrapfigure}
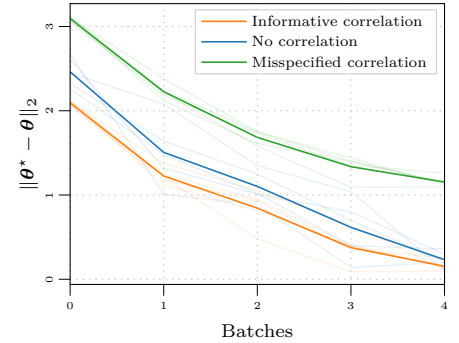
As outlined in \cref{sec:method_parameter_estimation}, we propose a Bayesian parameter estimation framework designed to estimate structural parameters in the presence of latent confounders. This approach incorporates language model (LM)-suggested prior knowledge into an Expectation-Maximization (EM) algorithm. The EM-step are defined in \cref{eq:e_step} and \cref{eq:m_step}.\looseness-1

In the experiment on the {\sc Red Wine Quality} dataset (\cf \cref{sec:exp_parameter_estimation}), we evaluate the impact of incorporating LM-suggested priors over the latent confounder. \cref{fig:parameter_estimation} illustrates how different priors affects estimation performance, demonstrating the robustness of the proposed Bayesian method.

Additionally, we analyze the effect of LM-suggested correlation values $\rho(A, B)$ and investigate the robustness of the proposed parameter estimation algorithm. From the observational data we know that the confounder \textit{alcohol} has a correlation of $\rho(\text{alcohol}, \text{density})=-0.50$ and
$\rho(\text{alcohol}, \text{quality})=0.48$. We set the correlation regularizer as $\lambda=5$. 

For parameter estimation algorithm, we use stochastic gradient descent with $0.001$ learning rate and run the algorithm for maximum for maximum of $20$ E-steps and $50$ M-steps with an early stop criteria based on the expected log-likelihood value.

\section{Experiment Details}
\label{app:experiment_details}
This section provides detailed descriptions of the experimental setup, including sequential data setup and batching strategies, LM prompts used in the experiments, details about the FCI variants, LLM-first, ILS-CSL, and \ours framework.
\begin{figure*}[t!]
	\centering
	\scriptsize
\begin{tikzpicture}[
	node/.style={draw, rounded corners=2pt, align=center, font=\scriptsize, text width=2cm},
	arrow/.style={-{Latex[length=2mm]}, thin}
	]

	\node[node] (Sugar) at (0, 2) {Residual Sugar};
	\node[node] (Density) at (3, 2) {Density};
	\node[node] (VolAcid) at (0, 0) {Volatile Acidity};
	\node[node] (Sulfur) at (3, -1) {Total \mbox{Sulfur} Dioxide};
	\node[node] (Alcohol) at (6, 1) {Alcohol};
	\node[node] (Quality) at (3, 1) {Quality};

	\draw[arrow] (Sugar) -- (Density);
	\draw[arrow] (Sugar) -- (Sulfur);
	\draw[arrow] (Alcohol) -- (Density);
	\draw[arrow] (VolAcid) -- (Quality);
	\draw[arrow] (Quality) -- (Sulfur);
	\draw[arrow] (Alcohol) -- (Quality);
	
\end{tikzpicture}
 	\caption{Ground truth causal structure for the {\sc Red Wine Quality} dataset illustrating the directional dependencies among key physicochemical attributes affecting wine quality. The graph highlights how alcohol, volatile acidity, total sulfur dioxide, and density (influenced by residual sugar) contribute directly or indirectly to the perceived quality of red wine.}
	\label{fig:redwine_true_graph}
	\vspace*{-1em}
\end{figure*}
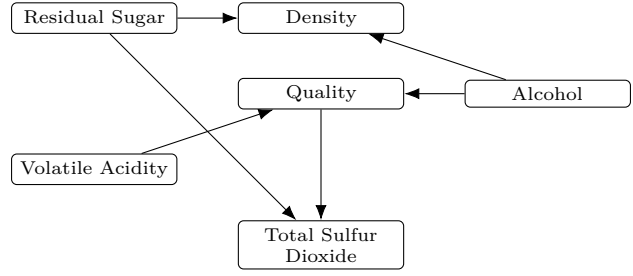

\subsection{Sequential Data Setup}
We describe the setup for each dataset category based on the availability of ground-truth causal structure and the strategy used for batch construction.

\textsc{Earthquake, Asia}
These datasets provide both the true causal structure and observational data. We use the ground-truth graphs shown in \cref{fig:earthquake_true_graph} and \cref{fig:asia_true_graph} as the true causal structures. The observational data are randomly partitioned into $6$ batches of $250$ samples each, without replacement. This results in data of shape $(6, 250, 5)$ for \textsc{Earthquake} and $(6, 250, 8)$ for \textsc{Asia}. Since all variables are binary (Yes/No), we preprocess the data by mapping \texttt{Yes} to 1 and \texttt{No} to 0.

\textsc{User-Level Data I, User-Level Data II}
For these datasets, ground-truth causal structures are unavailable. We apply DirectLiNGAM to the full observational dataset to estimate a causal graph, which we then treat as the true causal structure for generating synthetic data. The resulting graphs are visualized in \cref{fig:user_data_1_true_graph} and \cref{fig:user_data_2_true_graph}. Using the estimated structure, we simulate data via a linear SEM with additive Gaussian noise, sampled from $\N(0,0.05)$. We construct $7$ sequential batches with varying sizes. For {\sc user level data I}, the batchsize is $[3000,1000,2000,4000,3000,2000,5000]$ while for {\sc user level data II} we set the batch size as $[2000,1000,2000,1000,1000,3000,1000]$.  To simulate realistic distributions, in \textsc{User-Level Data I}, the parent node \textit{Proximity To Transaction} is sampled from $\N(10, 1)$ while in \textsc{User-Level Data II}, the parent node \textit{Number Of Page Hits Last Month} is sampled from $\N(27,10.5)$.

\textsc{Child, Alarm} 
These datasets provide both the true causal structure and observational data. We use the ground-truth graphs from \citet{beinlich1989alarm} and \citet{spiegelhalter1993bayesian} respectively as the true causal structures. For {\sc Child} dataset,  the observational data are randomly partitioned into $6$ batches of $2000$ samples each, without replacement. This results in data of shape $(6, 2000, 20)$. For {\sc Alarm} dataset, we split the observational data randomly, without replacement, into six batches with $50$ data points per batch. This is done with the aim to mimic the data-scarce scenario and showcase the robustness of \ours.

\textsc{Red Wine Quality}
This dataset provides observational data, which we use primarily to evaluate the proposed \textit{parameter estimation} framework. We estimate the causal structure using DirectLiNGAM on the full dataset, as shown in \cref{fig:redwine_true_graph}. Existence of a latent confounder, \textit{alcohol}, between \textit{density} and \textit{quality} makes this dataset suitable for testing latent-aware estimation. The full dataset consists of $1599$ samples across $6$ variables, which we partition into $5$ batches of sizes $[319,319,319,319,323]$ on which we perform parameter estimation.

\subsection{LM Prompts}
\label{app:prompts}
In this section, we provide the prompts used in the experiments. 

The pairwise prompt is:
\begin{lstlisting}[frame=single]
	System message: 
	
	You are an expert in Causal discovery and are studying {experiment_name}. You are using your knowledge to help build a causal model that contains  all the assumptions about {experiment_name}, where a causal model is a conceptual model that describes the causal mechanisms of a system. You will do this by by answering questions about cause and effect and using your domain knowledge as an expert in Causal discovery. We are considering the following variables: {variables}. The description of the variables is as follows:{variable_description}.
	
	User message:
	
	From your perspective as an expert in Causal discovery, which of the following is  most likely true? 
	(A) {var1} affects/causes {var2}; {var1} has a high likelihood of directly influencing {var2}; 
	(B) {var2} affects/causes {var1};  {var2} has a high likelihood of directly influencing {var1}; 
	(C) Neither A nor B; There is no causal relationship between {var1} and {var2}.
	(D) Do not know about the causal relationship between {var1} and {var2}. 
	Select the answer. Think step by step and provide your thoughts with the "thoughts" key and the answer with the "answer" key.
	Return a JSON with the following format:
	{
		"answer": "A/B/C/D",
		"thoughts": "step-by-step thought"
	}
	NOTE: Only return the JSON and nothing else.  
	
\end{lstlisting}

The triplet prompt is:
\begin{lstlisting}[frame=single]
	System message: 
	
	You are an expert in Causal discovery and are studying {experiment_name}. You are using your knowledge to help build a causal model that contains  all the assumptions about {experiment_name}, where a causal model is a conceptual model that describes the causal mechanisms of a system. We are considering the following variables: {variables}. The description of the variables is as follows:{variable_description}.
	
	User message:
	
	As an expert in Causal discovery, consider the following variables and output a causal DAG, {var1}, {var2}, {var3}. Only consider direct causal effects. If a variable has no causal relationship with any other, include it as an isolated node.
	For example, if Z is independent, and X causes/affects Y, the output DAG should be: [["X", "Y"], ["Z"]]. Think step by step and provide your thoughts with the "thoughts" key.
	Return a JSON in the following format:
	{
		"dag": [["source_node", "target_node"], ..., ["isloated_node"]],
		"thoughts": "step-by-step thought"
	}
	NOTE: Only return the JSON object and nothing else.
\end{lstlisting}

The PAG-Pairwise prompt is:
\begin{lstlisting}[frame=single]
	System message:
	
	You are an expert in Causal discovery and are studying {experiment_name}. You are using your knowledge to help build a causal model that contains  all the assumptions about {experiment_name}, where a causal model is a conceptual model that describes the causal mechanisms of a system. You will do this by by answering questions about cause and effect and using your domain knowledge as an expert in Causal discovery. We are considering the following variables: {variables}. The description of the variables is as follows:{variable_description}.
	
	User message:
	
	From your perspective as an expert in Causal discovery, which of the following is  most likely true?
	A: {var1} affects/causes {var2}; {var1} has a high likelihood of directly influencing {var2}; 
	B: {var2} affects/causes {var1};  {var2} has a high likelihood of directly influencing {var1}; 
	C: Neither A nor B; There is no causal relationship between {var1} and {var2}. 
	D: Do not know about the causal relationship between {var1} and {var2}. 
	E: There is a possible latent confounder between {var1} and {var2} i.e. {var1} <-> {var2} 
	F: Can not be sure about the causal relationship however {var1} is not an ancestor of {var2} i.e. {var1} o-> {var2} 
	G: Can not be sure about the causal relationship i.e. {var1} o-o {var2} 
	Select the answer. Think step by step and provide your thoughts with the "thoughts" key and the answer with the "causal_option" key.
	Return a JSON with the following format:
	{{
			"causal_option": "A/B/C/D/E/F/G",
			"thoughts": step-by-step thought
	}}
	NOTE: Only return the JSON and nothing else.
\end{lstlisting}
The \textit{LM-expert} prompt to get prior over the latent confounder
\begin{lstlisting}[frame=single]
	System message:
	
	You are an expert in Causal discovery and are studying {experiment_name}. You are using your knowledge to help information about the latent confounder variable in the causal structure. You will do this by giving a marginal Gaussian distribution over the detected confounder variable along with its correlation with the connected variables. We are considering the following variables: {variables}. The description of the variables is as follows:{variable_description}.
	
	User message : 
	
	We know that there is a confounder {confounder} between {variable_1} and {variable_2}. Provide an informative marginal Gaussian on the {confounder} between {variable_1} and {variable_2} using historical data and world knowledge.
	Think step-by-step.
	Return a JSON with the following format.
	{{
			"mean": "numerical mean value of prior N({latent})",
			"variance": "numerical variance value of prior N({latent})",
			"correlation": "dictionary of correlation numerical values",
	}}
	NOTE: Only return the JSON and nothing else.
\end{lstlisting}

The \textit{LM-expert} prompt:
\begin{lstlisting}[frame=single]
	System message:
	
	You are an expert in Causal discovery and are studying {experiment_name}. You are using your knowledge to help build a causal model that contains  all the assumptions about {experiment_name}, where a causal model is a conceptual model that describes the causal mechanisms of a system. You will do this by by answering questions about cause and effect and using your domain knowledge as an expert in Causal discovery. We are considering the following variables: {variables}. The description of the variables is as follows:{variable_description}. 
	
	User message:
	
	You are asked to determine the causal relationship between {A} and {B}. Only consider direct relationships and not indirect ones.The options available are limited to FCI output i.e. PAG edges (Use 0,2,3,4,6 options carefully and think more about 1,2,5 as we need more of these options):
	0: There is no causal relationship between {A} and {B}.
	1: Changing the state of node {A} causally affects a change in another node {B} i.e. A->B 
	2: There is a possible latent confounder between {A} and {B} 
	3: Can not be sure about the causal relationship however {A} is not an ancestor of {B} i.e. {A} o-> {B} 
	4. Can not be sure about the causal relationship, i.e., {A} o-o {B} or {B} o-o {A} 
	5. Changing the state of node which says {B} causally affects a change in another node which says {A}, i.e. B->A
	6. Can not be sure about the causal relationship however {B} is not an ancestor of {A}, {B} o-> {A}
	
	Response format:
	{
		"option": option_tag,
		"thoughts": "step-by-step thought"
	}
	We know the following causal relationships: {known_relationship}
	Be extra thoughtful and careful about the relationships you are describing by considering both ({A}, {B}) and ({B}, {A}) before answering. 
	NOTE: Only return the JSON object and nothing else.
\end{lstlisting}

The \textit{LM-confounder} prompt:
\begin{lstlisting}[frame=single]
	System message:
	
	Consider you are a causal discovery expert. We are creating a causal structure of {experiment_name} by refining the PAG obtained from fast causal inference (FCI) algorithm, where we are observing the following variables: {variables}. The description of the variables is as follows:{variable_description}. 
	Your goal is to output a possible confounder variable name with MAXIMUM 3 words. Be specific when providing the confounder variable and DO NOT provide a generic one like user behavior, etc.  In rare situations, when you are not sure about the confounder, return 'undefined' as the latent confounder name. Also, confounder should be outside the observed variables. 
	NOTE: Be precise, and ONLY return confounder variable names with no explanation.
	
	User message:
	
	We know that there is a latent confounder between {latent_0} and {latent_1}. Can you identify the latent confounder variable name? 
\end{lstlisting}

\subsection{Fast Causal Inference (FCI) Variants}
In the interested setting where data arrive sequentially in batches, running standard causal discovery algorithms like FCI on the entire dataset is infeasible. To address this, we adapt the Fast Causal Inference (FCI) algorithm to sequential processing by designing multiple baselines that reflect different trade-offs. Each variant operates under a restricted lookback window size, and adapts the FCI procedure accordingly.

For all FCI-based methods, we set the significance level to $\alpha = 0.3$ for the {\sc User Level Data II} dataset, and $\alpha = 0.1$ for all other datasets. We use the \textit{chisq} test for conditional independence on the {\sc Asia} and {\sc Earthquake} datasets (due to their binary variables), and the \textit{fisherz} test for all remaining experiments involving continuous data.

\paragraph{FCI-Cumulative} 
This variant applies the FCI algorithm to the cumulative data up to batch $i$, \ie $\GD{i}=\text{FCI}(\DD{_{1:i}})$. It serves as an upper bound on performance, assuming full access to all prior data.

\paragraph{FCI-Vanilla} 
This variant applies FCI independently to each incoming batch, $\GD{i}=\text{FCI}(\DD{i})$. Here, the algorithm forgets all previous knowledge, mimicking a naive local learner. This variant tests whether single-batch inference is sufficient and highlights the limitations of ignoring \textit{data bias}.

\paragraph{FCI-Iterative} 
This variant introduces background knowledge by passing the output of the previous batch's FCI run as input to the next, $\GD{i}=\text{FCI}(\DD{i},\B_i=\GD{(i-1)})$. Here, the background knowledge includes previously inferred causal structure. This allows FCI to refine its structure incrementally.

\paragraph{FCI-Heuristics}
In this variant, an edge from previous iterations is included in the background knowledge only if a heuristic threshold is met. Specifically, an edge is included in the background knowledge  if it has appeared in the FCI outputs of at least $h$ of all the past batches (we set $h{=}2$ in our experiments). This strategy balances between overfitting to noise (as in FCI-Iterative) and excessive forgetting (as in FCI-Vanilla).

\subsection{LLM-first}
In the \textbf{LLM-first} method, we reverse the standard causal discovery pipeline by first using a language model (LM) to propose an initial causal graph, which is then incrementally refined using sequential batches of observational data. To obtain the initial causal structure, we use a pairwise prompting strategy where the LM is queried on each variable pair. This results in a causal structure $\GE{}$, where each edge $A \rightarrow B$ indicates a predicted causal relation. While this initial graph often captures plausible structural patterns, it suffers from overly-optimistic behavior (\cf \cref{tbl:llm_optimistic_nature}) and lacks data grounding.  We then sequentially refine $\GE{}$ using batches of observed data. At each step $i$, we update the causal graph $\GE{(i-1)}$ using the FCI produced causal structure conditioned on the batch $i$. This adds, removes, or reorients edges based on statistical tests applied to the current data batch \( \mathcal{D}_i \), thus improving the reliability of the structure over time.

\subsection{Iterative LLM-Supervised Causal Structure Learning (ILS-CSL)}
\citet{ban2023causal} proposed Iterative LLM-Supervised Causal Structure Learning (ILS-CSL) algorithm that integrates natural language causal knowledge into the structure learning process by iteratively refining a causal graph using response of a large language model (LLM). Originally designed as a score-based method, ILS-CSL supervises structure learning by posing pairwise causal queries to the LLM and using its responses as soft constraints during graph optimization. For consistency and a fair comparison with our method, we adapt ILS-CSL to work with the FCI framework. Specifically, we treat the causal constraints inferred from the LLM as background knowledge and inject them into the FCI algorithm. This hybrid adaptation retains the core iterative supervision strategy of ILS-CSL while operating within a constraint-based causal discovery framework.\looseness-1

\begin{table*}[!t]
	\centering \scriptsize
	\caption{\textbf{Causal Discovery, Impact of LM:} We experiment with GPT-4o and GPT-5 on the {\sc Earthquake} and {\sc Asia} dataset and report \textit{Modified SHD} $\downarrow$ with temperature $1$. We report mean and standard deviation over $5$ showcasing the LM-agnostic nature of \ours and superior performance across models. The inference cost of recent models, GPT-4o and GPT-5, are quite a bit more than GPT-3.5\textsubscript{turbo}, constraining use of them for all large set of experiments.} 
	\label{tbl:gpt4o_gpt5_results}
	\begin{tabular}{lcccc}
		\toprule
		\textbf{Dataset} & \textbf{Method} & \textbf{Mod. SHD (GPT-3.5\textsubscript{turbo})} & \textbf{Mod. SHD (GPT-4o)} & \textbf{Mod. SHD (GPT-5)}\\
		\midrule
		\sc{Earthquake} & LLM-First & $6.00 \pm 0.82$ & $5.80 \pm 0.75$ & $5.50 \pm 0.50$ \\
		& ILS-CSL & $2.38 \pm 0.96$ & $1.50 \pm 1.12$ & $1.40 \pm 1.02$ \\
		& \ours & $1.00 \pm 0.63$ & $0.90 \pm 0.51$ & $0.80 \pm 0.74$ \\
		\midrule
		\sc{Asia} & LLM-First & $7.33 \pm 0.94$ & $7.90 \pm 0.40$ & $7.00 \pm 1.00$\\
		& ILS-CSL & $6.50 \pm 0.50$ & $5.00 \pm 1.00$ & $5.00 \pm 0.89$ \\
		& \ours & $4.60 \pm 1.02$ & $3.60 \pm 1.01$ & $3.20 \pm 0.98$ \\
		\bottomrule
	\end{tabular}
	\vspace*{-1.1em}
\end{table*}

\subsection{\ours}
\ours relies on a set of hyperparameters that guide expert interaction and edge selection throughout sequential batches. The selection score used for deciding which edge to query is computed using \cref{eq:score} with weights $w_1$, $w_2$, and $w_3$. Additionally, for background threshold \cref{eq:dynamic_bg}, $\alpha$ and a maximum expert budget $\me$ per batch are used to regulate expert calls.

For the {\sc Asia}, {\sc Earthquake}, and {\sc User-Level Data I} datasets, for selection score we set ${(w_1, w_2, w_3) = (0.1, 0.6, 0.3)}$, with and background knowledge threshold with $\alpha = 0.3$ and maximum expert call budgets $\me = 20$, $\me = 50$, and $\me = 70$ respectively. A minimum threshold of $10$ is also imposed to ensure reliability. For the {\sc User-Level Data II} dataset, we use ${(w_1, w_2, w_3) = (0.3, 0.4, 0.3)}$, with $\alpha = 0.5$, maximum expert call budget $\me = 20$, and a minimum threshold of $5$. \cref{tbl:causal_discovery_metrics_all_t} showcases the \ours---structure learning algorithm across LM temperatures.

\paragraph{GPT-4o and GPT-5}
In \cref{tbl:llm_optimistic_nature}, we illustrate the issue of LLM over-optimism using both GPT-3.5\textsubscript{turbo} and GPT-4o. We find that GPT-4o does not always outperform GPT-3.5\textsubscript{turbo}. We further demonstrate how an LM can be viewed as a noisy expert; yet can be integrated into a Bayesian causal structure discovery framework. The inference cost of recent models, GPT-4o and GPT-5, are quite a bit more than GPT-3.5\textsubscript{turbo}, constraining use of them for all large set of experiments. That said, \ours is fully model-agnostic, and stronger LMs are expected to further improve \ours's performance by providing more accurate priors. As an evidence, we experiment with GPT-4o and GPT-5 model on {\sc Earthquake} and {\sc Asia} dataset and showcase the superior performance (Modified SHD) in \cref{tbl:gpt4o_gpt5_results}.

\begin{table*}[t!]
	\centering \scriptsize
	\caption{\textbf{Causal Discovery, Impact of LM Temperature:} \cref{tbl:causal_discovery_metrics} shows evaluation metrics only for \textit{temperature=1}; here we show for other two \textit{temperatures=\{0.0,0.5\}}. The conclusions are remarkably similar and favor \ours. We show results for all six datasets using the two paradigms: \textit{Only-Data} and \textit{Data-LM}. We evaluate with the following metrics: \textit{Modified SHD, SID, Precision, Recall, F1}. All the methods use GPT-3.5\textsubscript{turbo} as an LM-expert and we report mean and standard deviation over $5$ runs. }   
	\label{tbl:causal_discovery_metrics_all_t}
	\setlength{\tabcolsep}{3.5pt}
	\resizebox{\textwidth}{!}{
\begin{tabular}{cclccccc}
	\toprule
	\textbf{Dataset} & \textbf{Approach} & \textbf{Method} & \textbf{Mod. SHD $\downarrow$} & \textbf{SID $\downarrow$} & \textbf{Precision $\uparrow$} & \textbf{Recall $\uparrow$} & \textbf{F1 Score $\uparrow$} \\
	\midrule
\multirow{10}{*}{\rotatebox{90}{\dataset{Earthquake}{5}}} & \multirow{4}{*}{Only-Data} & FCI-Cumulative & \val{2.00}{0.00} & \val{(0.00, 5.00)}{(0.00, 0.00)} & \val{1.00}{0.00} & \val{0.50}{0.00} & \val{0.67}{0.00} \\
	&& FCI-Vanilla & \val{3.60}{0.80} & \val{(8.20, 9.20)}{(3.60, 1.60)} & \val{0.20}{0.40} & \val{0.05}{0.10} & \val{0.08}{0.16} \\
	&& FCI-Iterative & \val{5.00}{1.67} & \val{(12.20, 12.20)}{(4.66, 4.66)} & \val{0.30}{0.27} & \val{0.20}{0.19} & \val{0.24}{0.22} \\
	&& FCI-Heuristics & \val{3.60}{0.80} & \val{(8.20, 9.20)}{(3.60, 1.60)} & \val{0.20}{0.40} & \val{0.05}{0.10} & \val{0.08}{0.16} \\
	\cdashline{2-8}[.4pt/1pt]
	& \multirow{3}{*}{Data-LM\textsubscript{(t=0)}} & LLM-first & \val{6.00}{0.82} & \val{(15.00, 15.00)}{(0.82, 0.82)} & \val{0.11}{0.16} & \val{0.08}{0.12} & \val{0.09}{0.13} \\
	&& ILS-CSL & \val{2.00}{0.89} & \val{(5.00, 5.00)}{(0.89, 0.89)} & \val{0.93}{0.13} & \val{0.55}{0.19} & \val{0.67}{0.17} \\
	 & & 	\cellcolor{highlight} \ours &	\cellcolor{highlight} \val{1.00}{0.63} &	\cellcolor{highlight} \val{(2.20, 2.20)}{(1.60, 1.60)} & 	\cellcolor{highlight} \val{1.00}{0.00} & 	\cellcolor{highlight} \val{0.75}{0.16} & 	\cellcolor{highlight} \val{0.85}{0.11}  \\
	\cdashline{2-8}[.4pt/1pt]
	& \multirow{3}{*}{Data-LM\textsubscript{(t=0.5)}} & LLM-first & \val{6.00}{0.82} & \val{(15.00, 15.00)}{(0.82, 0.82)} & \val{0.11}{0.16} & \val{0.08}{0.12} & \val{0.09}{0.13} \\
	&& ILS-CSL & \val{2.00}{0.89} & \val{(5.00, 5.00)}{(2.61, 2.61)} & \val{0.93}{0.13} & \val{0.55}{0.19} & \val{0.67}{0.17} \\
	\rowcolor{highlight}\cellcolor{white} &\cellcolor{white}& \ours & \val{1.00}{0.63} & \val{(2.20, 2.20)}{(1.60, 1.60)} & \val{1.00}{0.00} & \val{0.75}{0.16} & \val{0.85}{0.11}  \\
	\midrule
	\multirow{10}{*}{\rotatebox{90}{\dataset{Asia}{8}}}  & \multirow{4}{*}{Only-Data} & FCI-Cumulative & \val{7.00}{0.00} & \val{(23.00, 49.00)}{(0.00, 0.00)} & \val{0.00}{0.00} & \val{0.00}{0.00} & \val{0.00}{0.00} \\
	&& FCI-Vanilla & \val{7.80}{0.75} & \val{(30.00, 35.00)}{(5.90, 2.45)} & \val{0.00}{0.00} & \val{0.00}{0.00} & \val{0.00}{0.00} \\
	&& FCI-Iterative & \val{8.00}{1.26} & \val{(33.00, 33.00)}{(7.46, 7.46)} & \val{0.45}{0.24} & \val{0.23}{0.15} & \val{0.29}{0.18} \\
	&& FCI-Heuristics & \val{7.80}{0.75} & \val{(30.00, 35.00)}{(5.90, 2.45)} & \val{0.00}{0.00} & \val{0.00}{0.00} & \val{0.00}{0.00} \\
	\cdashline{2-8}[.4pt/1pt]
	& \multirow{3}{*}{Data-LM\textsubscript{(t=0)}} & LLM-first &\val{7.33}{0.94} & \val{(27.67, 27.67)}{(2.49, 2.49)} & \val{0.58}{0.12} & \val{0.29}{0.06} & \val{0.39}{0.08} \\
	&& ILS-CSL & \val{5.20}{0.75} & \val{(21.80, 21.80)}{(1.94, 1.94)} & \val{0.90}{0.10} & \val{0.38}{0.08} & \val{0.53}{0.09}  \\
   && \cellcolor{highlight}  \ours & \cellcolor{highlight}\val{4.60}{1.02}  & \cellcolor{highlight}\val{(13.60, 13.60)}{(3.83, 3.83)}  & \cellcolor{highlight}\val{0.80}{0.12} & \cellcolor{highlight}\val{0.60}{0.12} & \cellcolor{highlight}\val{0.67}{0.08}  \\
	\cdashline{2-8}[.4pt/1pt]
	& \multirow{3}{*}{Data-LM\textsubscript{(t=0.5)}} & LLM-first &\val{7.33}{0.94} & \val{(27.67, 27.67)}{(2.49, 2.49)} & \val{0.58}{0.12} & \val{0.29}{0.06} & \val{0.39}{0.08} \\
	&& ILS-CSL & \val{5.60}{1.36} & \val{(22.80, 22.80)}{(3.19, 3.19)} & \val{0.85}{0.15} & \val{0.35}{0.09} & \val{0.49}{0.12}  \\
	&& \cellcolor{highlight}  \ours & \cellcolor{highlight}\val{4.60}{1.02}  & \cellcolor{highlight}\val{(13.60, 13.60)}{(3.83, 3.83)}  & \cellcolor{highlight}\val{0.80}{0.12} & \cellcolor{highlight}\val{0.60}{0.12} & \cellcolor{highlight}\val{0.67}{0.08}  \\
	\midrule
	\multirow{10}{*}{\rotatebox{90}{\tiny\dataset{User Level Data-I}{9}}} & \multirow{4}{*}{Only-Data} & FCI-Cumulative & \val{15.00}{2.77} & \val{(47.80, 47.80)}{(4.62, 4.62)} & \val{0.72}{0.18} & \val{0.37}{0.09} & \val{0.47}{0.09} \\
	&& FCI-Vanilla & \val{21.30}{3.50} & \val{(61.60, 61.60)}{(5.54, 5.54)} & \val{0.41}{0.18} & \val{0.22}{0.10} & \val{0.29}{0.13} \\
	&& FCI-Iterative & \val{\phantom{0}5.60}{1.20} & \val{(23.40, 23.40)}{(7.94, 7.94)} & \val{0.93}{0.04} & \val{0.77}{0.02} & \val{0.84}{0.03} \\
	&& FCI-Heuristics & \val{21.30}{3.50} & \val{(61.60, 61.60)}{(5.54, 5.54)} & \val{0.41}{0.18} & \val{0.22}{0.10} & \val{0.29}{0.13} \\
	\cdashline{2-8}[.4pt/1pt]
	& \multirow{3}{*}{Data-LM\textsubscript{(t=0)}} & LLM-first & \val{8.40}{1.20} & \val{(35.20, 35.20)}{(2.92, 2.92)} & \val{0.83}{0.05} & \val{0.69}{0.02} & \val{0.76}{0.03}  \\
	&& ILS-CSL & \val{8.50}{2.60} & \val{(30.25, 30.25)}{(9.09, 9.09)} & \val{0.83}{0.08} & \val{0.69}{0.08} & \val{0.75}{0.07}  \\
	&& \cellcolor{highlight}  \ours & \cellcolor{highlight}\val{5.40}{0.49}  & \cellcolor{highlight}\val{(13.20, 13.20)}{(2.40, 2.40)} & \cellcolor{highlight}\val{0.88}{0.02} & \cellcolor{highlight}\val{0.83}{0.02} & \cellcolor{highlight}\val{0.85}{0.01} \\
	\cdashline{2-8}[.4pt/1pt]
	& \multirow{3}{*}{Data-LM\textsubscript{(t=0.5)}} & LLM-first & \val{9.60}{1.69} & \val{(38.70, 38.70)}{(4.94, 4.94)} & \val{0.80}{0.06} & \val{0.67}{0.03} & \val{0.73}{0.05}  \\
	&& ILS-CSL & \val{7.67}{1.70} & \val{(26.33, 26.33)}{(3.68, 3.68)} & \val{0.84}{0.05} &\val{0.74}{0.04}& \val{0.79}{0.05}  \\
	&& \cellcolor{highlight}  \ours & \cellcolor{highlight}\val{4.60}{0.49}  & \cellcolor{highlight}\val{(12.00, 12.00)}{(0.00, 0.00)} & \cellcolor{highlight}\val{0.91}{0.03} & \cellcolor{highlight}\val{0.84}{0.00} & \cellcolor{highlight}\val{0.87}{0.01}  \\
	\midrule
	\multirow{10}{*}{\rotatebox{90}{\tiny\dataset{User Level Data-II}{8}}} & \multirow{4}{*}{Only-Data} & FCI-Cumulative & \val{19.80}{2.04} & \val{(40.00, 40.00)}{(3.03, 3.03)} & \val{0.15}{0.07} & \val{0.10}{0.04} & \val{0.12}{0.05} \\
	&& FCI-Vanilla & \val{17.40}{1.83} & \val{(36.80, 39.40)}{(2.04, 3.38)} & \val{0.07}{0.13} & \val{0.02}{0.03} & \val{0.03}{0.05} \\
	&& FCI-Iterative & \val{20.60}{1.77} & \val{(42.80, 43.40)}{(4.07, 4.22)} & \val{0.06}{0.08} & \val{0.05}{0.07} & \val{0.05}{0.07} \\
	&& FCI-Heuristics & \val{17.40}{1.83} & \val{(36.80, 39.40)}{(2.04, 3.38)} & \val{0.07}{0.13} & \val{0.02}{0.03} & \val{0.03}{0.05} \\
	\cdashline{2-8}[.4pt/1pt]
	& \multirow{3}{*}{Data-LM\textsubscript{(t=0)}} & LLM-first & \val{18.33}{0.94}&\val{(42.67, 42.67)}{(0.47, 0.47)}&\val{0.21}{0.04}&\val{0.19}{0.04}&\val{0.20}{0.04} \\
	&& ILS-CSL &\val{19.00}{1.22}&\val{(46.75, 46.75)}{(4.87, 4.87)}&\val{0.14}{0.06}&\val{0.13}{0.07}&\val{0.13}{0.07} \\
	&& \cellcolor{highlight}  \ours & \cellcolor{highlight}\val{18.67}{1.70}  & \cellcolor{highlight}\val{(41.00, 41.00)}{(1.63,1.63)} & \cellcolor{highlight}\val{0.22}{0.02} & \cellcolor{highlight}\val{0.22}{0.04} & \cellcolor{highlight}\val{0.22}{0.02} \\
	\cdashline{2-8}[.4pt/1pt]
	& \multirow{3}{*}{Data-LM\textsubscript{(t=0.5)}} & LLM-first &\val{18.33}{0.94}&\val{(42.67, 42.67)}{(0.47, 0.47)}&\val{0.21}{0.04}&\val{0.19}{0.04}&\val{0.20}{0.04} \\
	&& ILS-CSL &\val{17.60}{2.33}&\val{(41.20, 41.20)}{(4.12, 4.12)}&\val{0.21}{0.11}&\val{0.18}{0.12}&\val{0.19}{0.12}\\
	&& \cellcolor{highlight}  \ours & \cellcolor{highlight}\val{18.67}{2.05}  & \cellcolor{highlight}\val{(41.00, 41.00)}{(1.63,1.63)} & \cellcolor{highlight}\val{0.23}{0.02} & \cellcolor{highlight}\val{0.22}{0.04} & \cellcolor{highlight}\val{0.22}{0.02}  \\
	\midrule
	\multirow{10}{*}{\rotatebox{90}{\dataset{Child}{19}}}  & \multirow{4}{*}{Only-Data} & FCI-Cumulative & \val{27.50}{0.00} & \val{(111.00, 131.00)}{(\phantom{0}0.00, \phantom{0}0.00)} & \val{0.38}{0.00} & \val{0.36}{0.00} & \val{0.37}{0.00} \\
	&& FCI-Vanilla & \val{28.00}{1.48} & \val{(129.20, 133.20)}{(10.46, 10.76)} & \val{0.38}{0.04} & \val{0.26}{0.05} & \val{0.31}{0.04} \\
	&& FCI-Iterative & \val{32.10}{1.16} & \val{(149.00, 164.40)}{(\phantom{0}7.16, 10.33)} & \val{0.27}{0.03} & \val{0.26}{0.04} & \val{0.26}{0.03} \\
	&& FCI-Heuristics & \val{28.00}{1.48} & \val{(129.20, 133.20)}{(10.46, 10.76)} & \val{0.38}{0.04} & \val{0.26}{0.05} & \val{0.31}{0.04} \\
	\cdashline{2-8}[.4pt/1pt]
	& \multirow{3}{*}{Data-LM\textsubscript{(t=0)}} & LLM-first & \val{29.67}{0.47} & \val{(154.83, 154.83)}{(\phantom{0}9.75, \phantom{0}9.75)} & \val{0.33}{0.01} & \val{0.35}{0.02} & \val{0.34}{0.01} \\
	&& ILS-CSL & \val{29.80}{2.48} & \val{(160.00, 160.00)}{(21.25, 21.25)} & \val{0.33}{0.05} & \val{0.35}{0.04} & \val{0.34}{0.04}  \\
	 & &  \cellcolor{highlight} \ours &  \cellcolor{highlight}\val{27.80}{0.75} &  \cellcolor{highlight} \val{(114.80, 114.80)}{(\phantom{0}7.76, \phantom{0}7.76)} &  \cellcolor{highlight} \val{0.39}{0.01} &  \cellcolor{highlight} \val{0.45}{0.03} &  \cellcolor{highlight} \val{0.42}{0.02} \\
	\cdashline{2-8}[.4pt/1pt]
		& \multirow{3}{*}{Data-LM\textsubscript{(t=0.5)}} & LLM-first & \val{30.67}{1.11} & \val{(146.83, 146.83)}{(\phantom{0}4.81, \phantom{0}4.81)} & \val{0.31}{0.03} & \val{0.32}{0.03} & \val{0.32}{0.03} \\
	&& ILS-CSL & \val{32.00}{2.28} & \val{(152.80, 152.80)}{(12.95, 12.95)} & \val{0.28}{0.05} & \val{0.29}{0.06} & \val{0.28}{0.06}  \\
	\rowcolor{highlight}\cellcolor{white} &\cellcolor{white}& \ours & \val{28.10}{0.86} & \val{(100.60, 112.00)}{(\phantom{0}9.31, \phantom{0}0.00)} & \val{0.40}{0.01} & \val{0.45}{0.00} & \val{0.42}{0.01} \\
	\midrule
	\multirow{10}{*}{\rotatebox{90}{\dataset{Alarm}{37}}} & \multirow{4}{*}{Only-Data} & FCI-Cumulative & \val{45.00}{0.00} & \val{(626.00, 626.00)}{(\phantom{0}0.00, \phantom{0}0.00)} & \val{0.25}{0.00} & \val{0.02}{0.00} & \val{0.04}{0.00} \\
	&& FCI-Vanilla & \val{49.50}{1.61} & \val{(617.80, 699.20)}{(29.23, 68.43)} & \val{0.00}{0.00} & \val{0.00}{0.00} & \val{0.00}{0.00} \\
	&& FCI-Iterative & \val{52.40}{6.21} & \val{(612.20, 636.80)}{(49.87, 43.83)} & \val{0.33}{0.14} & \val{0.12}{0.05} & \val{0.17}{0.07} \\
	&& FCI-Heuristics & \val{49.50}{1.61} & \val{(617.80, 699.20)}{(29.23, 68.43)} & \val{0.00}{0.00} & \val{0.00}{0.00} & \val{0.00}{0.00} \\
	\cdashline{2-8}[.4pt/1pt]
	& \multirow{3}{*}{Data-LM\textsubscript{(t=0)}} & LLM-first & \val{57.33}{1.89} & \val{(695.33, 695.33)}{(21.91, 21.91)} & \val{0.22}{0.03} & \val{0.10}{0.01} & \val{0.13}{0.01} \\
	&& ILS-CSL & \val{49.50}{1.50} & \val{(647.50, 647.50)}{(18.94, 18.94)} & \val{0.39}{0.05} & \val{0.10}{0.02} & \val{0.17}{0.03} \\
	 & & \cellcolor{highlight} \ours & \cellcolor{highlight} \val{51.60}{1.77} & \cellcolor{highlight} \val{(595.00, 595.80)}{(16.63, 16.94)} & \cellcolor{highlight} \val{0.40}{0.04} & \cellcolor{highlight} \val{0.22}{0.01} & \cellcolor{highlight} \val{0.28}{0.01} \\
	\cdashline{2-8}[.4pt/1pt]
	& \multirow{3}{*}{Data-LM\textsubscript{(t=0.5)}} & LLM-first & \val{56.67}{2.49} & \val{(684.67, 684.67)}{(16.34, 16.34)} & \val{0.23}{0.05} & \val{0.10}{0.02} & \val{0.14}{0.03} \\
   && ILS-CSL & \val{49.75}{1.48} & \val{(633.25, 633.25)}{(13.05, 13.05)} & \val{0.41}{0.02} & \val{0.12}{0.01} & \val{0.19}{0.01}  \\
	& & \cellcolor{highlight} \ours & \cellcolor{highlight} \val{51.70}{1.29} & \cellcolor{highlight} \val{(594.00, 595.20)}{(38.75, 38.75)} & \cellcolor{highlight} \val{0.41}{0.03} & \cellcolor{highlight} \val{0.22}{0.02} & \cellcolor{highlight} \val{0.29}{0.02} \\
	\bottomrule
\end{tabular}

 	}
	\vspace*{-1.1em}
\end{table*}

\section{Author Contributions}
The core research idea was conceived by Prakhar Verma and modified in discussion with other authors during an internship with Adobe Research. Prakhar Verma was responsible for modeling the scoring rule, parameter estimation, implementing the framework and conducting the experiments. Other authors pitched in with suggestions, some of which were incorporated in the final version. The first draft was written by Prakhar Verma and Atanu R. Sinha. All authors contributed to finalizing the manuscript.

\end{document}